\documentclass[10pt,twocolumn,letterpaper]{article}

\usepackage{wacv}
\usepackage{times}
\usepackage{epsfig}
\usepackage{graphicx}
\usepackage{amsmath}
\usepackage{amssymb}
\usepackage{subcaption}
\usepackage{adjustbox}
\usepackage[accsupp]{axessibility}
\def\wacvPaperID{****}

\wacvfinalcopy 

\ifwacvfinal
\usepackage[breaklinks=true,bookmarks=false]{hyperref}
\else
\usepackage[pagebackref=true,breaklinks=true,colorlinks,bookmarks=false]{hyperref}
\fi
\usepackage{cleveref}
\newcommand{\paperTitle}[0]{DeepPrivacy2: Towards Realistic Full-Body Anonymization}

\newcommand{\appFDH}[0]{Appendix \ref{sec:fdh_filtering}\xspace}
\newcommand{\appFDF}[0]{Appendix \ref{sec:fdf}\xspace}
\newcommand{\appDetails}[0]{Appendix \ref{sec:experimental_details}\xspace}
\newcommand{\appRandom}[0]{Appendix \ref{sec:market1501}\xspace}

\newcommand{\authorskip}{\hspace{10mm}}

\newcommand{\paperAuthor}[0]{
    H{\aa}kon Hukkel{\aa}s  \authorskip Frank Lindseth \\
Norwegian University of Science and Technology \\
{\tt\small hakon.hukkelas@ntnu.no}
}

\begin{document}

\title{\paperTitle}

\author{\paperAuthor}
\twocolumn[{%
\renewcommand\twocolumn[1][]{#1}%
\maketitle
\begin{center}
    \centering
    \captionsetup{type=figure}
    \includegraphics[width=\linewidth]{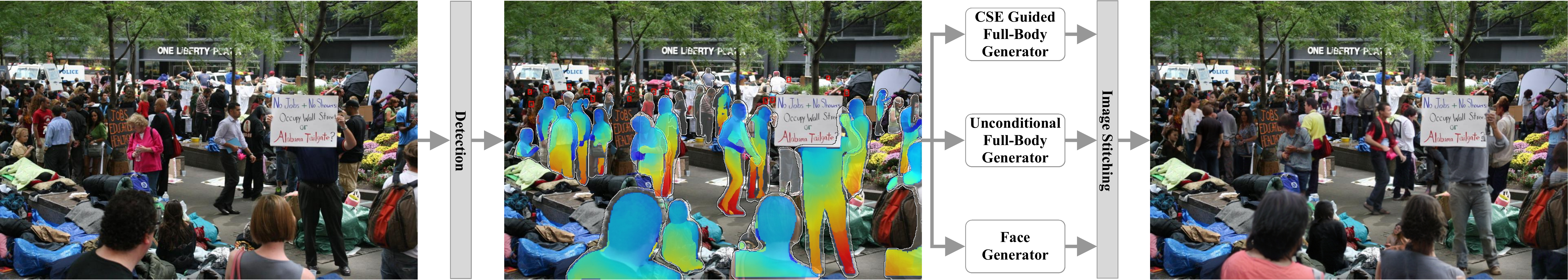}
    \captionof{figure}{
        DeepPrivacy2 detects and anonymizes individuals via three detection and synthesis networks;
        (1) a CSE-guided generator for individuals detected with dense pose (by CSE \cite{Neverova2020}),
        (2) an unconditional full-body generator for cases where CSE fails to detect (note the segmented persons without color-coded CSE detections),
        and  (3) a face generator for the remaining individuals (marked in red).
        The original image is from Wider-Face \cite{Yang2016WIDER}.
    }
    \label{fig:task_figure}
\end{center}%
}]
\begin{abstract}
Generative Adversarial Networks (GANs) are widely adopted for anonymization of human figures.
However, current state-of-the-art limits anonymization to the task of face anonymization.
In this paper, we propose a novel anonymization framework (DeepPrivacy2) for realistic anonymization of human figures and faces.
We introduce a new large and diverse dataset for full-body synthesis, which significantly improves image quality and diversity of generated images.
Furthermore, we propose a style-based GAN that produces high-quality, diverse, and editable anonymizations.
We demonstrate that our full-body anonymization framework provides stronger privacy guarantees than previously proposed methods.
Source code and appendix is available at: \href{http://github.com/hukkelas/deep_privacy2}{github.com/hukkelas/deep\_privacy2}.
 \end{abstract}

\begin{figure*}[t]
    \begin{subfigure}{0.1\textwidth}
        \includegraphics[width=\textwidth]{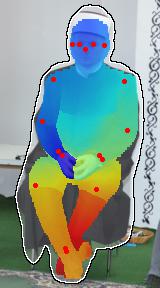}
    \end{subfigure}%
    \begin{subfigure}{0.1\textwidth}
        \includegraphics[width=\textwidth]{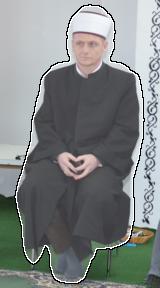}
    \end{subfigure}%
    \begin{subfigure}{0.1\textwidth}
        \includegraphics[width=\textwidth]{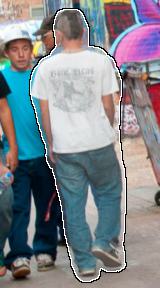}
    \end{subfigure}%
    \begin{subfigure}{0.1\textwidth}
        \includegraphics[width=\textwidth]{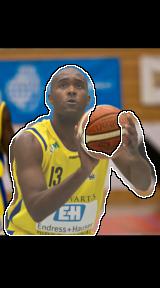}
    \end{subfigure}%
    \begin{subfigure}{0.1\textwidth}
        \includegraphics[width=\textwidth]{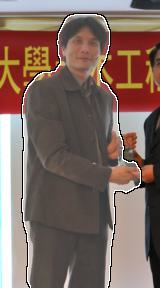}
    \end{subfigure}%
    \begin{subfigure}{0.1\textwidth}
        \includegraphics[width=\textwidth]{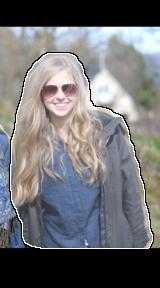}
    \end{subfigure}%
    \begin{subfigure}{0.1\textwidth}
        \includegraphics[width=\textwidth]{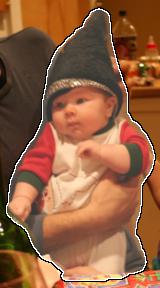}
    \end{subfigure}%
    \begin{subfigure}{0.1\textwidth}
        \includegraphics[width=\textwidth]{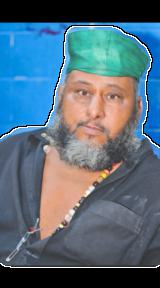}
    \end{subfigure}%
    \begin{subfigure}{0.1\textwidth}
        \includegraphics[width=\textwidth]{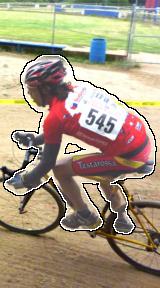}
    \end{subfigure}%
    \begin{subfigure}{0.1\textwidth}
        \includegraphics[width=\textwidth]{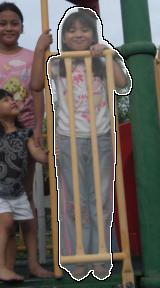}
    \end{subfigure}%
    \caption{Examples from the FDH dataset. Each image is annotated with keypoints, pixel-to-vertex correspondences (from CSE \cite{Neverova2020}) and a segmentation mask.
    The leftmost image shows annotations for the first image.}
    \label{fig:fdh_dataset_examples}
\end{figure*}

\section{Introduction}
Collecting and storing images is ubiquitous in our modern society, where a range of applications requires collecting privacy-sensitive data.
However, collecting such data without anonymization or consent from the individual is troublesome due to recently introduced legislation in many areas (\eg GDPR in EU).
Traditional image anonymization (\eg blurring) is widely adopted in practice; however, it severely distorts the data, making it unusable for future applications.
Recently, \emph{realistic anonymization} has been introduced as an alternative to traditional methods, where generative models can generate realistic faces fitting into a given context \cite{Gafni2019,Hukkelas2019,Maximov2020,Sun2018}.
However, current methods focus on face anonymization, which does not prevent recognition through identifiers outside the face, including both primary (\eg ears, gait \cite{Abarca2021handbook}) and secondary (\eg gender) identifiers.

Surface Guided GANs (SG-GAN) \cite{Hukkelas2022} propose a full-body anonymization GAN guided on dense pixel-to-surface correspondences from Continuous Surface Embeddings (CSE) \cite{Neverova2020}.
SG-GAN shows promising results for full-body anonymization, but their method often includes visual artifacts, degrading the image quality.
The authors attribute the limited visual quality to the dataset, where they use a derivate of COCO \cite{Lin2014COCO} containing 40K human figures.
Furthermore, the CSE segmentation used for anonymization  does not include accessories/hair on the human body; thus, the anonymized individual often "wears" these unsegmented areas (see \cref{fig:segmentation_comparison_SGGAN}).
Additionally, SG-GAN fails to anonymize many individuals, as the CSE detector often fails to  detect persons that are further away from the camera.

In this work, we extend Surface Guided GANs to address the limited visual quality and the insufficient anonymization due to poor segmentation.
Furthermore, we address cases where the CSE detector fails to detect individuals.
We summarize our contributions in the following.

First, we introduce the Flickr Diverse Humans (FDH) dataset.
The FDH dataset consists of 1.5M images of human figures in diverse contexts extracted from the YFCC100M \cite{Thomee2015} dataset.
We show that the larger dataset greatly benefits the visual quality of generated human figures.

Secondly, we propose a novel anonymization framework that combines detections from multiple modalities to improve the segmentation and detection of human figures.
Our anonymization framework divides image anonymization into three individual anonymizers; (1) for human figures that are detected with a dense pose estimation (CSE), (2) for human figures that CSE does not detect, and (3) for the remaining faces (see \cref{fig:task_figure}).
For each category, our framework employs a simple inpainting GAN that follows established GAN training techniques for unconditional image generation \cite{Karras2018,Karras2019b}.
We show that our GAN generates high-quality and diverse identities with few task-specific modeling choices.

Finally, we extend our GAN for face anonymization on an updated version of the Flickr Diverse Faces (FDF) dataset \cite{Hukkelas2019}.
In contrast to previous face anonymization techniques \cite{Gafni2019,Hukkelas2019,Maximov2020,Sun2018}, our GAN uses no pose guidance, enabling it to anonymize individuals where pose information is challenging to detect.
Furthermore, we show that our style-based generator can adapt methods from unconditional GANs to find global semantically meaningful directions in the GAN latent space.
This enables text-guided attribute editing for our anonymization pipeline.

DeepPrivacy2 surpasses all previous state-of-the-art realistic anonymization methods in terms of image quality and anonymization guarantees.
We validate the effectiveness of DeepPrivacy2 with extensive qualitative and quantitative evaluation.
Our code, pre-trained models, and the FDH dataset is available at \href{http://github.com/hukkelas/deep_privacy2}{github.com/hukkelas/deep\_privacy2}.
\section{Related Work}

\paragraph{Image Anonymization}
Naive image anonymization (\eg masking, blurring, pixelation) is widely adopted in practice;
however these methods severely degrade the quality of the anonymized image, making the data unusable for many applications.
Early work focused on the K-same family of algorithms \cite{Gross2006,Jourabloo2015,Newton2005}, which provides better privacy guarantees and data usability than naive methods , but generate highly corrupted images.
Recent work on deep generative models reflects that learning-based anonymization can realistically anonymize data while retaining its usability for downstream applications.
These methods anonymize face regions by either inpainting missing regions \cite{Hukkelas2019,Maximov2020,Sun2018,Sun2018a} or transforming \cite{Gafni2019,Ren2018a} the original face.
Our method anonymizes by inpainting, as inpainting-based methods provide stronger privacy guarantees than transformative methods, as they never observe the original privacy-sensitive information.
The majority of prior work focuses on face anonymization, which compromises privacy for many use cases, as they leave several primary (\eg ears, gait) and secondary (\eg gender) identifiers  on the human body untouched.
There is a limited amount of work focusing on full-body anonymization \cite{Brkic2017,Hukkelas2022,Maximov2020}, where prior methods are limited to low-resolution images \cite{Maximov2020} or generate images with visual artifacts \cite{Brkic2017,Hukkelas2022}.

\paragraph{Full-body Synthesis}
Recent work on full-body synthesis focus on limited tasks, such as transferring source appearances into new poses \cite{Balakrishnan2018,Chan2019,Li2019,Neverova2018,Sarkar2020}, with different garments \cite{Han2018,Sarkar2021,Sarkar2020}, or with new motion \cite{Chan2019}.
These methods are often guided on dense pixel-to-surface correspondences  or sparse keypoints annotations.
In contrast to these methods, our anonymization approach does not rely on a source appearance to transfer, and the majority of the aforementioned methods do not handle large variations in background contexts.
Furthermore, a number of these methods focus on low-variance datasets (\eg DeepFashion \cite{Liu2016a}), which consists of a limited number of identities in similar poses and a close-to static context (white background).
There is a limited amount of work focusing on full-body synthesis without a source appearance, where Ma \etal \cite{Sun2018person} proposes a pose-guided GAN for novel full-body synthesis.

\section{The Flickr Diverse Humans Dataset}
\label{sec:fdh_dataset}
The Flickr Diverse Humans (FDH) dataset consists of 1.53M images of human figures from the YFCC100M \cite{Thomee2015} dataset.
\Cref{fig:fdh_dataset_examples} shows examples from the dataset.
Each image contains a single human figure in the center, with a pixel-wise dense pose estimation from CSE \cite{Neverova2020}, 17 keypoint annotations from a keypoint R-CNN model \cite{He2017}, and a segmentation mask.
The segmentation mask is the union of the mask from a CSE detector and  Mask R-CNN \cite{He2017} trained on COCO.
The dataset is automatically filtered through confidence thresholding, automatic image quality assessment, the number of body parts visible in the image, and overlap between keypoint and CSE predictions (see \appFDH for more details).
Otherwise, we perform no further filtering such that the dataset includes individuals in all various contexts.
The resolution of each image is $288 \times 160$, and the dataset is split into 1,524,845 images for training and 30K images for validation.
Compared to previously adopted datasets for full-body synthesis \cite{Hukkelas2022,Liu2016a,Zheng2015}, FDH is much larger and contains a diverse set of individuals from in-the-wild images.
Additionally, FDH is less curated than typical datasets for generative modeling, where it includes human figures with unusual poses, perspectives, lighting conditions, and contexts.
This is to ensure that our anonymization method can handle such conditions.

\section{The DeepPrivacy2 Anonymization Pipeline}
This section outlines the core technologies used for our anonymization pipeline.
First, we present our ensemble detection pipeline, then our GAN-based synthesis method.

\subsection{Detection}
\label{sec:detection}
\begin{figure}
    \begin{subfigure}{0.333\linewidth}
        \includegraphics[width=\textwidth,trim={0cm 1.7cm 0cm 1.65cm },clip]{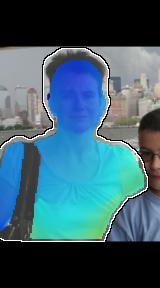}
        \caption*{Detection (Ours)}
    \end{subfigure}%
    \begin{subfigure}{0.333\linewidth}
        \includegraphics[width=\textwidth,trim={0cm 1.7cm 0cm 1.65cm },clip]{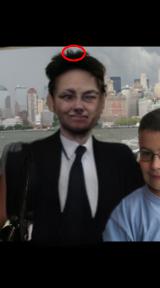}
        \caption*{SG-GAN \cite{Hukkelas2022}}
    \end{subfigure}%
    \begin{subfigure}{0.333\linewidth}
        \includegraphics[width=\textwidth,trim={0cm 1.7cm 0cm 1.65cm },clip]{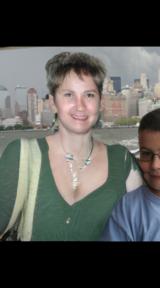}
        \caption*{Ours}
    \end{subfigure}
    \caption{SG-GAN \cite{Hukkelas2022} anonymizes only the area from a CSE segmentation (marked in blue tint), which does not include accessories/hair.
    This results in SG-GAN \cite{Hukkelas2022} anonymization often wearing the original hair/accessories of the original identity (marked in red).
    In contrast, DeepPrivacy2 anonymizes the segmentation from Mask R-CNN (outlined), which includes hair and clothing.}
    \label{fig:segmentation_comparison_SGGAN}
\end{figure}

The main objective of the detection module is to ensure that all individuals in the image are detected.
DeepPrivacy2 uses an ensemble of three detection networks from different modalities; DSFD \cite{Li2018c} for face detection, CSE \cite{Neverova2020} for dense pose estimation, and Mask R-CNN \cite{He2017} for instance segmentation.
The pipeline categorizes the detections into three categories; individuals with dense pose estimation (detection w/ CSE), individuals not detected by CSE (detection w/o CSE), and faces that are not included in the former categories.
For each category, we propose individual anonymization methods, introduced in \cref{sec:synthesis_method}.
For human figures, we anonymize the union of Mask R-CNN and CSE segmentations, such that accessories/hair detected by Mask R-CNN (but not CSE) are anonymized (see \cref{fig:segmentation_comparison_SGGAN}).
Note that dense pose estimation is not essential for privacy, but it substantially improves synthesized image quality.
Furthermore, the detections are tracked with a bounding box tracker, such that the anonymization can retain the same identity over a sequence of frames.
Compared to SG-GAN \cite{Hukkelas2022}, the ensemble of detectors significantly improves detection recall, as DeepPrivacy2 uses Mask R-CNN and DSFD for fallback detection when the CSE detector fails.

\begin{figure*}
    \begin{subfigure}[t]{0.3333\textwidth}
        \includegraphics[width=\textwidth,trim={0cm 0cm 0cm 3cm}, clip]{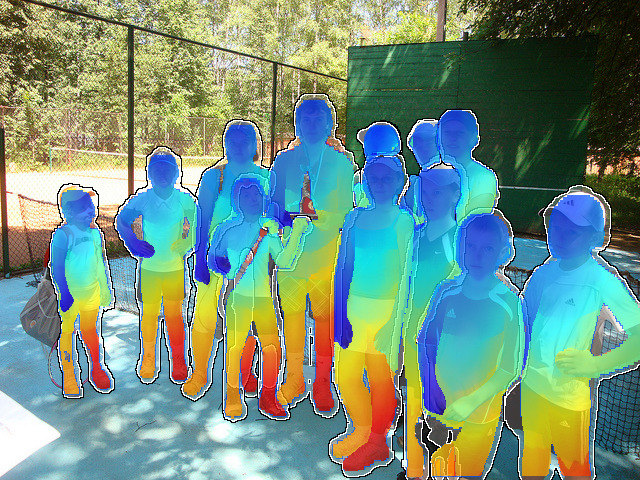}
        \caption{Detections}
    \end{subfigure}
    \begin{subfigure}[t]{0.3333\textwidth}
        \includegraphics[width=\textwidth,trim={0cm 0cm 0cm 3cm}, clip]{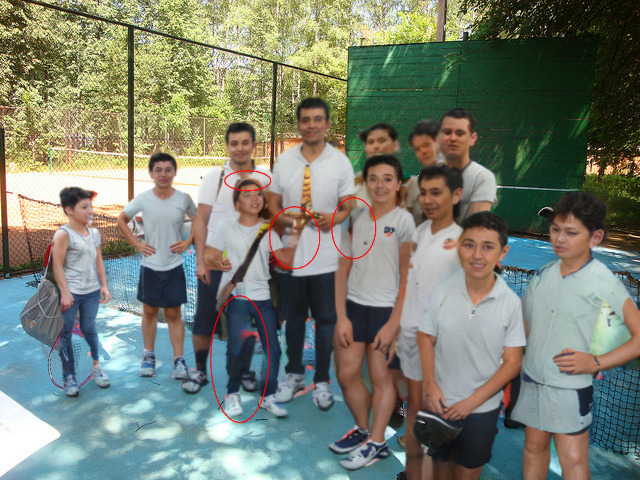}
        \caption{Descending ordering}
    \end{subfigure}
    \begin{subfigure}[t]{0.3333\textwidth}
        \includegraphics[width=\textwidth,trim={0cm 0cm 0cm 3cm}, clip]{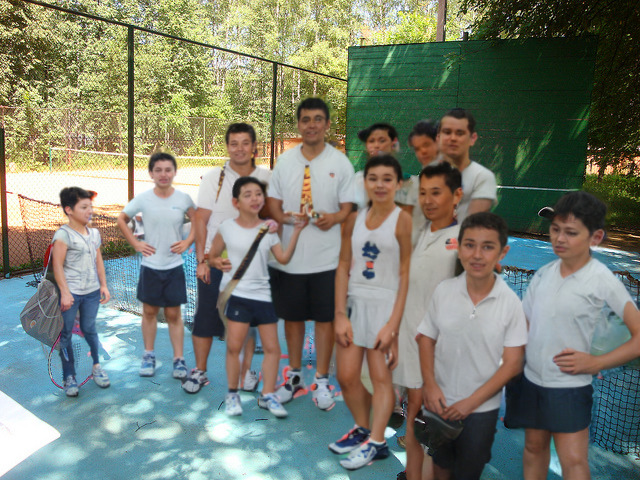}
        \caption{Ascending ordering}
    \end{subfigure}
    \caption{
        Anonymization results comparing our method with descending and ascending image stitching order.
        The ascending ordering stitches foreground objects last, which improves image quality at detection borders (\eg marked in red).
        }
    \label{fig:stitching}
\end{figure*}

\textbf{Implementation Details.}
Instance segmentation and CSE segmentation are combined via simple Intersection over Union (IoU) thresholding, where we assume all detections with an IoU higher than $0.4$ are the same individual.
All instance segmentations from Mask R-CNN not combined with a CSE detection are categorized as a detection without CSE.
All face detections within the CSE or Mask R-CNN segmentation are discarded.
All detections are tracked with simple Kalman filtering on bounding boxes, following the implementation of motpy \cite{motpy}.
We use Mask R-CNN and the CSE implementations from detectron2 \cite{wu2019detectron2},
specifically, the \href{https://github.com/facebookresearch/detectron2/blob/3cc9908ba3301785ec946ae7e37d7091fa1d5045/configs/COCO-InstanceSegmentation/mask_rcnn_X_101_32x8d_FPN_3x.yaml}{ResNeXt-101 FPN} \cite{Xie2017resnext} Mask R-CNN, and
\href{https://github.com/facebookresearch/detectron2/blob/3cc9908ba3301785ec946ae7e37d7091fa1d5045/projects/DensePose/configs/cse/densepose_rcnn_R_101_FPN_DL_soft_s1x.yaml}{ResNet-101} \cite{He2016resnet} CSE.
We adapt DSFD \cite{Li2018c} from the official implementation of the authors.

\subsection{Synthesis Method}
\label{sec:synthesis_method}
DeepPrivacy2 uses three independently trained generators for the three different detection categories introduced in \cref{sec:detection} (detection w/ cse, detection w/o cse,  faces).
While the tasks significantly differ in complexity, they share training setup and architecture to a high degree.
Here, we first present our style-based generator, then present task-specific modeling choices for full-body and face synthesis.
Each generator frame the anonymization task as an image inpainting task, where we remove areas to be anonymized and let a generator complete the missing region.
Specifically, the input and output of each generator is given by
\footnote{For the CSE-guided generator, the CSE-embedding is concatenated with $I \odot M$ to the input of the generator. See \cref{sec:generator}.},
\begin{equation}
    \tilde{I} = G(I \odot M, M, z) \odot M + (1 - M) \odot I,
\end{equation}
$\odot$ is element-wise multiplication, $I$ the original image, and $M$ indicates missing regions ($M$ is 1 for known pixels and 0 for pixels to be anonymized).

\subsubsection{A Style-Based U-Net Generator}
\label{sec:generator}
Our synthesis method follows the implementation of Surface-guided GANs \cite{Hukkelas2022}.
The generator is a U-Net \cite{Ronneberger2015} with limited task-specific modeling choices, consisting of a context encoder and a style-based decoder.
The context encoder uses a sequence of convolutions and downsampling layers, with residual connections at every feature map resolution.
We use no normalization layers in the encoder, as it performs similarly without it.
However, we find it essential to apply instance normalization for the features in the U-net skip connections, where we combine features from the encoder and decoder as additive residuals.
The decoder follows the design of Stylegan2 \cite{Karras2019b}, with the operation order $\textit{instance normalization} \rightarrow \textit{convolution} \rightarrow \textit{style modulation}$.
Note that we replace the baked-in weight demodulation in Stylegan2 with instance normalization, as we find that normalization on expected statistics works poorly when large areas of the input are missing.

Furthermore, we increase the depth of the U-net to 5 downsampling layers, such that the minimum feature resolution is $9 \times 5$.
In contrast, SG-GAN has three downsampling blocks, where most parameters are at the $36 \times 20$ resolution.
We observe no performance degradation by increasing the depth while improving inference speed, as more parameters are located at lower resolution layers.
Finally, we remove V-SAM and discriminator surface supervision used in SG-GAN.
\appDetails includes further details.

\paragraph{Full-Body Synthesis}
The full-body generator is trained on the FDH dataset at a resolution of $288 \times 160$.
We train two independent generators for full-body synthesis; one that concatenates the CSE embedding to the input image and one that does not.
The CSE embedding has a resolution of $16 \times 288 \times 160$, where we use the pixel-to-vertex embedding map released in the official implementation of CSE \cite{Neverova2020,wu2019detectron2}.

\paragraph{Face Synthesis}
In contrast to previous face anonymization methods \cite{Hukkelas2019,Maximov2020, Sun2018}, we propose a generator that does not use keypoints for synthesis.
Removing the keypoint detector improves detection recall in cases where keypoints are difficult to detect.
We train the face generator on an updated version of the FDF dataset \cite{Hukkelas2019}, which increases the image resolution to $256 \times 256$ from the original $128 \times 128$.

\subsection{Recursive Stitching}
The final stage of our pipeline is pasting the anonymized identities into the original image.
Unlike face anonymization, full-body anonymization has many detection overlaps.
If not handled correctly, these overlaps generates visually annoying artifacts at the border between individuals.

Our stitching approach recursively stitches each individual in ascending order depending on the number of pixels the person covers.
The recursive stitching assumes that the synthesis method handles overlapping artifacts when generating each individual.
Additionally, our ordering assumes that objects in the foreground cover a larger area, where foreground objects are stitched in last.
The reverse order (foreground objects first) results in background objects "overwriting" foreground objects, as the detections can overlap (see \cref{fig:stitching}).
This naive ordering significantly reduces visual artifacts at borders between individuals.

\section{Experimental Evaluation}
\begin{figure}[t]
\centering
\begin{subfigure}[t]{0.1\textwidth}
\centering
\includegraphics[width=\textwidth]{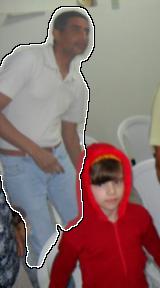}
\includegraphics[width=\textwidth]{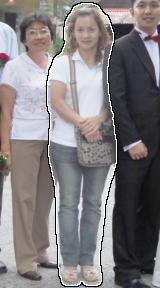}
\includegraphics[width=\textwidth,trim={0cm 0cm 0cm 3.3cm},clip]{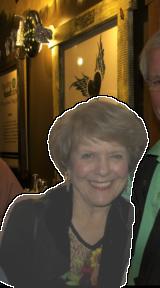}
\includegraphics[width=\textwidth,trim={0cm .8cm 0cm 1.4cm},clip]{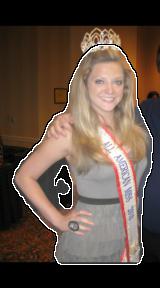}
\includegraphics[width=\textwidth,trim={0cm 0cm 0cm .3cm},clip]{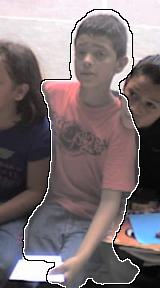}
\caption{Detected}
\end{subfigure}
\begin{subfigure}[t]{0.1\textwidth}
\centering
\includegraphics[width=\textwidth]{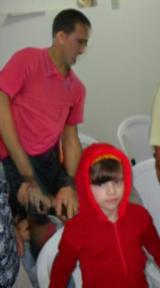}
\includegraphics[width=\textwidth]{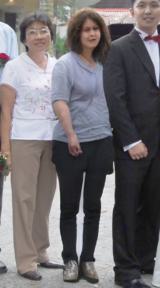}
\includegraphics[width=\textwidth,trim={0cm 0cm 0cm 3.3cm},clip]{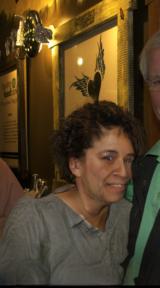}
\includegraphics[width=\textwidth,trim={0cm .8cm 0cm 1.4cm},clip]{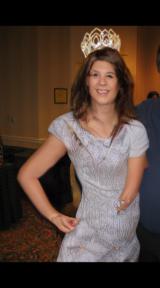}
\includegraphics[width=\textwidth,trim={0cm 0cm 0cm .3cm},clip]{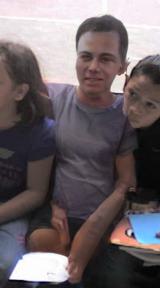}
\caption{No CSE}
\end{subfigure}
\begin{subfigure}[t]{0.1\textwidth}
\centering
\includegraphics[width=\textwidth]{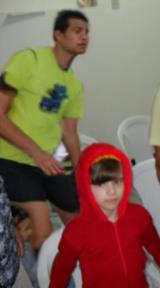}
\includegraphics[width=\textwidth]{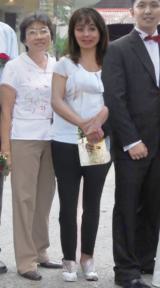}
\includegraphics[width=\textwidth,trim={0cm 0cm 0cm 3.3cm},clip]{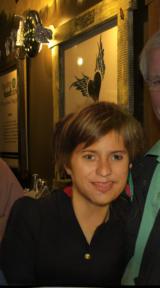}
\includegraphics[width=\textwidth,trim={0cm .8cm 0cm 1.4cm},clip]{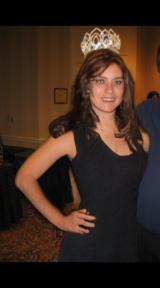}
\includegraphics[width=\textwidth,trim={0cm 0cm 0cm .3cm},clip]{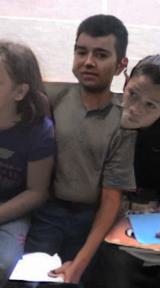}
\caption{}
\end{subfigure}
\begin{subfigure}[t]{0.1\textwidth}
\centering
\includegraphics[width=\textwidth]{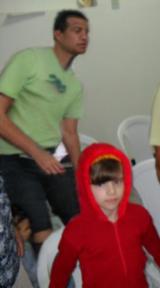}
\includegraphics[width=\textwidth]{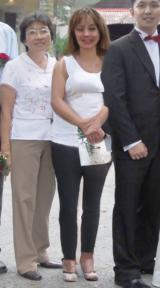}
\includegraphics[width=\textwidth,trim={0cm 0cm 0cm 3.3cm},clip]{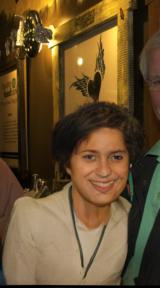}
\includegraphics[width=\textwidth,trim={0cm .8cm 0cm 1.4cm},clip]{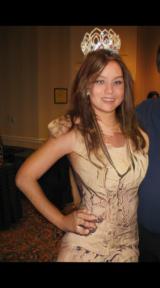}
\includegraphics[width=\textwidth,trim={0cm 0cm 0cm .3cm},clip]{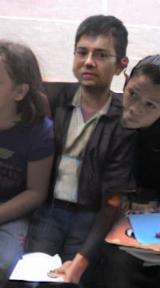}
\caption{}
\end{subfigure}
\begin{subfigure}[t]{0.1\textwidth}
\centering
\includegraphics[width=\textwidth]{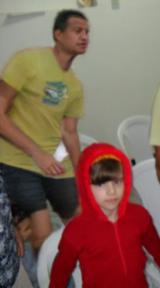}
\includegraphics[width=\textwidth]{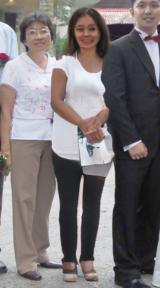}
\includegraphics[width=\textwidth,trim={0cm 0cm 0cm 3.3cm},clip]{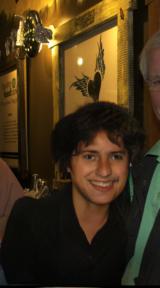}
\includegraphics[width=\textwidth,trim={0cm .8cm 0cm 1.4cm},clip]{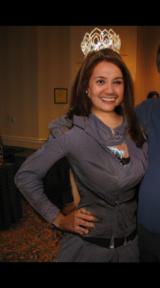}
\includegraphics[width=\textwidth,trim={0cm 0cm 0cm .3cm},clip]{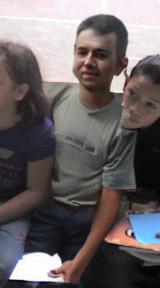}
\caption{}
\end{subfigure}
\caption{Synthesis results on FDH. (a) is the original identity and the anonymization mask, (b) is the unconditional generator, and (c-e) is the CSE-guided generator}
\label{fig:fdh_examples}
\end{figure}
\begin{figure}[t]
\centering
\begin{subfigure}[t]{0.1\textwidth}
\centering
\includegraphics[width=\textwidth]{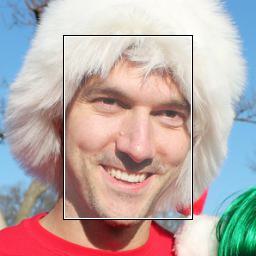}
\includegraphics[width=\textwidth]{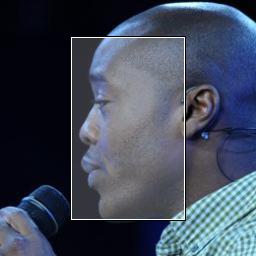}
\includegraphics[width=\textwidth]{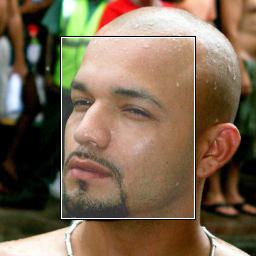}
\includegraphics[width=\textwidth]{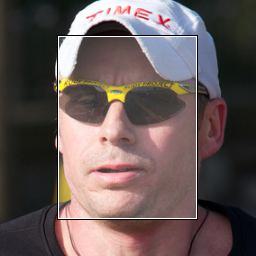}
\includegraphics[width=\textwidth]{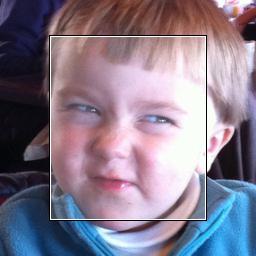}
\end{subfigure}
\begin{subfigure}[t]{0.1\textwidth}
\centering
\includegraphics[width=\textwidth]{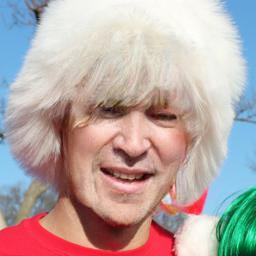}
\includegraphics[width=\textwidth]{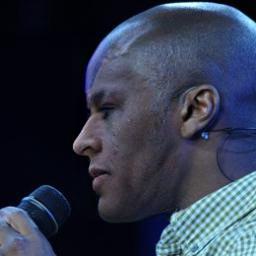}
\includegraphics[width=\textwidth]{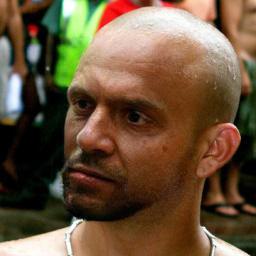}
\includegraphics[width=\textwidth]{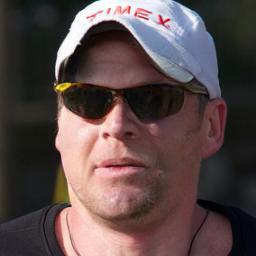}
\includegraphics[width=\textwidth]{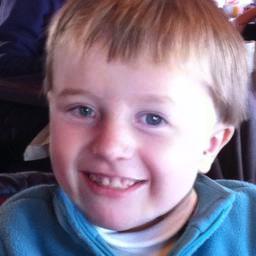}
\end{subfigure}
\begin{subfigure}[t]{0.1\textwidth}
\centering
\includegraphics[width=\textwidth]{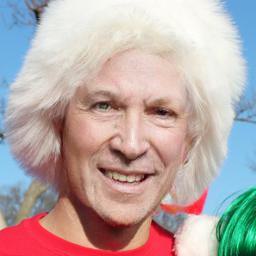}
\includegraphics[width=\textwidth]{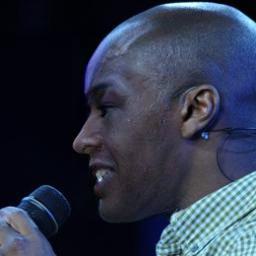}
\includegraphics[width=\textwidth]{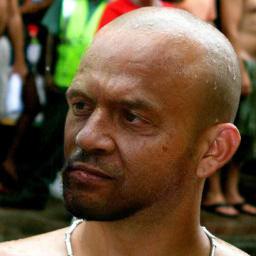}
\includegraphics[width=\textwidth]{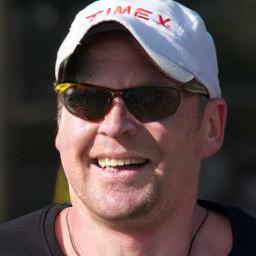}
\includegraphics[width=\textwidth]{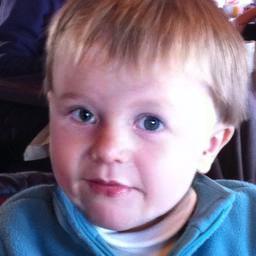}
\end{subfigure}
\begin{subfigure}[t]{0.1\textwidth}
\centering
\includegraphics[width=\textwidth]{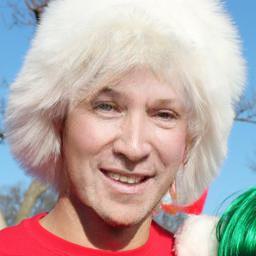}
\includegraphics[width=\textwidth]{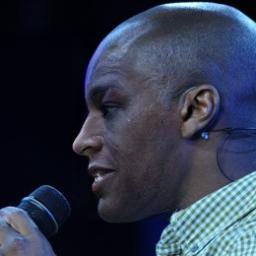}
\includegraphics[width=\textwidth]{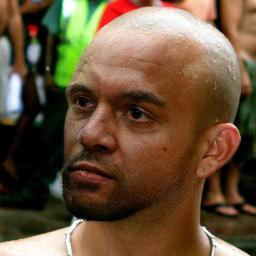}
\includegraphics[width=\textwidth]{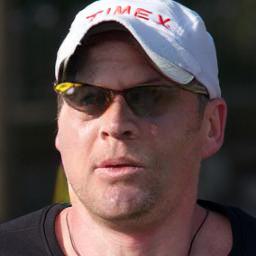}
\includegraphics[width=\textwidth]{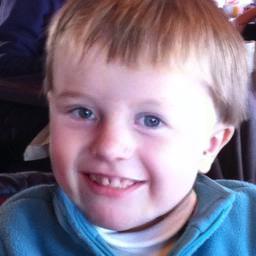}
\end{subfigure}
\begin{subfigure}[t]{0.1\textwidth}
\centering
\includegraphics[width=\textwidth]{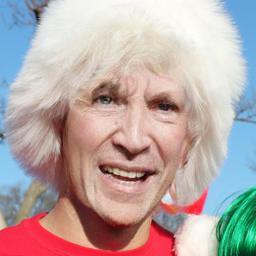}
\includegraphics[width=\textwidth]{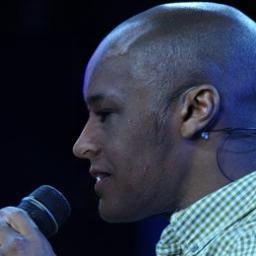}
\includegraphics[width=\textwidth]{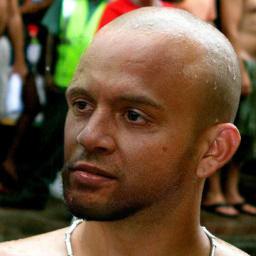}
\includegraphics[width=\textwidth]{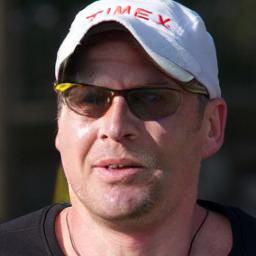}
\includegraphics[width=\textwidth]{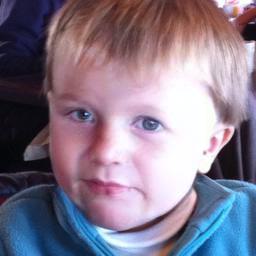}
\end{subfigure}
\caption{
    Synthesis results on FDF256. First column shows the original identity and the anonymization mask.
    Columns 2-5 shows generated identities from DeepPrivacy2.}
\label{fig:cherry_fdf}
\end{figure}
\begin{figure*}
    \begin{subfigure}[t]{0.33\textwidth}
        \includegraphics[width=\textwidth]{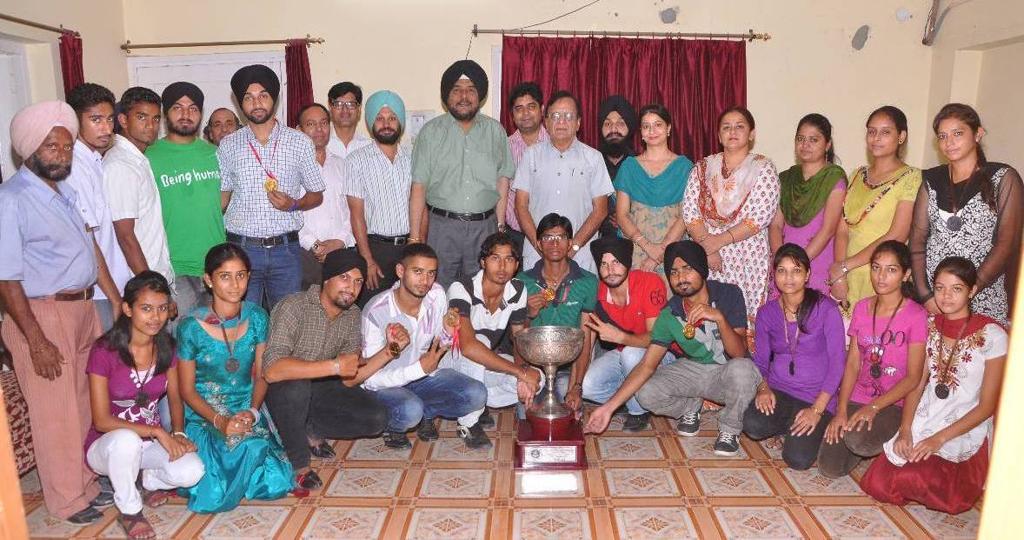}
        \caption{Original}
    \end{subfigure}%
    \begin{subfigure}[t]{0.33\textwidth}
        \includegraphics[width=\textwidth]{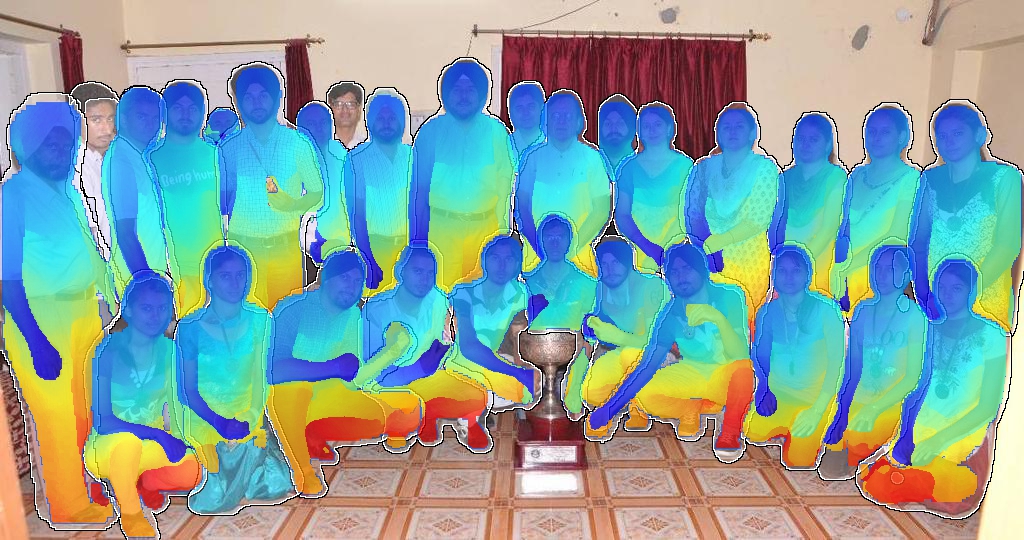}
        \caption{Detections}
    \end{subfigure}%
    \begin{subfigure}[t]{0.33\textwidth}
        \includegraphics[width=\textwidth]{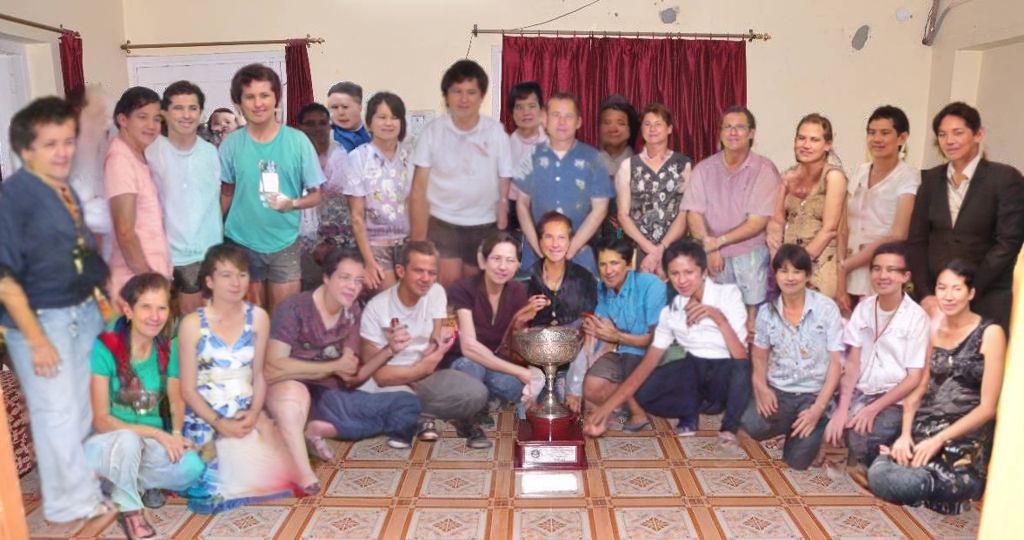}        
        \caption{Anonymized}
    \end{subfigure}
    \caption{DeepPrivacy2 anonymization result on an image from WIDER-Face. \appRandom includes random examples.}
    \label{fig:wider_anonymization}
\end{figure*}
\begin{figure*}
    \begin{subfigure}[t]{0.33\textwidth}
        \includegraphics[width=\textwidth]{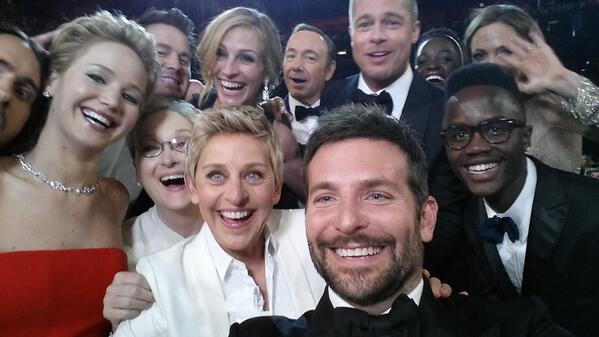}
        \caption{Original}
    \end{subfigure}
    \begin{subfigure}[t]{0.33\textwidth}
        \includegraphics[width=\textwidth]{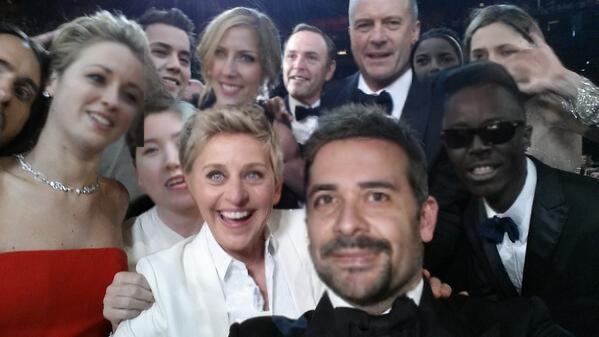}
        \caption{DeepPrivacy}
    \end{subfigure}
    \begin{subfigure}[t]{0.33\textwidth}
        \includegraphics[width=\textwidth]{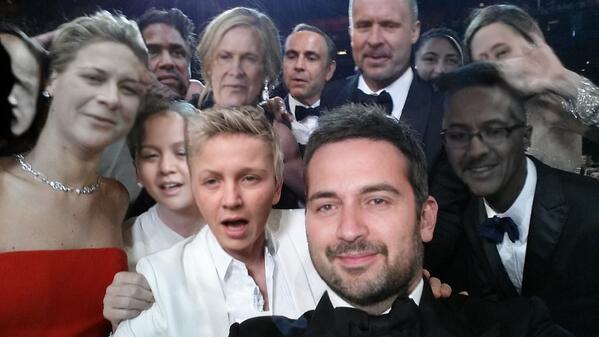}
        \caption{Ours}
    \end{subfigure}
    \caption{
        Face anonymization comparison between our method and DeepPrivacy \cite{Hukkelas2019}.}
    \label{fig:face_comparison}
\end{figure*}

We validate our proposed anonymization pipeline in terms of synthesis quality, using anonymized data for future development, and anonymization guarantees.
There are no standard baselines to compare against for realistic anonymization of data.
Thus, we compare against traditional anonymization techniques, and DeepPrivacy \cite{Hukkelas2019}, a widely adopted realistic face anonymizer.
Additionally, we compare our full-body generator to Surface Guided GANs \cite{Hukkelas2022}.
\appRandom includes random anonymized images on Cityscapes \cite{cityscapes}, COCO \cite{Lin2014COCO}, FDH, and FDF256.

\paragraph{Experimental Details.}
All models are trained with Pytorch 1.10 \cite{paszke2019pytorch} on 4 NVIDIA V100-32GB.
For qualitative examples, we use multi-modal truncation to improve image quality while retaining diversity \cite{Mokady2022}.
We report Fr\'echet Inception Distance (FID) \cite{Heusel2017FID} and FID$_{\text{CLIP}}$
\footnote{FID$_{\text{CLIP}}$ is less sensitive to ImageNet classes.
ImageNet-FID is insensitive to faces and scores images containing ImageNet objects (\eg tie) higher \cite{kynkaanniemi2022role}.} \cite{kynkaanniemi2022role} to evaluate image quality using Torch Fidelity \cite{obukhov2020torchfidelity}.
The three generators (for CSE-guided, unconditional and face) has 43M parameters each.

\paragraph{Datasets.}
For training, we use the FDH dataset (see \cref{sec:fdh_dataset}) for full-body synthesis, and FDF256 for face synthesis.
The FDF256 dataset is an updated version of FDF \cite{Hukkelas2019}, where the image resolution is increased from $128 \times 128$ to $256 \times 256$ (see \appFDF). 
For evaluation, we use Market1501 \cite{Zheng2015}, Cityscapes \cite{cityscapes}, and COCO \cite{Lin2014COCO}.
We follow the standard train/validation split for all datasets.

\paragraph{Runtime Analysis.}
The DeepPrivacy2 architecture is computationally efficient, where the CSE-guided generator processes $\sim 11.6$ frames per second (FPS), and the face generator at $\sim 7.9$ FPS on an NVIDIA 1080 8GB GPU.
In contrast, the SG-GAN \cite{Hukkelas2022} generator processes  $\sim 7.3$ FPS, where our improved runtime originate from the removal of V-SAM and moving the majority of parameters to lower resolution layers.
The entire pipeline (detection, synthesis and stitching) require $\sim 2.8$ seconds to process an image with 12 persons on an NVIDIA 1080 8GB GPU.

\subsection{Synthesis Quality}

\begin{figure}
    \includegraphics[width=0.5\textwidth, trim={0cm 0cm 0 1.5cm},clip]{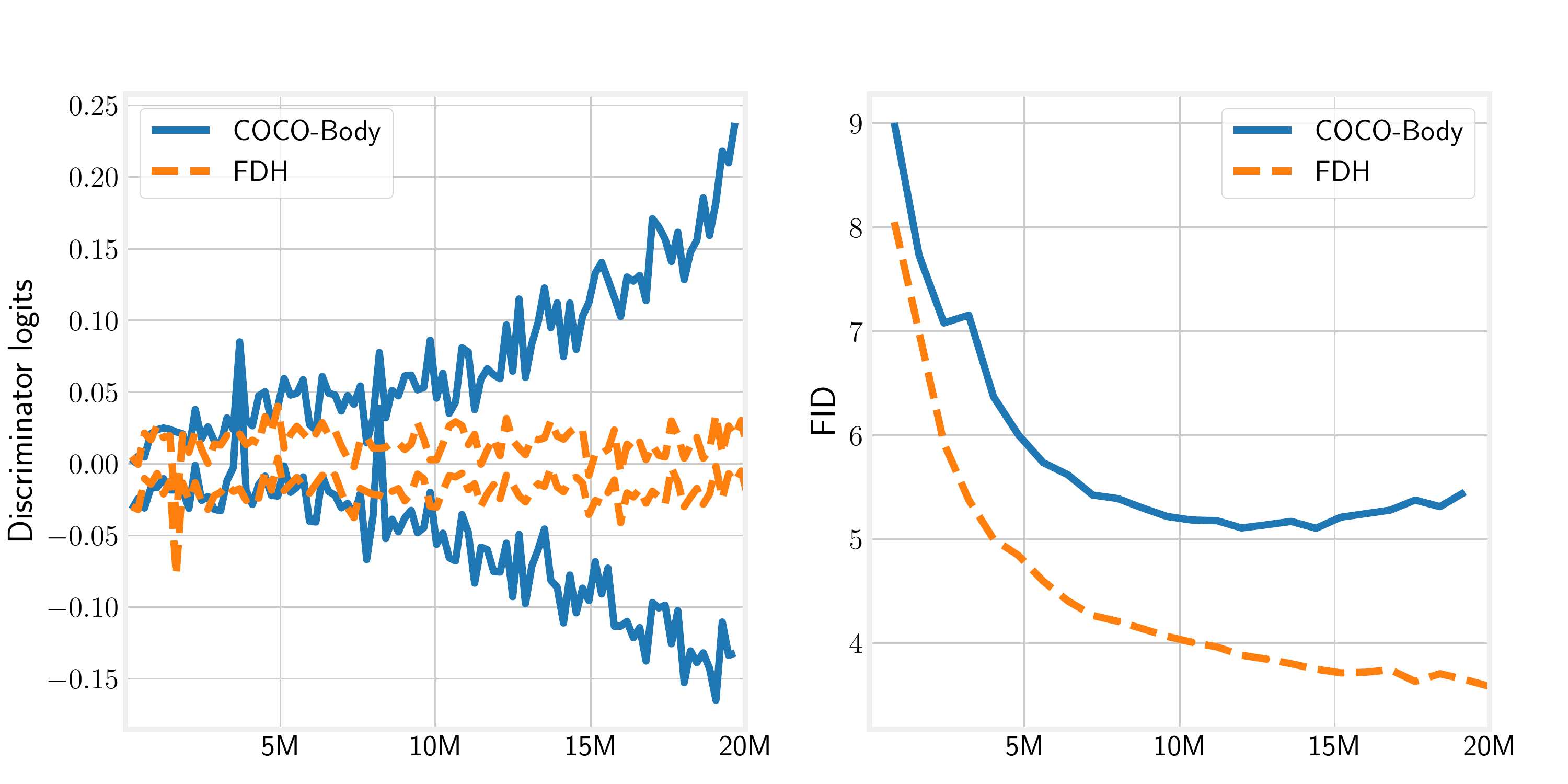}
    \caption{
        Our generator overfits early when trained on the small COCO-Body dataset (\textcolor{blue}{blue}) \cite{Hukkelas2022}, which consists of $\sim 40$K images of human figures.
        Note that this occurs even with the strong data augmentation used by \cite{Hukkelas2022}. No data augmentation, other than horizontal flip, is used for FDH.}
    \label{fig:diverging_D}
\end{figure}

\paragraph{Full-Body Synthesis.}
\Cref{fig:fdh_examples} shows diverse synthesized examples on the FDH dataset.
Our model generates high-quality human figures that seamlessly transition into the original image.
Furthermore, the model can handle a large variety of background contexts, poses, and overlapping objects.
We find CSE guidance necessary for high-quality anonymization, where the unconditional generator often generates human figures with unnatural poses (see \cref{fig:fdh_examples}).
This is reflected in quantitative metrics, where the CSE-guided generator (FID=5.6, FID$_{\text{CLIP}}$=1.7) substantially improves over the unconditional generator (FID=6.1, FID$_{\text{CLIP}}$=2.30).
Furthermore, the primary improvement of our model compared to Surface-guided GANs \cite{Hukkelas2022} is the larger and more diverse FDH dataset.
The same model trained on the COCO-Body dataset \cite{Hukkelas2022} starts to overfit early in training, reflected by the diverging discriminator logits and increasing FID (\cref{fig:diverging_D}).

\paragraph{Face Synthesis.}

\Cref{fig:cherry_fdf} shows generated results on the FDF256 dataset.
Directly comparing our face anonymizer to DeepPrivacy \cite{Hukkelas2019} is not straightforward, as we synthesize higher resolution images ($256 \times 256$, not $128 \times 128$).
Additionally, the FDF256 dataset does not include the same images as the original dataset, as FDF256 filter out  lower resolution images.
Nevertheless, to validate our modeling choices, we retrain our GAN for $128 \times 128$ synthesis on FDF \cite{Hukkelas2019}.
Our GAN achieves a FID of 0.56, a significant improvement compared to DeepPrivacy (FID=0.68) \cite{Hukkelas2019}.
Note that this is without using face keypoints, which the original DeepPrivacy uses to improve quality.

\Cref{fig:face_comparison} compares the open-source DeepPrivacy \cite{Hukkelas2019} to our method.
Our method generates higher quality faces and handles overlaps between detections better.
Also, note that DeepPrivacy does not anonymize all faces in the image, as it is unable to detect keypoints for all individuals 
\footnote{Even with confidence threshold of 0.05, DeepPrivacy is unable to detect keypoints for all individuals.}.

\paragraph{Attribute-Guided Anonymization}
\begin{figure}
    \begin{subfigure}{0.25\textwidth}
        \includegraphics[width=\textwidth,trim={5cm 10cm 15cm 1cm},clip]{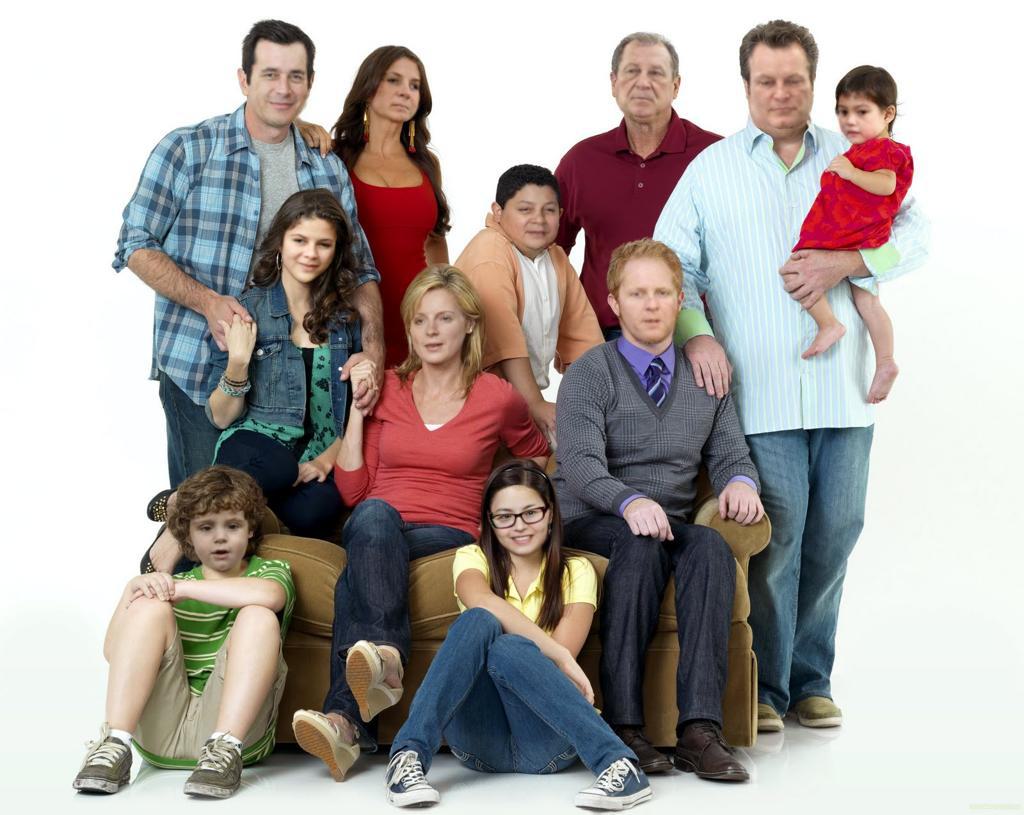}
        \caption*{Original Anonymization}
    \end{subfigure}%
    \begin{subfigure}{0.25\textwidth}
        \includegraphics[width=\textwidth,trim={5cm 10cm 15cm 1cm},clip]{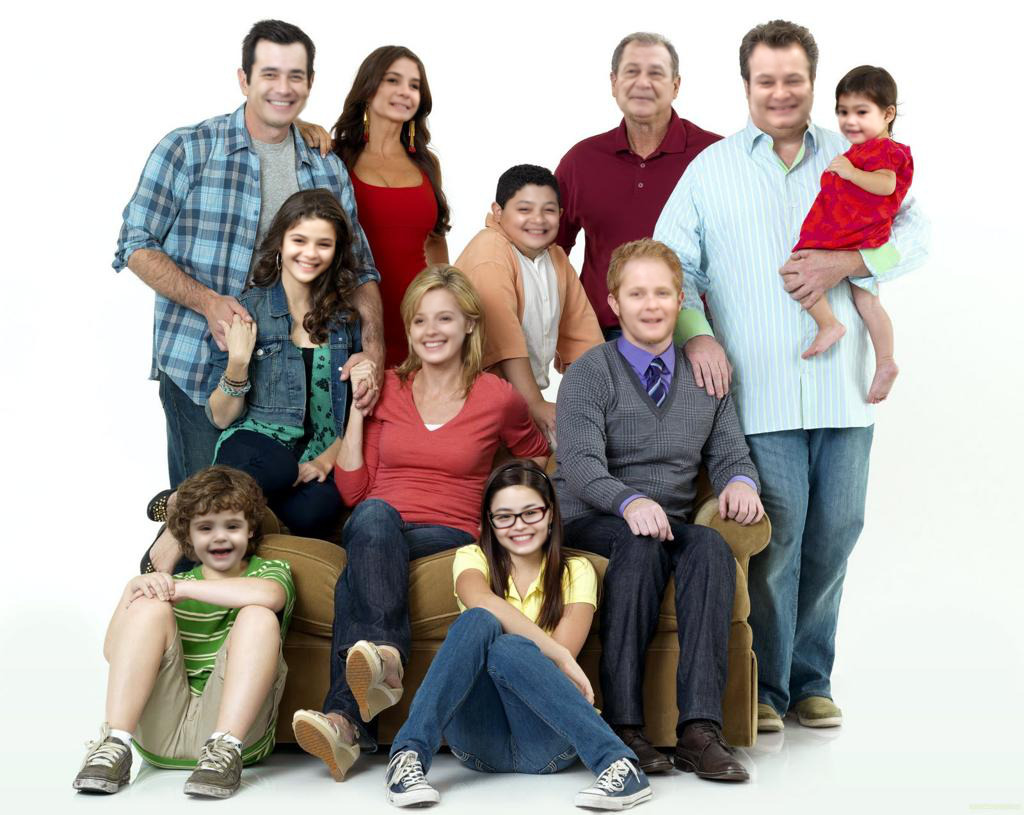}
        \caption*{Happy}
    \end{subfigure}
    \begin{subfigure}{0.25\textwidth}
        \includegraphics[width=\textwidth,trim={5cm 10cm 15cm 1cm},clip]{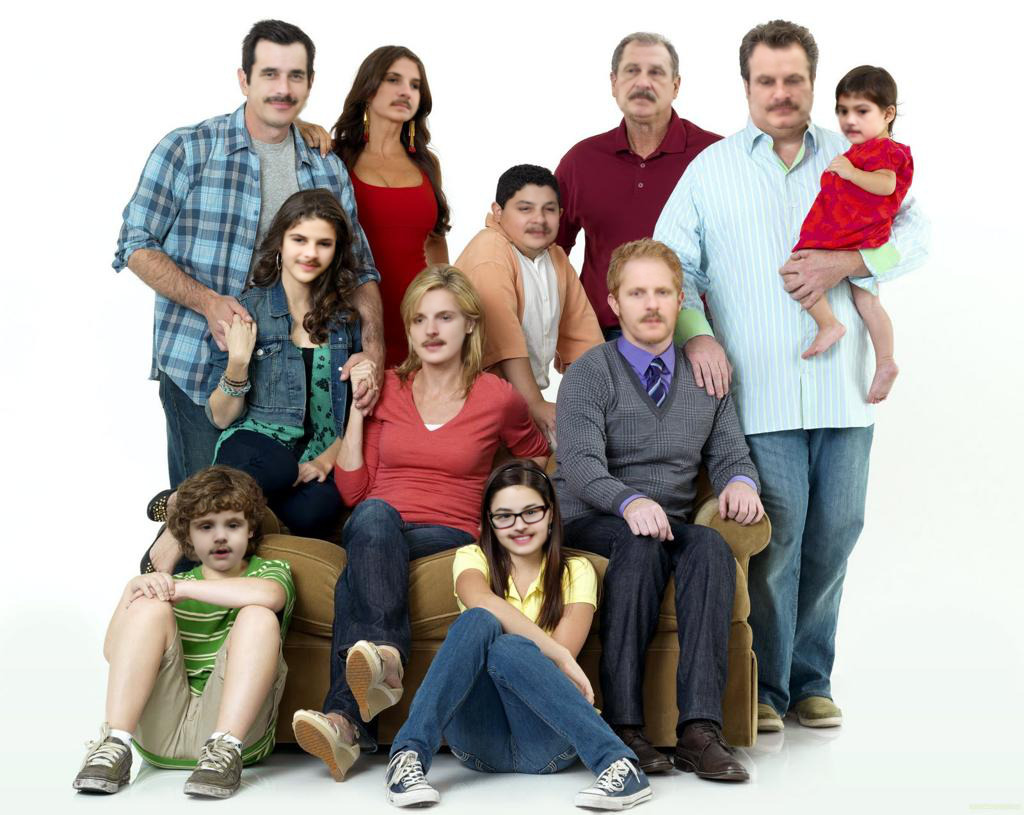}
        \caption*{Mustache}
    \end{subfigure}%
    \begin{subfigure}{0.25\textwidth}
        \includegraphics[width=\textwidth,trim={5cm 10cm 15cm 1cm},clip]{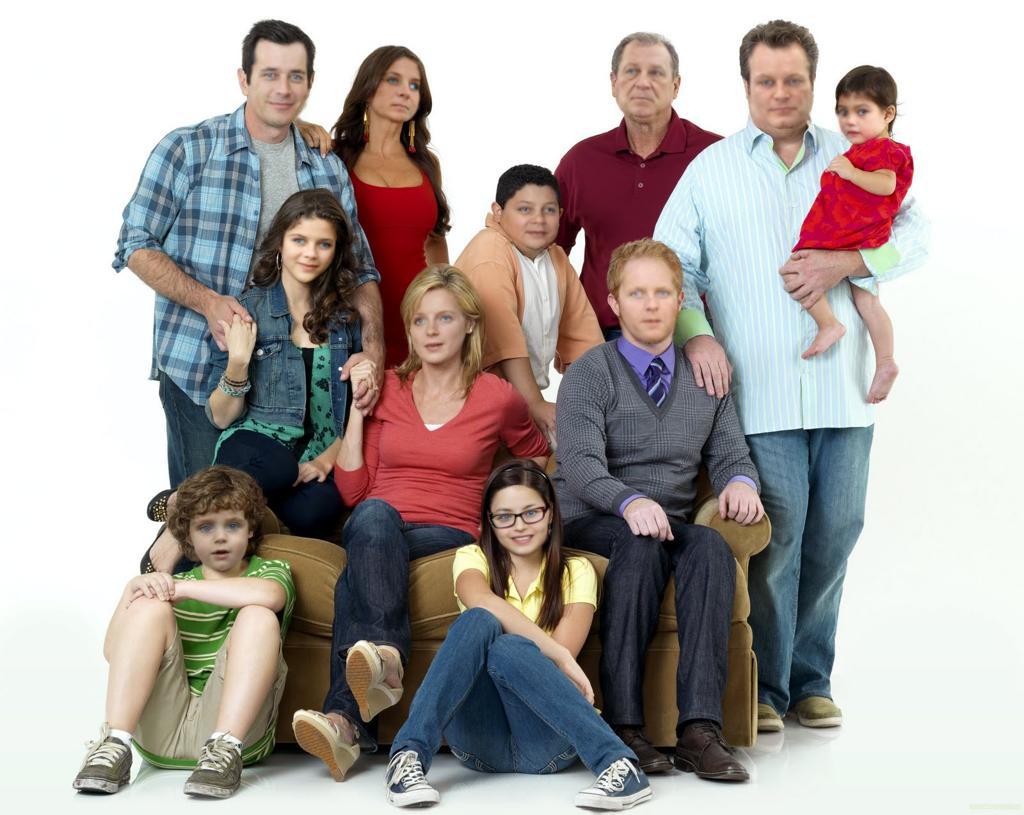}
        \caption*{Blue eyes}
    \end{subfigure}%
    \caption{Latent manipulations with StyleMC \cite{Kocasari2022stylemc}. The text prompt used for each edit is shown below each image. See our video for an interactive demo: \href{https://youtu.be/faoNyaaORts}{https://youtu.be/faoNyaaORts}.}
    \label{fig:stylemc}
\end{figure}
DeepPrivacy2 allows for controllable anonymization through text prompts by adapting StyleMC \cite{Kocasari2022stylemc}.
StyleMC finds global semantically meaningful directions in the GAN latent space by manipulating images towards a given text prompt with a CLIP-based \cite{Radford2021CLIP} loss.
\Cref{fig:stylemc} shows attribute-guided anonymization, where the global directions are found over 256 images.
As far as we know, DeepPrivacy2 is the first to enable controllable anonymization through text prompts, whereas previous methods are limited to no control or attribute preservation from the original identity \cite{Gafni2019} .

\subsection{Anonymization Evaluation}

\paragraph{Anonymization Guarantee}
\begin{table}
    \centering 
    \begin{tabular}{|l|c|c|}
    \hline
        Anonymization & R1 $\downarrow$ & mAP $\downarrow$\\ 
        \hline 
        Original & 94.4 & 82.5 \\ 
        \hline
        Pixelation $8 \times 8$ & 54.6 & 16.1   \\
        Pixelation $16 \times 16$ & 70.3 & 36.6   \\
        Mask-out   & 45.5 & \textbf{8.0}  \\
        \hline
        SG-GAN \cite{Hukkelas2022} & 74.4 & 30.2 \\
        Full-body anonymization (Ours) & \textbf{44.7} & 8.5 \\
        \hline
    \end{tabular}
    \caption{
        Re-identification mAP and rank-1 accuracy on Market1501 \cite{Zheng2015} using OSNet \cite{Zhou2019}.
        }
    \label{tab:re_identification}
\end{table}

To evaluate the anonymization guarantee of DeepPrivacy2, we evaluate how well automatic re-identification tools can identify anonymized individuals.
Specifically, we evaluate the re-identification rate of OSNet \cite{Zhou2019} by anonymizing Market1501 \cite{Zheng2015}.
In this case, a lower re-identification mAP and rank-1 accuracy (R1) reflects worse re-identification, indicating improved anonymization.
\appDetails details the experiment further.

\Cref{tab:re_identification} reflects that pixelation enables re-identification of several of the anonymized individuals.
Our full-body anonymizer yields similar anonymization guarantees as masking out the area and significantly improve compared to pixelation.
Furthermore, the full-body anonymization of SG-GAN \cite{Hukkelas2022} provides poorer anonymization results than DeepPrivacy2.
This is caused by the CSE detector failing in several cases and the poor segmentation of accessories/hair in SG-GAN, where the anonymized identity often "wears" parts of the original identity (see \cref{sec:detection}).

\subsubsection{Training and Evaluating on Anonymized Data}
A typical use case for anonymization is collecting and anonymizing data for the development of computer vision systems.
We evaluate DeepPrivacy2 on two established computer vision benchmarks:
COCO \cite{Lin2014COCO} person keypoint estimation and Cityscapes \cite{cityscapes} instance segmentation.
We evaluate two use cases; (1) using anonymized data for training, and (2) using anonymized data for validation with a pre-trained model.
For the former, we report evaluation metrics on the original validation dataset.

\textbf{COCO Person Keypoint Estimation.}
\begin{table}

    \begin{center}
        \resizebox{\linewidth}{!}{
        \begin{tabular}{|c|c|c||c|c|}
            \cline{2-5}
             \multicolumn{1}{c|}{}& \multicolumn{2}{c||}{Train w/ Anon. Data} & \multicolumn{2}{c|}{Validate w/ Anon. Data} \\ 
            \hline
            Dataset & Box AP $\uparrow$& Kp. AP $\uparrow$& Box AP $\uparrow$& Kp. AP$\uparrow$\\ 
            \hline
            \hline
             Original &  53.6 & 64.0 & 53.6 & 64.0\\ 
             \hline
             Masked Out & 10.1 & 0.5 & 17.0 & 1.8\\ 
             Pixelation $8 \times 8$ & 10.4 & 1.0 & 29.1 & 2.2\\ 
             Pixelation $16 \times 16$ & 10.1 & 1.5 & 36.5 & 12.0 \\
             DeepPrivacy2 (w/o CSE) & 21.4 & 10.2 & \textbf{49.9} & 11.5 \\
             DeepPrivacy2 & \textbf{26.0} & \textbf{31.9} & 49.4 & \textbf{48.4} \\ 
             \hline
        \end{tabular}}
        \end{center}
        \caption{
            Keypoint (Kp.) AP on the COCO \cite{Lin2014COCO} validation set with a Keypoint \href{https://github.com/facebookresearch/detectron2/blob/main/configs/COCO-Keypoints/keypoint_rcnn_R_50_FPN_1x.yaml}{R-50 FPN}  R-CNN  \cite{He2017}.
            }
        \label{tab:keypoint_COCO}
\end{table}
\Cref{tab:keypoint_COCO} analyzes the effect of anonymization on the COCO dataset for person keypoint estimation.
Pixelation greatly affects model training for the fine-grained task of keypoint estimation, whereas DeepPrivacy2 significantly improves over traditional methods.
Note that CSE fails to detect many individuals in the COCO dataset, which yields poor pose preservation for individuals anonymized by the unconditional generator.

\textbf{Cityscapes Instance Segmentation.}
\Cref{tab:cityscapes} analyzes the effect of anonymization on the Cityscapes dataset.
DeepPrivacy2 improves over pixelation and mask-out, but the gap is less prevalent than for keypoint estimation.
We believe this originates from model weight initialization
\footnote{
The Cityscapes model is initialized from a COCO pre-trained Mask R-CNN, while the keypoint R-CNN from an ImageNet \cite{Deng2010ImageNet} backbone.}.

\textbf{Is surface guidance necessary?}
Section 5.1 established that the CSE-guided generator improves image quality compared to the unconditional generator.
We now ask the question; does the improved image quality translate to improvements when using the anonymized data?
In \cref{tab:cityscapes} and \cref{tab:keypoint_COCO}, we replace the CSE-guided generator with the unconditional generator, such that all persons are anonymized without CSE-guidance (denoted \emph{DeepPrivacy2 w/o CSE}).
Removing CSE-guidance severly hurts performance, especially when using the anonymized data for training.

\begin{table}

    \begin{center}
        \resizebox{\linewidth}{!}{
        \begin{tabular}{|c|c|c||c|c|}
            \cline{2-5}
             \multicolumn{1}{c|}{} & \multicolumn{2}{c||}{Train w/ Anon. Data} & \multicolumn{2}{c|}{Validate w/ Anon. Data} \\ 
            \hline
            Dataset& mAP $\uparrow$ & AP$_\text{person}$  $\uparrow$& mAP $\uparrow$& AP$_\text{person}$ $\uparrow$\\ 
            \hline
            \hline
             Original &  36.5 & 35.0 & 36.5 & 35.0\\ 
             \hline
             Masked Out & 34.0 & 26.4  & 27.7 & 4.7 \\ 
             Pixelation $8 \times 8$ & 34.7 & 27.1 & 29.4 & 10.2 \\
             Pixelation $16 \times 16$ & 34.7 & 29.6 & 32.0 & 21.8\\
             DeepPrivacy2 (w/o CSE) & 33.4 & 27.5 & 33.1 & \textbf{27.8} \\
             DeepPrivacy2 &\textbf{35.2} & \textbf{30.3} & \textbf{33.2} & 27.3 \\ 
             \hline
        \end{tabular}}
        \end{center}
        \caption{
            Instance segmentation AP on the Cityscapes \cite{cityscapes} validation set with a Mask R-CNN \cite{He2017} \href{https://github.com/facebookresearch/detectron2/blob/main/configs/Cityscapes/mask_rcnn_R_50_FPN.yaml}{R-50 FPN}.
            }
        \label{tab:cityscapes}
\end{table}

\section{Conclusion}
DeepPrivacy2 is an automatic realistic anonymization framework for human figures and faces, and is a practical tool for anonymization without degrading the image quality.
Compared to previously proposed anonymization frameworks, we show that DeepPrivacy2 substantially improves image quality and privacy guarantees.
Furthermore, we introduce the FDH dataset, a large-scale full-body synthesis dataset that includes a wide variety of identities in different poses and contexts.
Our new FDH dataset, combined with our simple style-based GAN, improves image quality and diversity of human figure synthesis for in-the-wild images.
Furthermore, we show that our simple style-based GAN generates high-quality human faces that are controllable through user-guided anonymization via text prompts.
We believe that our open-source framework will be a useful tool for computer vision researchers and other entities requiring anonymization while retaining image quality.

\paragraph{Societal Impact}
Recently introduced legislation in many regions has complicated collecting privacy-sensitive data, where consent from individuals is required for storing the data.
This can act as a barrier for developing applications relying on high-quality images, such as computer vision models.
This paper proposes an automatic realistic image anonymization framework that simplifies the collection of privacy-sensitive data while retaining the original image quality.
We believe this will be a highly useful tool for the computer vision field.
Nevertheless, our work focuses on synthesizing realistic humans, which has a potential for misuse (\eg DeepFakes).
There is a large focus in the community to mitigate this, for example, the DeepFake Detection Challenge \cite{Dolhansky2020} and model watermarking \cite{Yu2021}.

\subsection{Limitations}
\begin{figure}
    \begin{subfigure}{0.2\linewidth}
        \includegraphics[width=\textwidth]{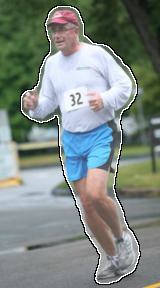}
        \includegraphics[width=\textwidth]{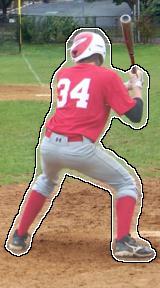}
    \end{subfigure}%
    \begin{subfigure}{0.2\linewidth}
        \includegraphics[width=\textwidth]{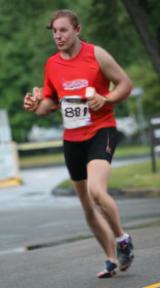}
        \includegraphics[width=\textwidth]{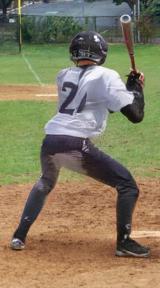}
    \end{subfigure}%
    \begin{subfigure}{0.2\linewidth}
        \includegraphics[width=\textwidth]{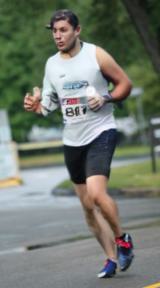}
        \includegraphics[width=\textwidth]{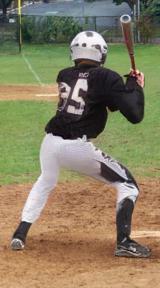}
    \end{subfigure}%
    \begin{subfigure}{0.2\linewidth}
        \includegraphics[width=\textwidth]{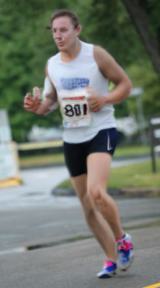}
        \includegraphics[width=\textwidth]{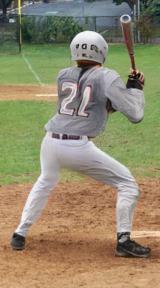}
    \end{subfigure}%
    \begin{subfigure}{0.2\linewidth}
        \includegraphics[width=\textwidth]{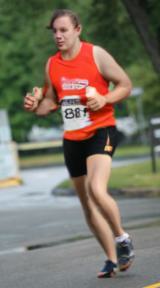}
        \includegraphics[width=\textwidth]{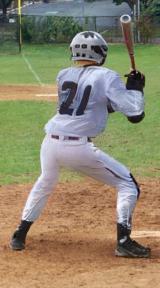}
    \end{subfigure}
    \caption{The generator samples from a small subset of different identities given the condition.}
    \label{fig:conditioning_bias}
\end{figure}

DeepPrivacy2 generates a limited set of identities given a particular input condition.
The input condition is highly descriptive of the shape of the original identity and the context that the identity should fit into.
Thus, the generator learns a sampling probability of identities given the condition. 
For example, if the generator observes a baseball field, the synthesized identity is likely to be a baseball player (\cref{fig:conditioning_bias}).

As with any anonymization framework, DeepPrivacy2 cannot guarantee anonymization without human supervision, as the detector can fail.
However, DeepPrivacy2 uses a set of detectors from different modalities to improve detection in cases where one or more of the detectors fail.
Also, DeepPrivacy2 uses dense pose description for anonymization, which allows identity recognition through gait \cite{Abarca2021handbook}.

\paragraph{Synthesis Quality}
DeepPrivacy2 significantly improves full-body synthesis for in-the-wild images; however, it struggles in several scenarios.
First, DeepPrivacy2 relies on dense pose estimation to synthesize high-quality human figures, where the image quality is severely degraded in cases where the pose description is incorrect.
Furthermore, we find our full-body GAN harder to edit (\eg attribute edit via text prompts \cite{Kocasari2022stylemc}), and we observe that common directions in the latent space do not translate to semantically equivalent transformations for different poses/background contexts.

\paragraph{Acknowledgement}
The computations were performed on resources provided by the NTNU IDUN/EPIC computing cluster \cite{sjalander2019epic} and the Tensor-GPU project led by Prof. Anne C. Elster through support from The Department of Computer Science and The Faculty of Information Technology and Electrical Engineering, NTNU.
\setcounter{section}{0}
\renewcommand*{\thepart}{\Alph{part}}
\renewcommand*{\thesection}{\Alph{section}}
\renewcommand*{\thesubsection}{\Roman{subsection}}
\renewcommand*{\thesubsubsection}{\Roman{subsubsection}}
\renewcommand*{\theparagraph}{\alph{paragraph}}
\renewcommand*{\thesubparagraph}{(\Roman{subparagraph})}
\section{Flickr Diverse Humans Dataset}
\label{sec:fdh_filtering}
The FDH dataset is generated by running the detection pipeline of DeepPrivacy2, where we only keep detections that are not filtered out by the set of criterias mentioned below.
We filter out poor quality images in the FDH dataset through the following criterias:
\begin{itemize}
    \item \textbf{Confidence thresholding:} All detections with confidence with less than 0.98.
    \item \textbf{Low resolution images:} All detections where the resulting image has an area lower than $144 \times 80$.
    \item \textbf{Grayscale images:} All grayscale images.
    \item \textbf{Automatic image quality assessment:} We use an open-source implementation \cite{idealods2018imagequalityassessment} of NIMA \cite{Talebi2018} and filter every detection with a score less than 3.
    \item \textbf{CSE Vertices per part:} 
    We filter out detections with corrupted CSE detections by counting the number of vertices that belong each body part. 
    Specifically, by finding pixel-to-vertex correspondences to the extended SMPL \cite{Loper2015} body model used by CSE \cite{Neverova2020}, we find the number of unqiue vertexes in the image that belong to each body part for the image (split into 26 regions).
    Then, if the average number of unique vertices per body part is less than 135, we filter the detection out.
    \item \textbf{Mask R-CNN and CSE IoU:} Detections where Mask R-CNN and CSE segmentations have an IoU threshold lower than 0.5.
    \item \textbf{Keypoint-to-CSE Correspondence:} We filter detections where the keypoints and CSE annotations do not match.
    Specifically, by using pixel-to-vertex correspondences we segment each CSE embedding into a semantic body part \footnote{We use an open source semantic segmentation of the SMPL model, found \href{https://raw.githubusercontent.com/Meshcapade/wiki/main/assets/SMPL_body_segmentation/smpl/smpl_vert_segmentation.json}{here}.}.
    Then, we count the number of keypoints that match to its corresponding body part (\eg eyes should match to the body part "head").
    We include any detection where at least 8 keypoints matches the corresponding body part.
    The keypoint annotations are predicted from a pre-trained Keypoint R-CNN \cite{He2017} on COCO \cite{Lin2014COCO} from torchvision \cite{paszke2019pytorch}.
\end{itemize}

\section{The Updated Flickr Diverse Faces Dataset}
\label{sec:fdf}
The Flickr Diverse Faces 256 (FDF256) dataset is a derivate from 1.08M images from the YFCC100M \cite{Thomee2015} dataset, following  the dataset generation of the original FDF dataset \cite{Hukkelas2019}.
The training dataset consists of 242,031 images and the validation set of 6533 images, where each face is up/downsampled to $256 \times 256$.
We filter out all faces where the original resolution is smaller than $64 \times 64$.
Each face is annotated with keypoints from a pre-trained keypoint R-CNN \cite{He2017} R50-FPN from torchvision, and the bounding box is from the official implementation of DSFD \cite{Li2018c}.

\section{Experimental Details}
\label{sec:experimental_details}
\paragraph{GAN Implementation Details}
We follow the implementation setup and hyperparameters of SG-GAN \cite{Hukkelas2022} for the optimizer and loss objective, unless stated otherwise.
We train all networks with the Adam optimizer \cite{Kingma2014Adam} with batch size of 32 and learning rate of 0.002.
For both the FDF and FDH dataset, we use no data augmentation except random horizontal flip.
The discriminator use the same architecture as SG-GAN \cite{Hukkelas2022}, except that we remove the layers used for the feature pyramid network.

\paragraph{Market1501 Anonymization.}
\label{sec:market1501}
The market1501 dataset consists of 19,732 test images and 3368 query images.
We anonymize the 19,732 test images and use the original 3368 query images to perform re-identification on the anonymized test images.
We use a confidence threshold of 0.1 for Mask R-CNN, 0.5 for the face detector and 0.3 for the CSE detector.
DSFD \cite{Li2018c} and CSE \cite{Neverova2020} detects a large amount of false positives on the Market1501 dataset, which we believe is caused by the low resolution of the images.
Out of all the detections, 45\% are detected with CSE, 52\% are detected by Mask R-CNN, and the remaining by DSFD.
The detector fails to detect any individual for 10\% of the test images.
The mAP/R1 metric is calculated from all test examples, including cases where our detector fails to detect any individual.

\section{Random Anonymized Examples}
\begin{figure*}[t]
\centering
\begin{subfigure}[t]{0.3333333333333333\textwidth}
\centering
\includegraphics[width=\textwidth]{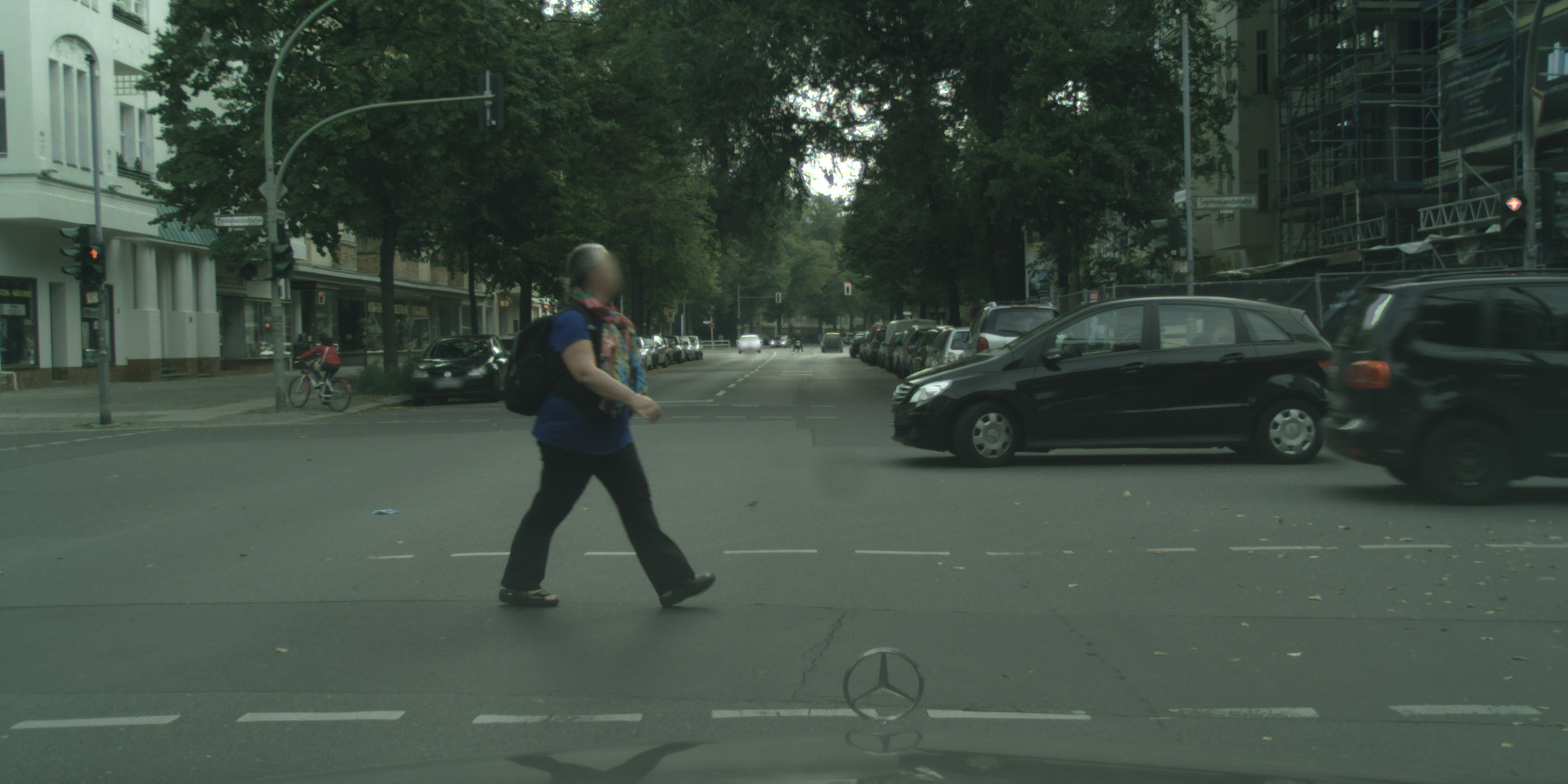}
\includegraphics[width=\textwidth]{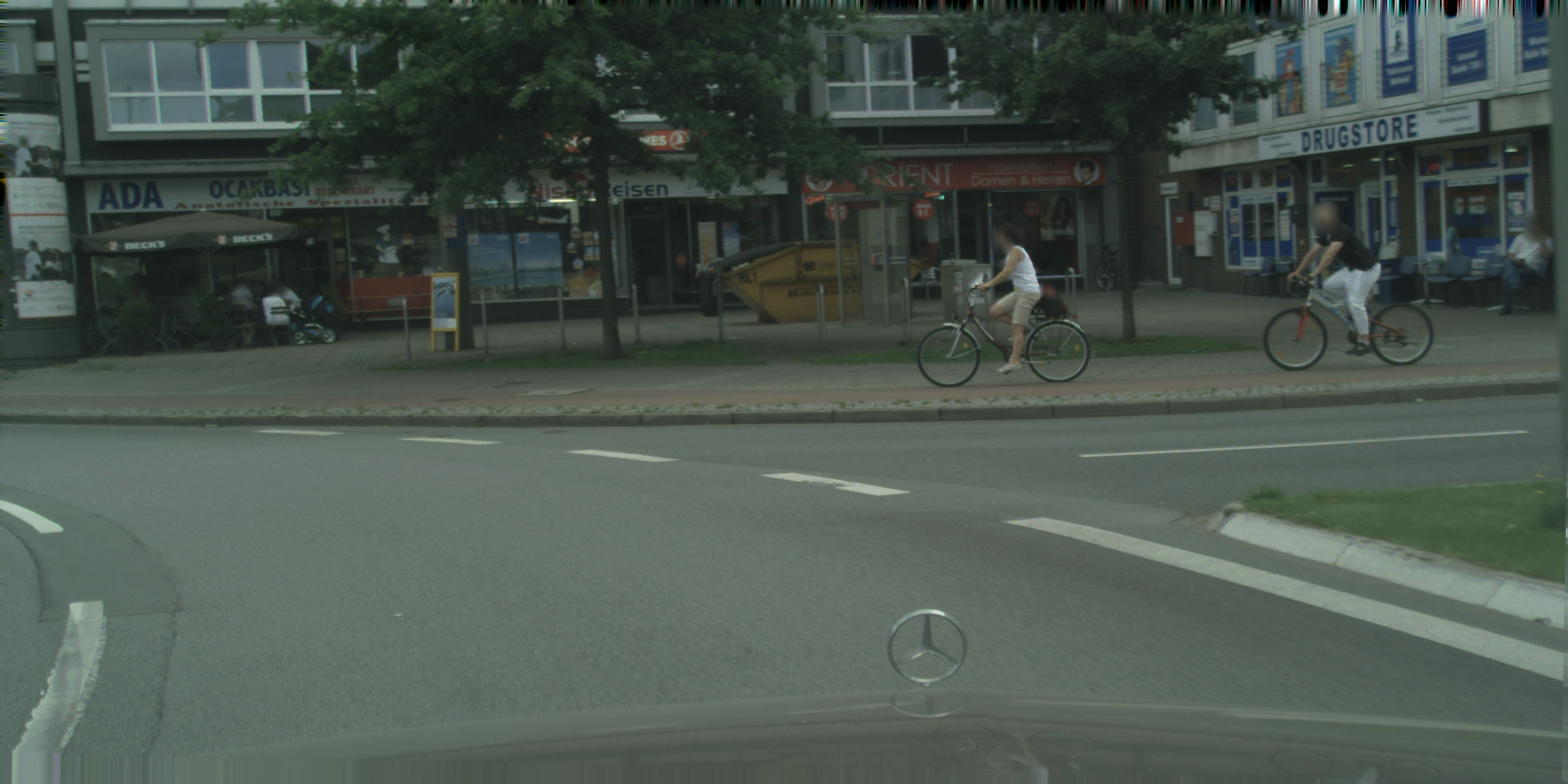}
\includegraphics[width=\textwidth]{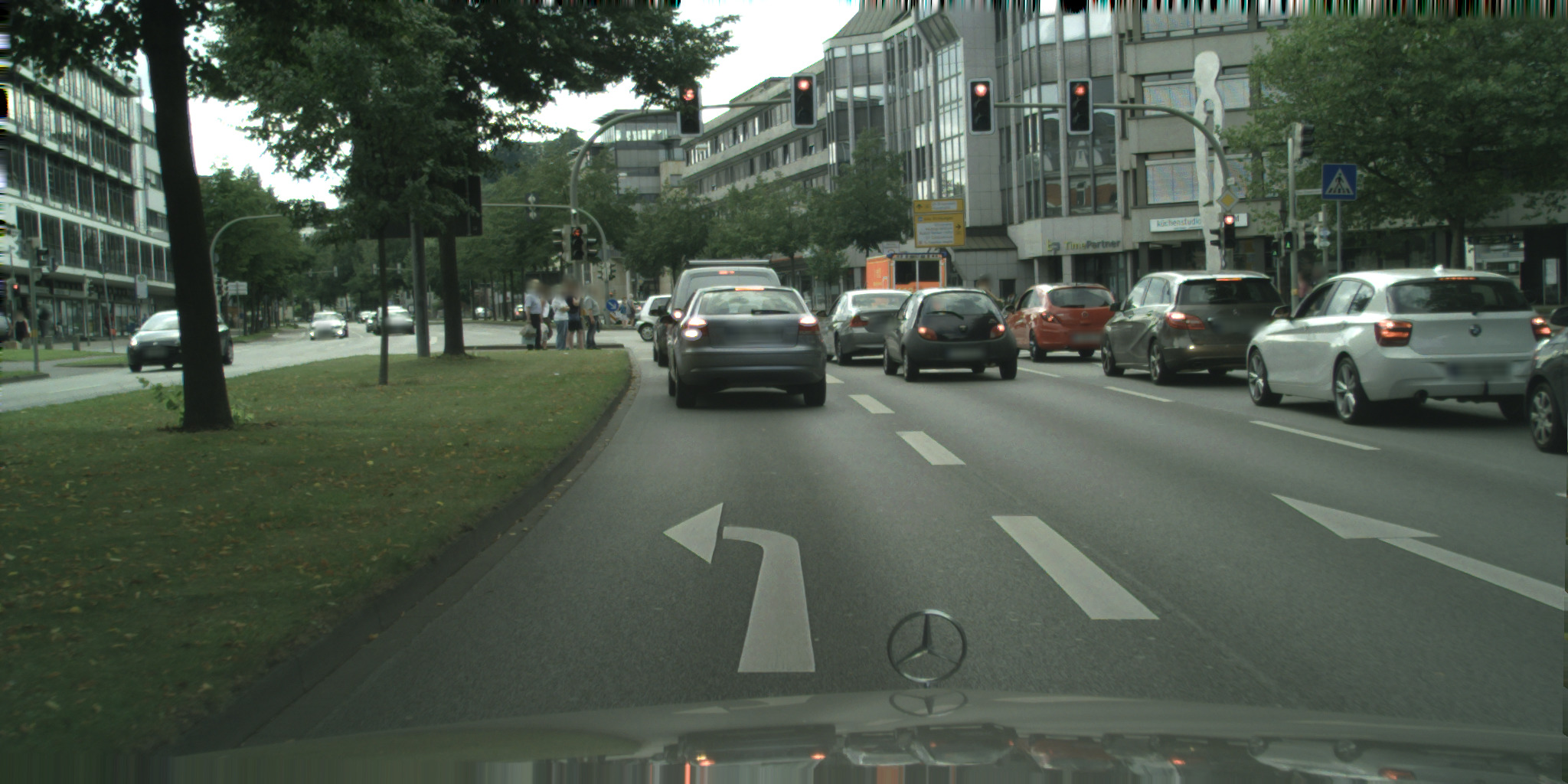}
\includegraphics[width=\textwidth]{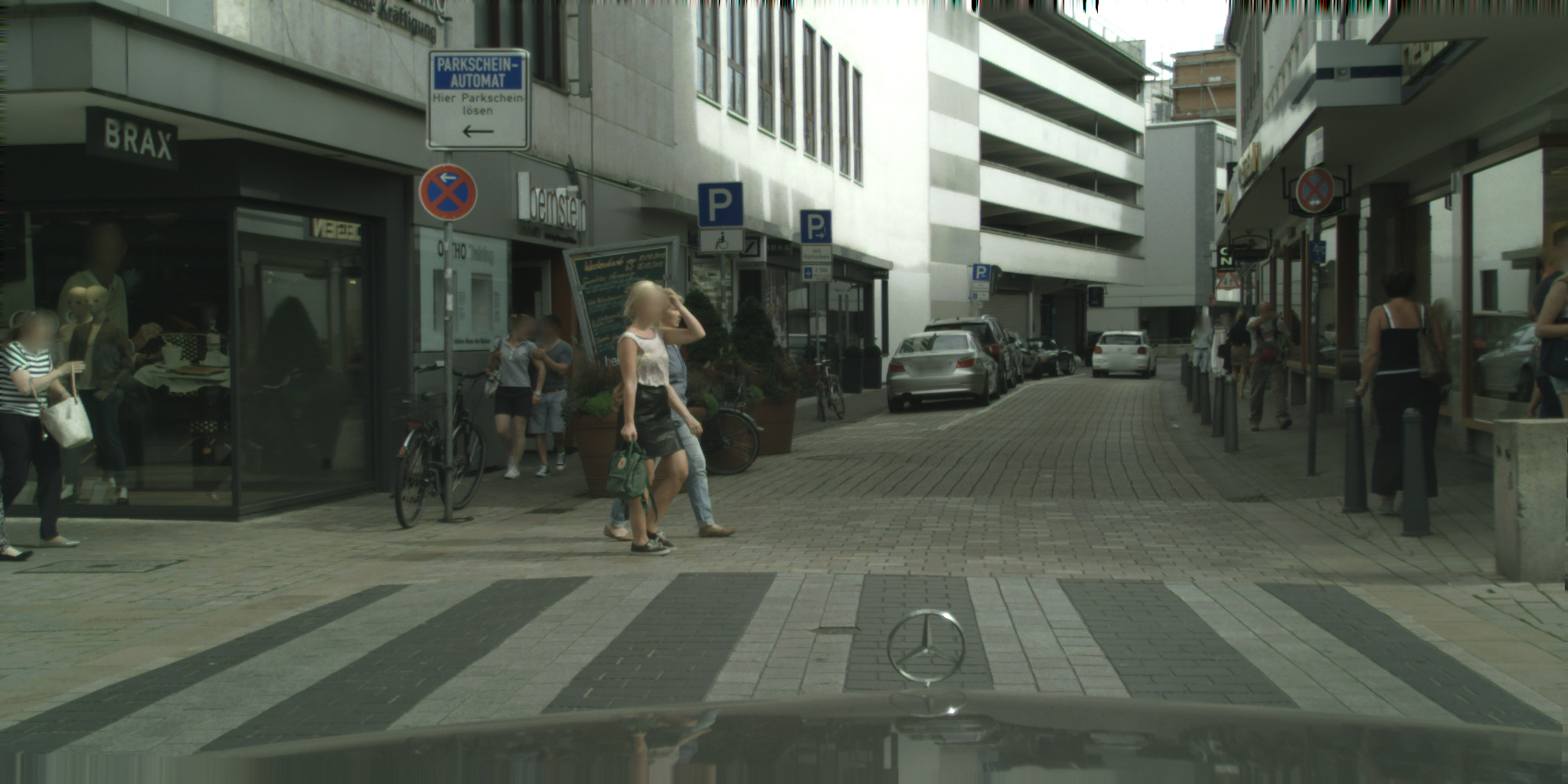}
\includegraphics[width=\textwidth]{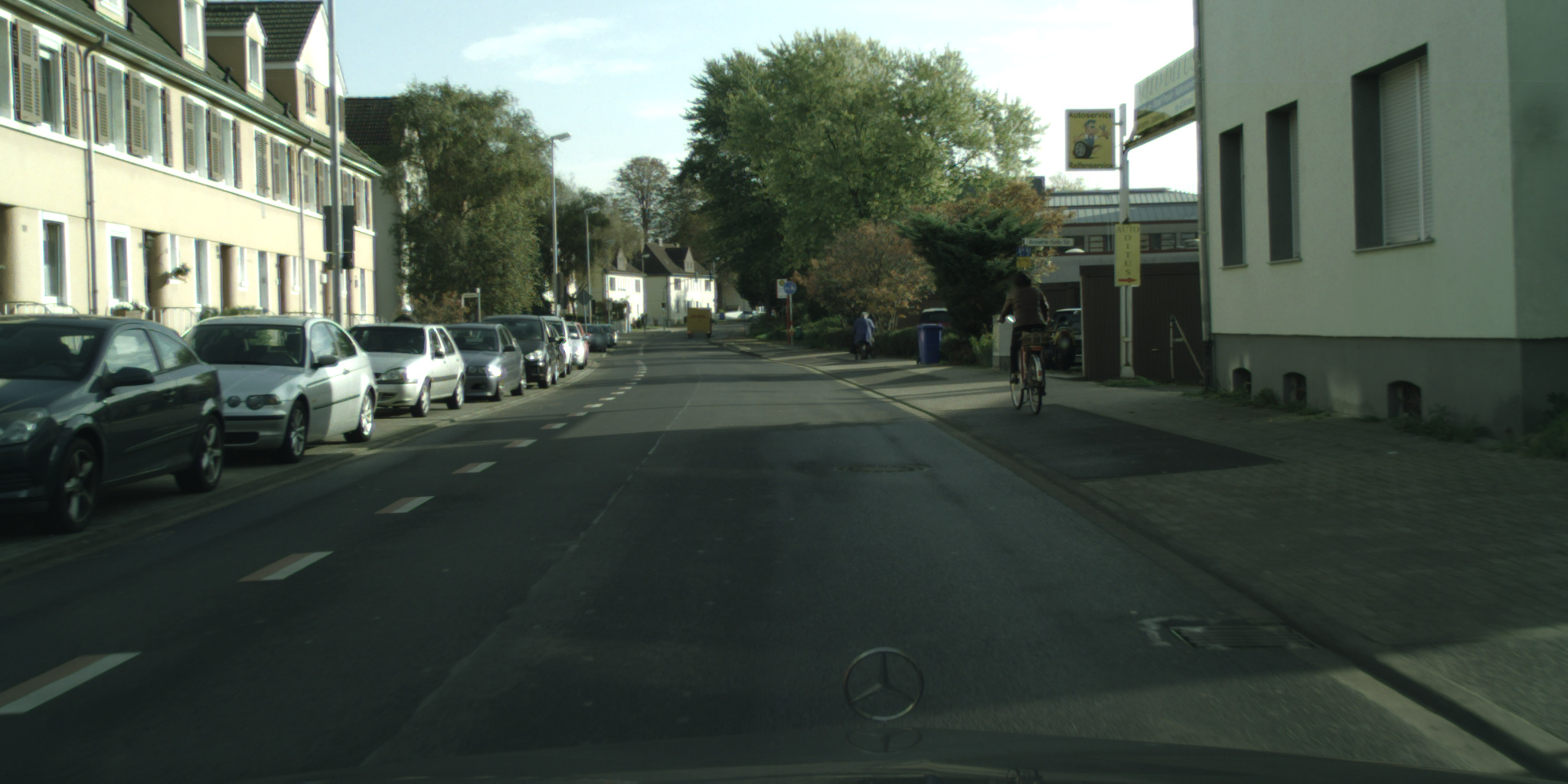}
\includegraphics[width=\textwidth]{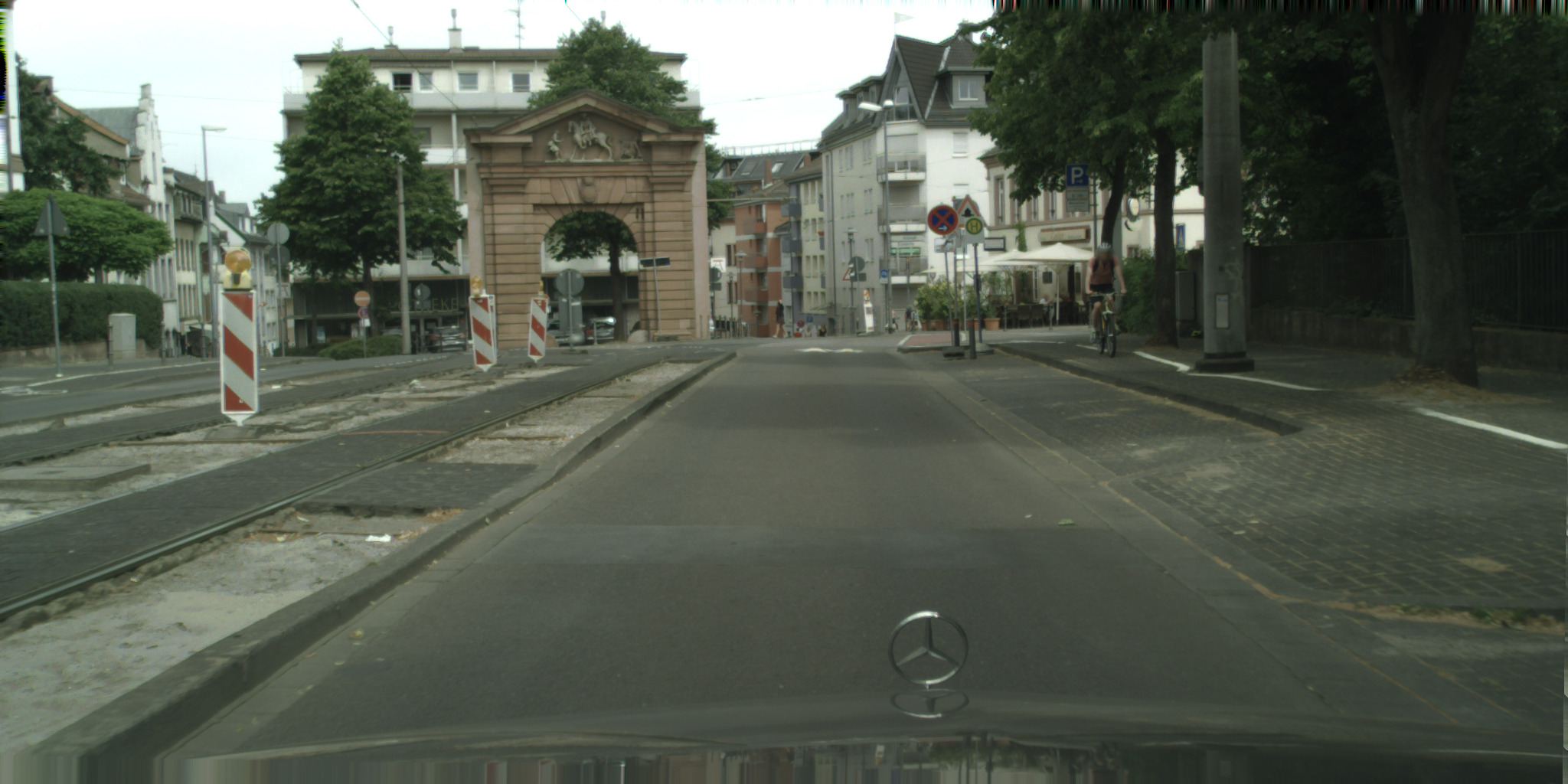}
\caption{Original}
\end{subfigure}
\begin{subfigure}[t]{0.3333333333333333\textwidth}
\centering
\includegraphics[width=\textwidth]{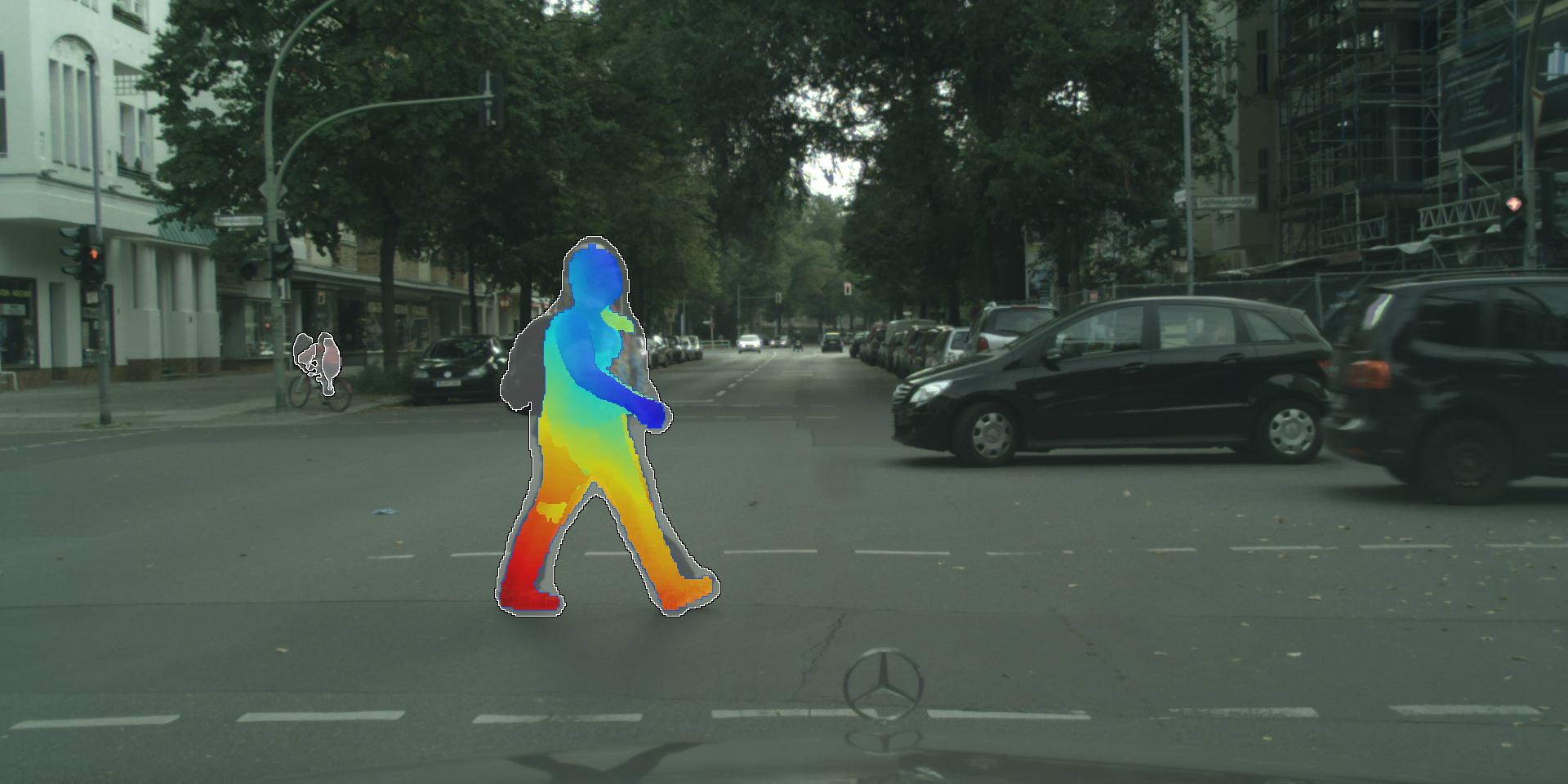}
\includegraphics[width=\textwidth]{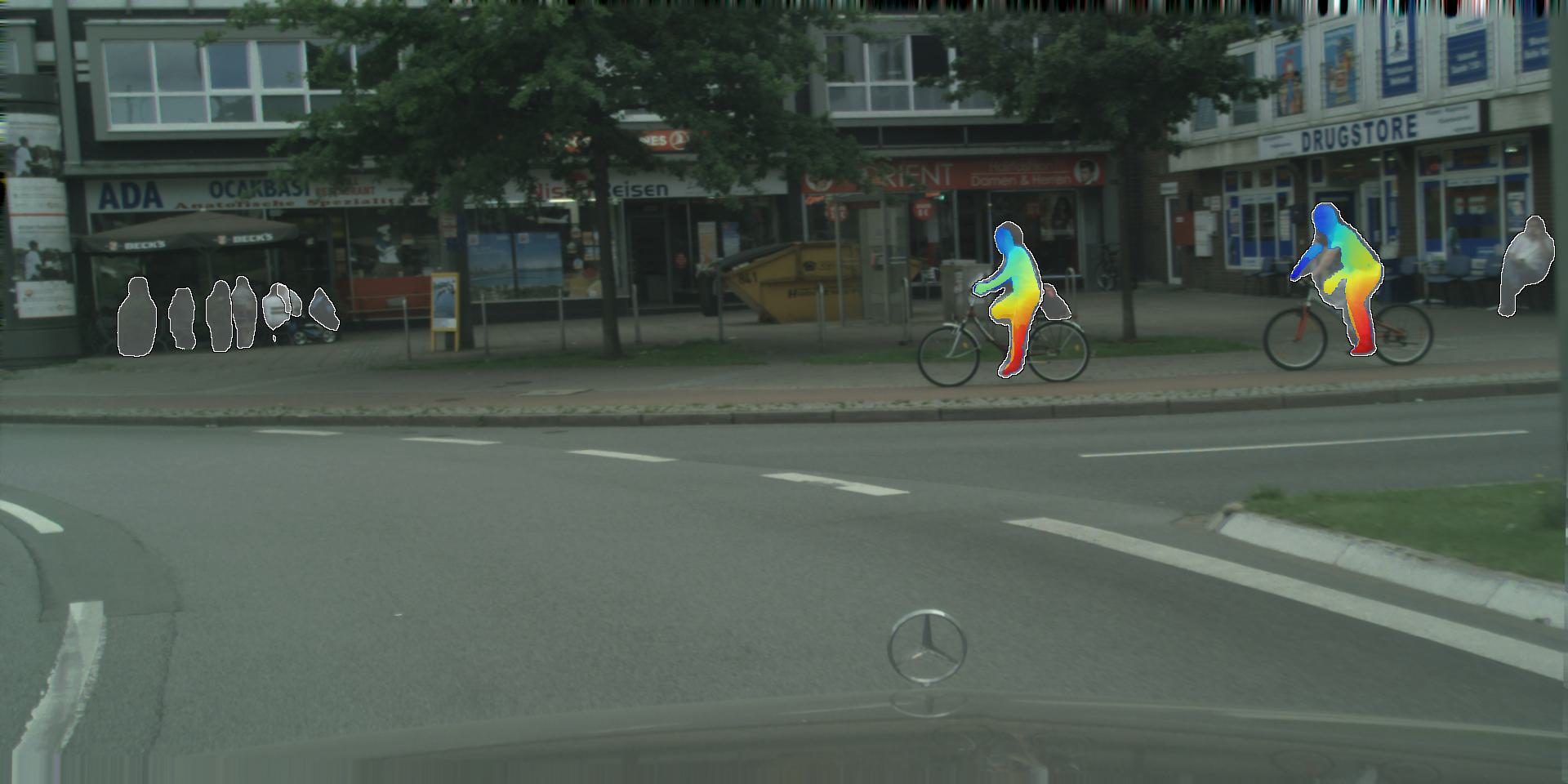}
\includegraphics[width=\textwidth]{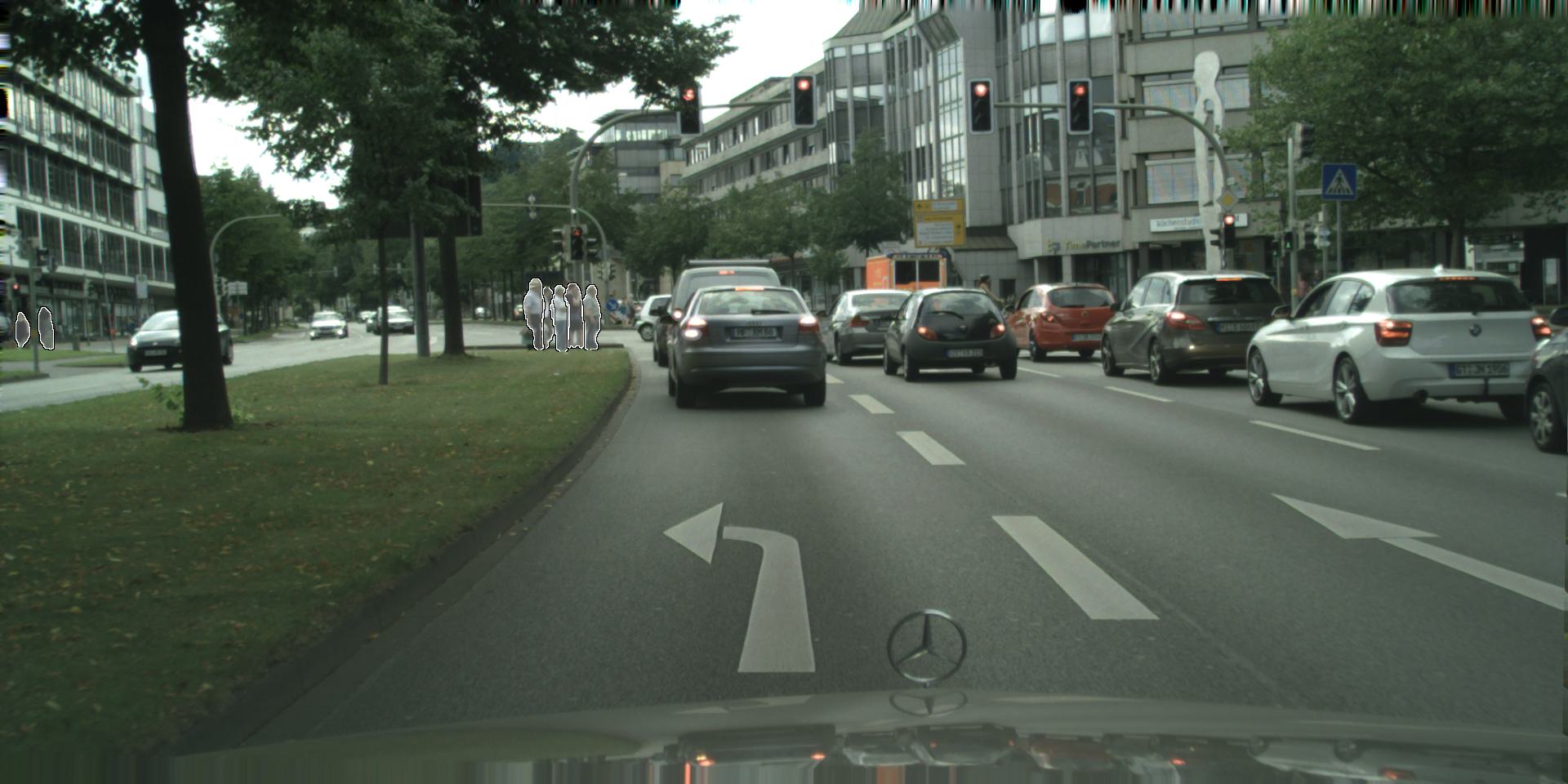}
\includegraphics[width=\textwidth]{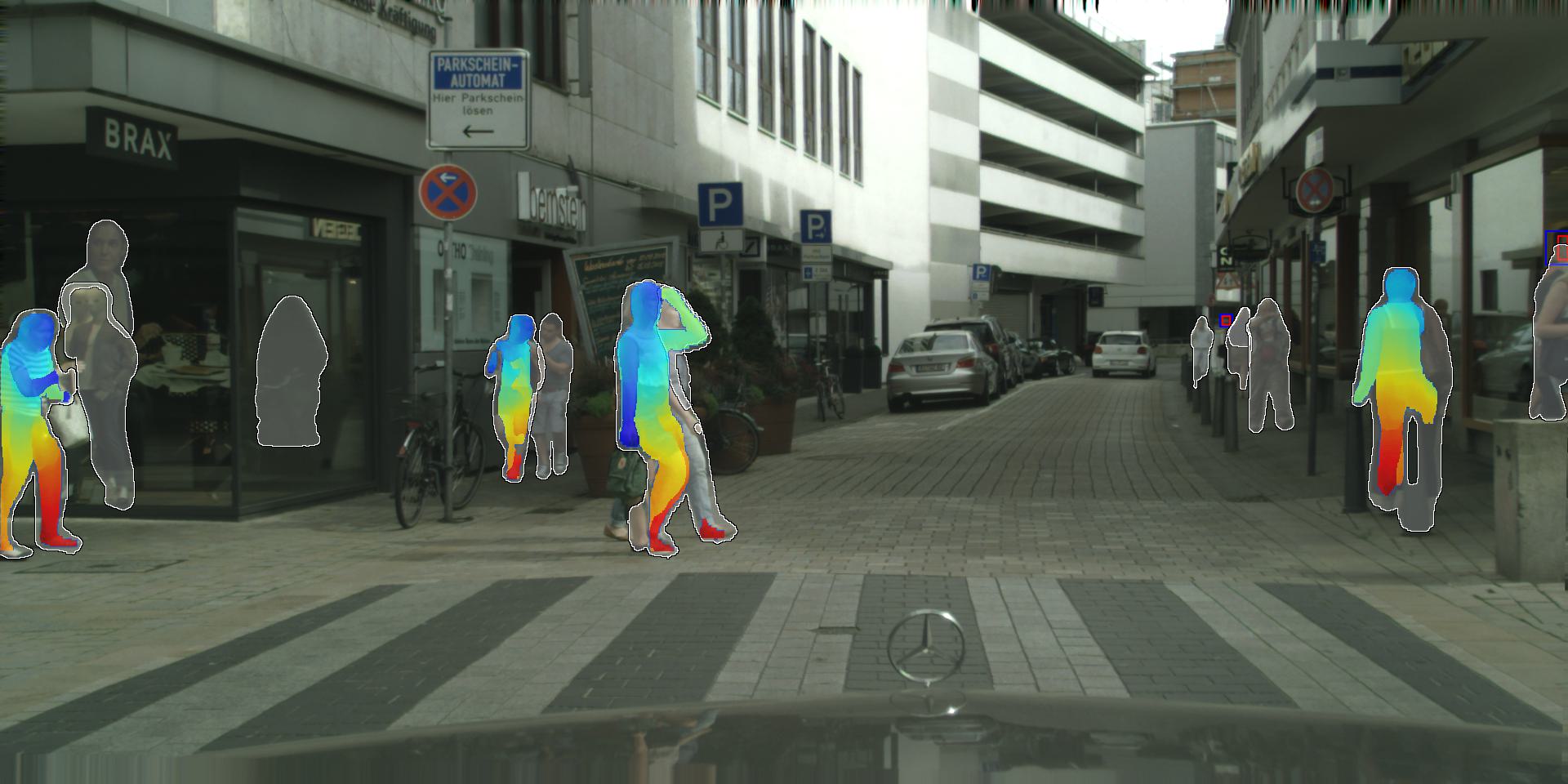}
\includegraphics[width=\textwidth]{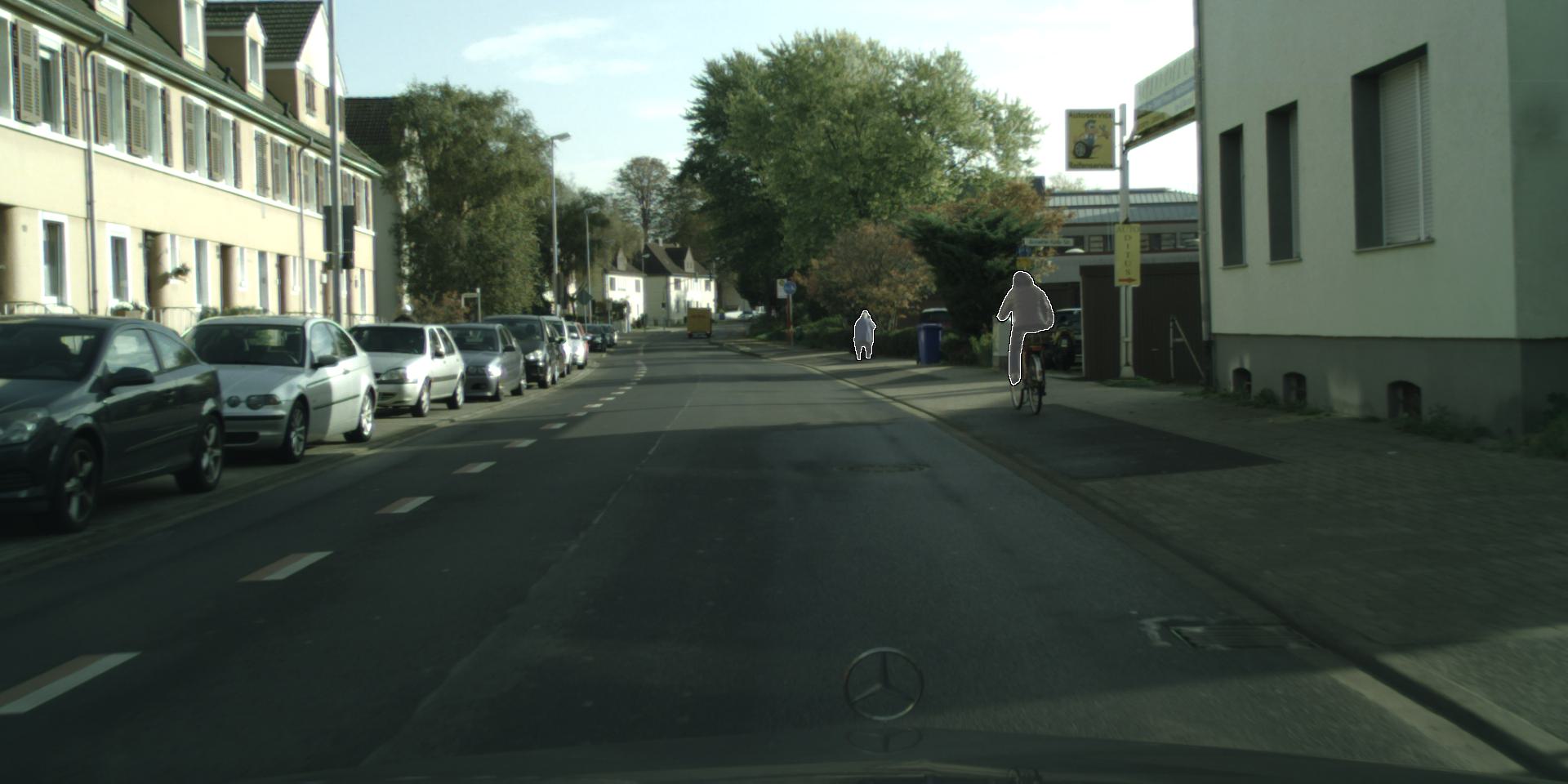}
\includegraphics[width=\textwidth]{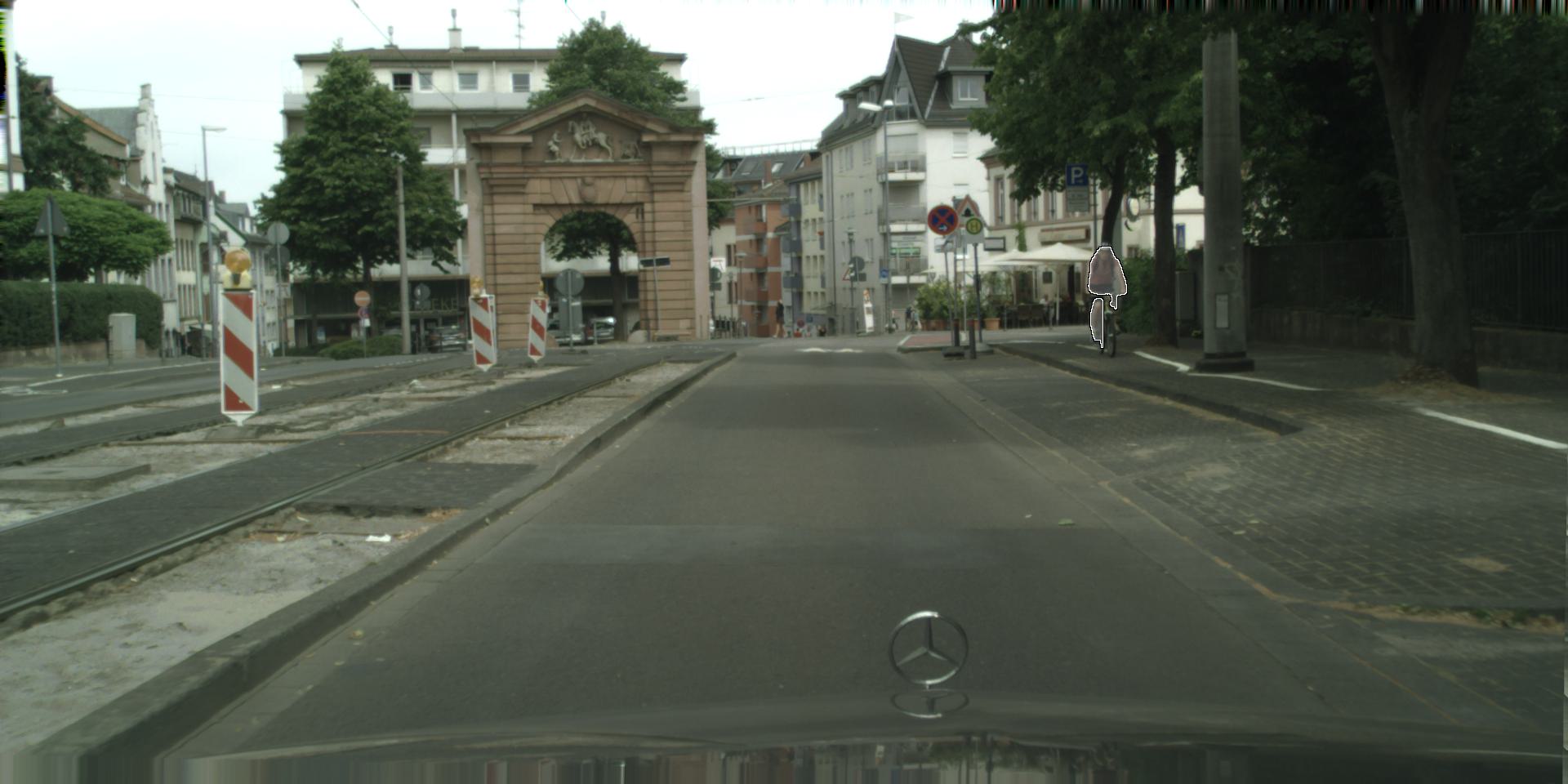}
\caption{Detections}
\end{subfigure}
\begin{subfigure}[t]{0.3333333333333333\textwidth}
\centering
\includegraphics[width=\textwidth]{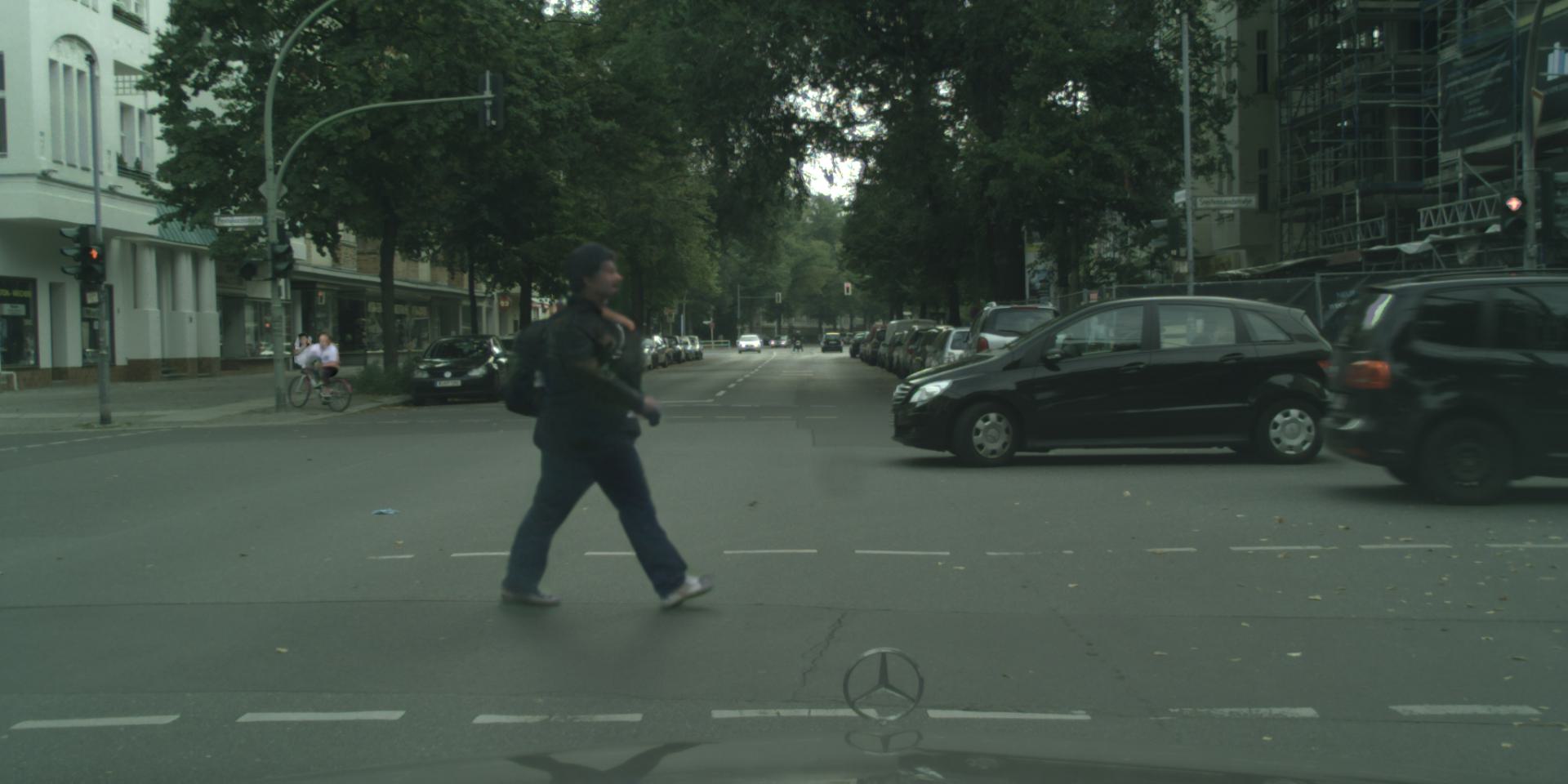}
\includegraphics[width=\textwidth]{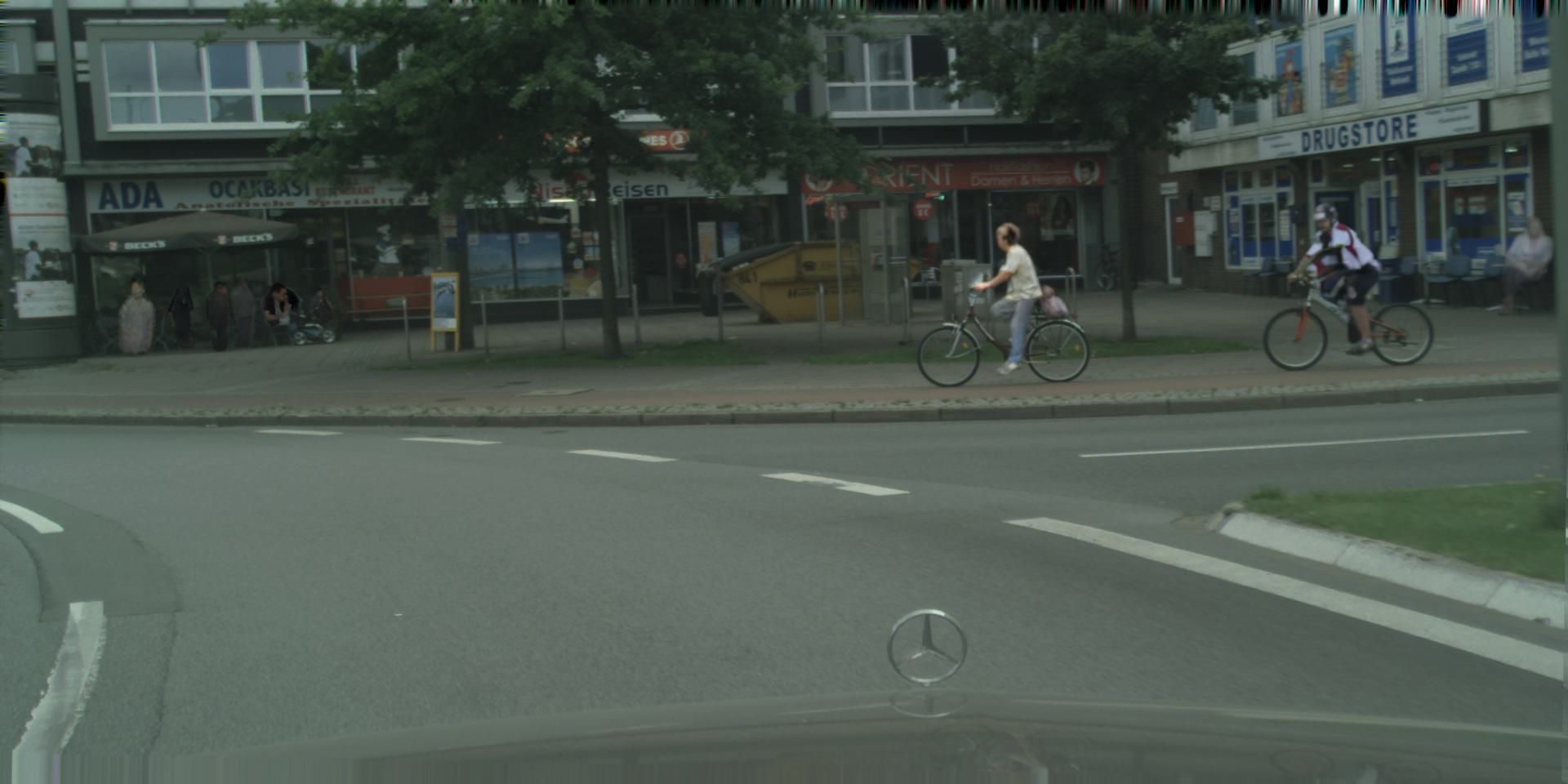}
\includegraphics[width=\textwidth]{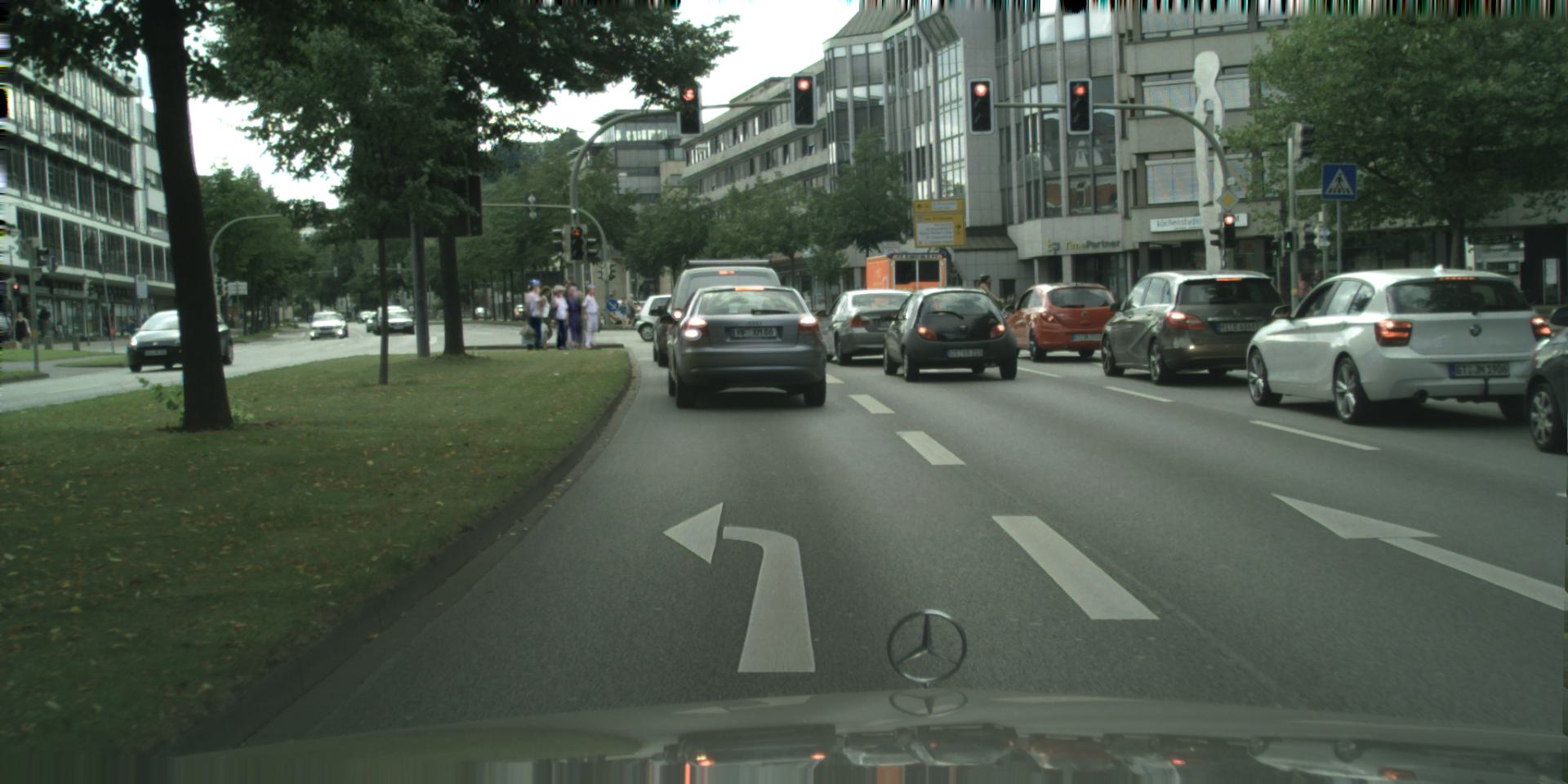}
\includegraphics[width=\textwidth]{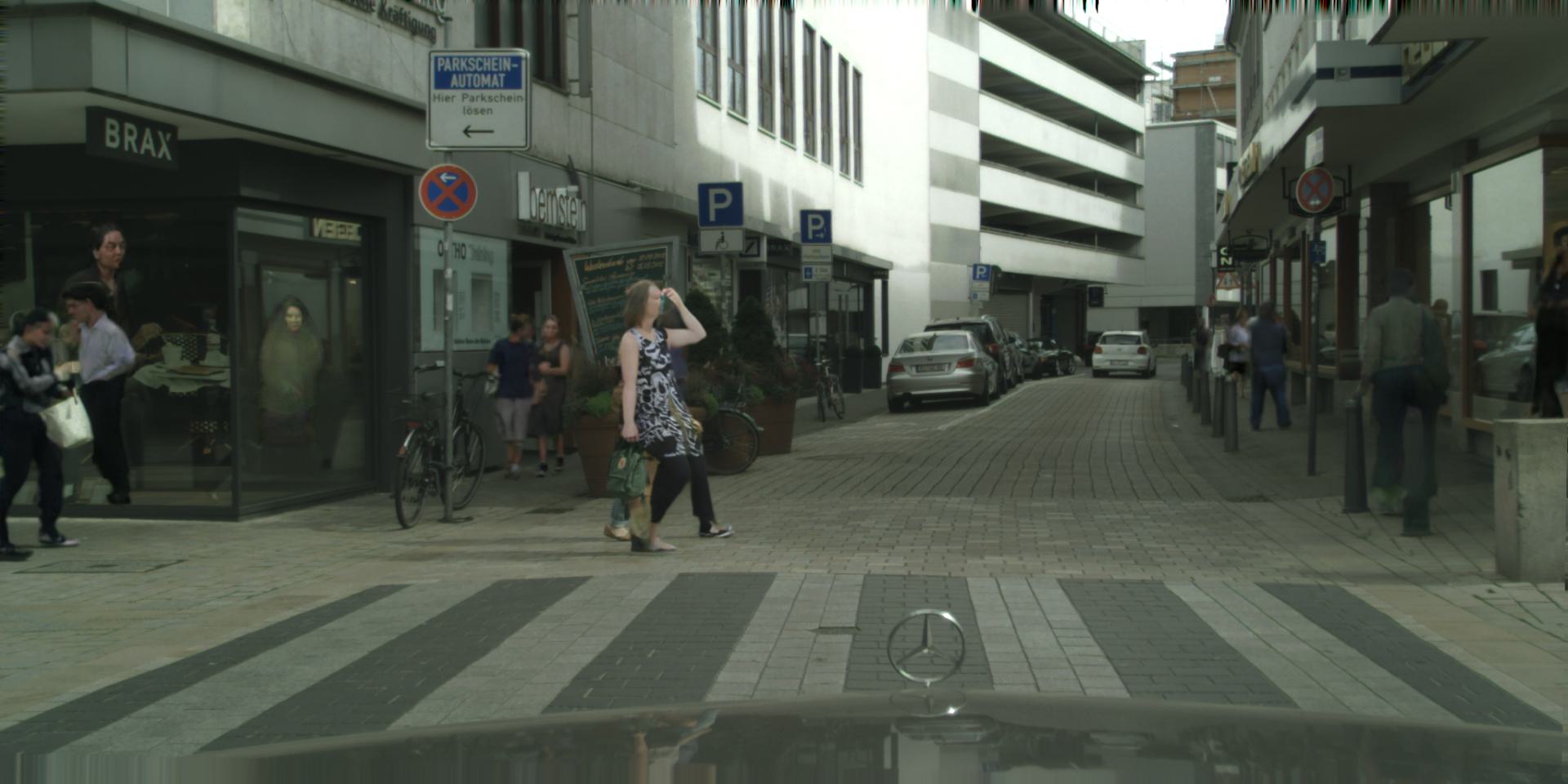}
\includegraphics[width=\textwidth]{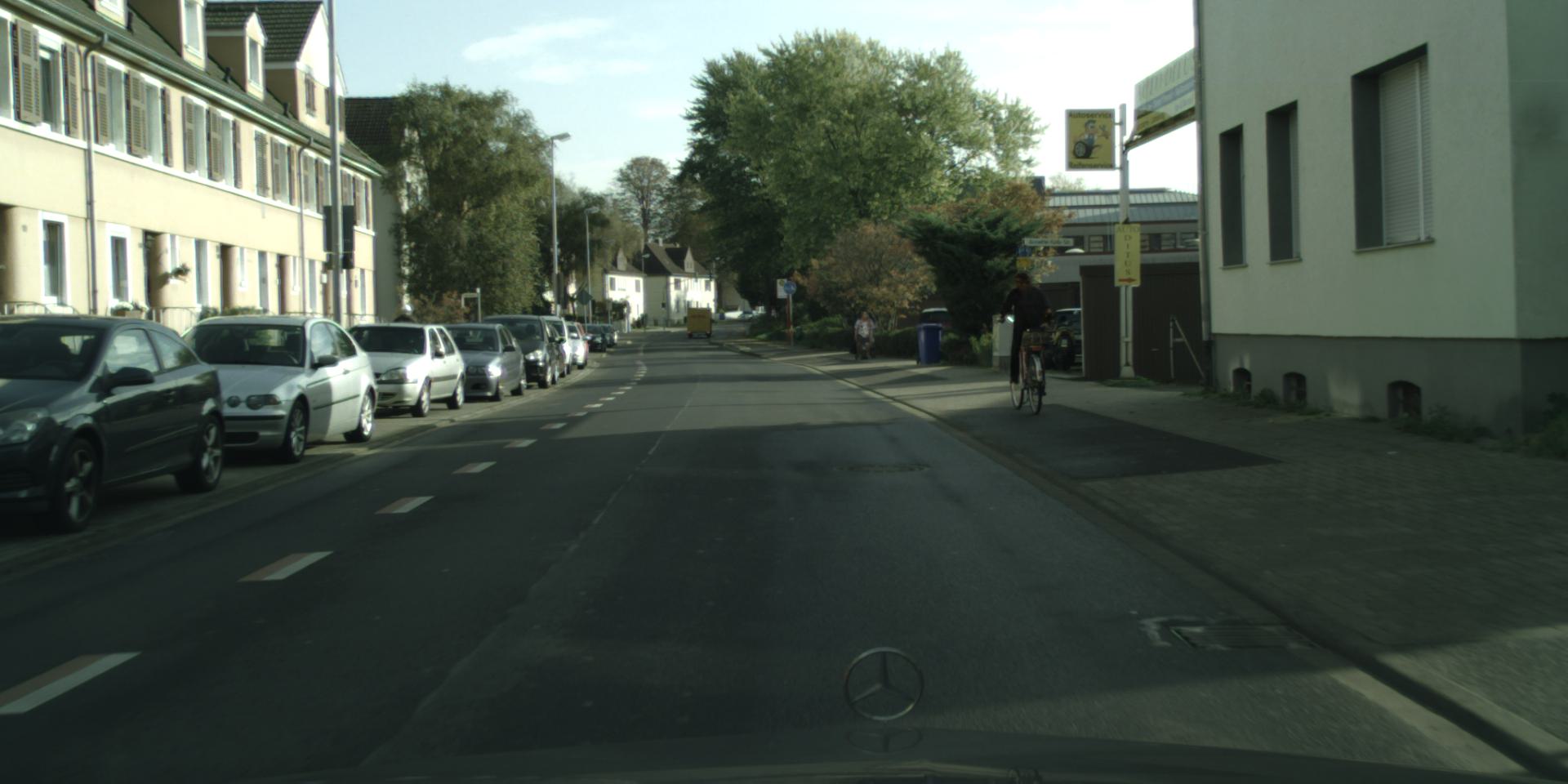}
\includegraphics[width=\textwidth]{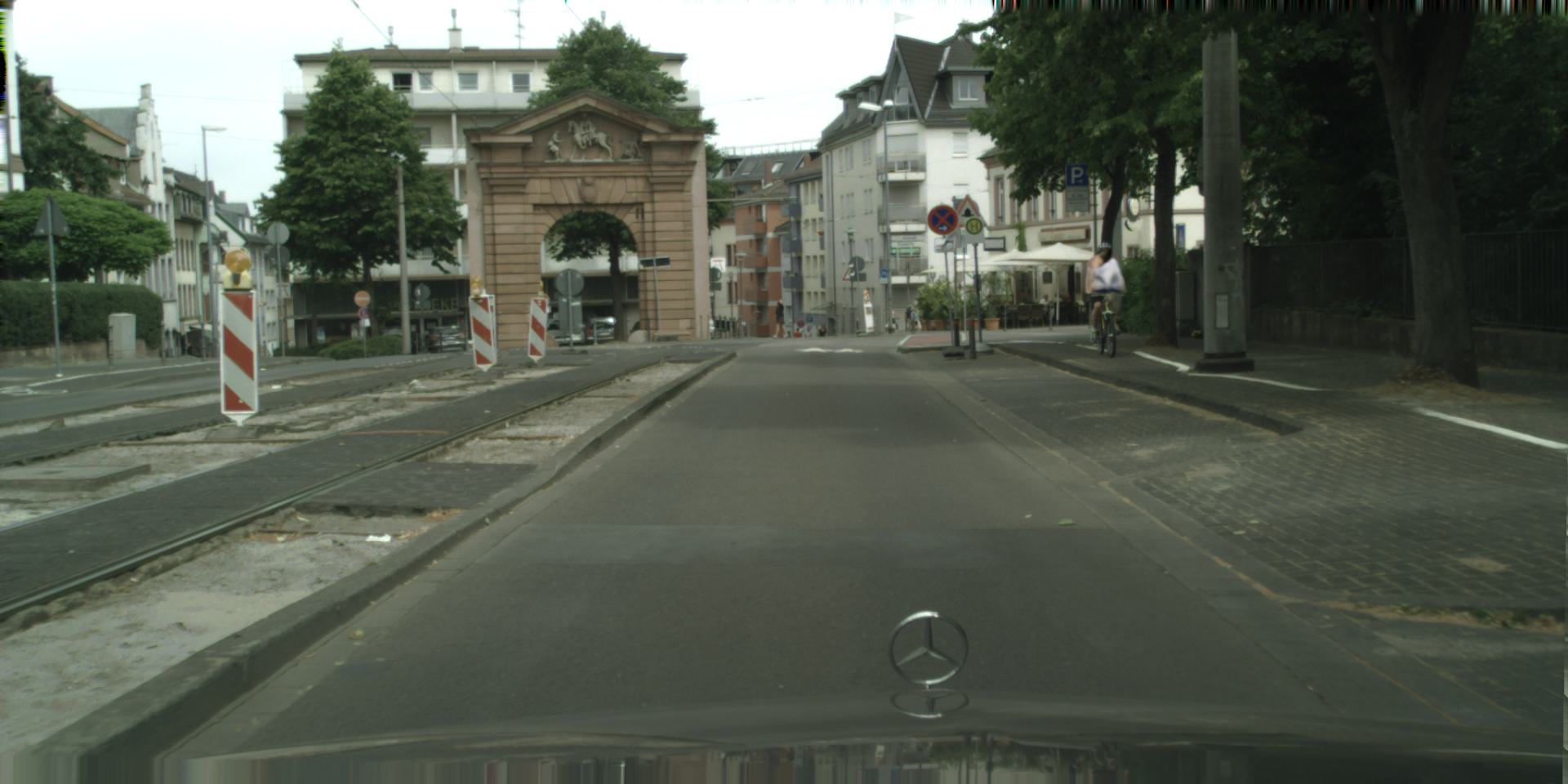}
\caption{Anonymized}
\end{subfigure}
\caption{Randomly generated images from the Cityscapes \cite{cityscapes} test set. (a) shows the original image (with blurred face to follow Cityscapes licensing), (b) the detections and (c) anonymized image}
\label{fig:citscapes_random1}
\end{figure*}

\begin{figure*}[t]
\centering
\begin{subfigure}[t]{0.3333333333333333\textwidth}
\centering
\includegraphics[width=\textwidth]{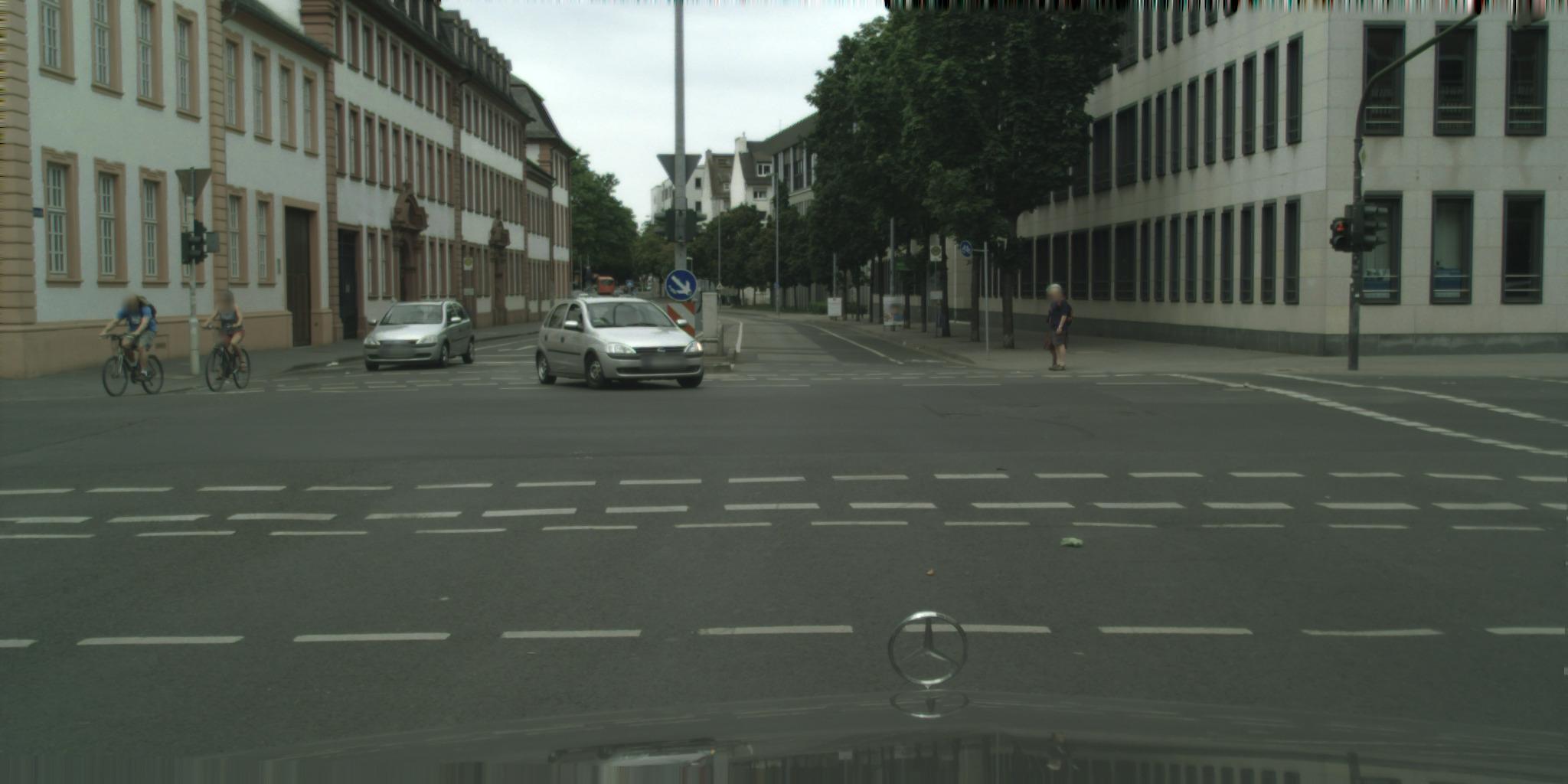}
\includegraphics[width=\textwidth]{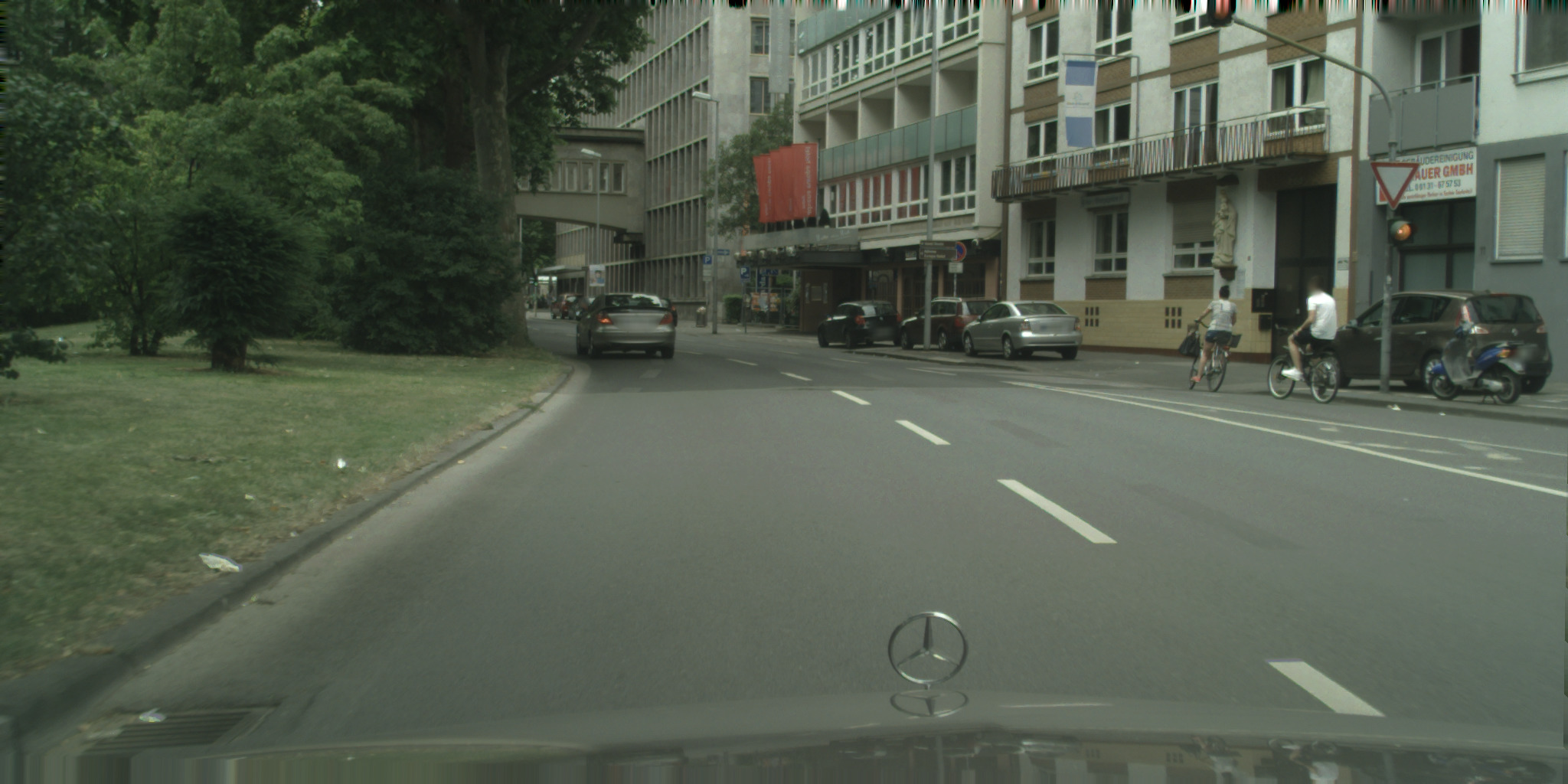}
\includegraphics[width=\textwidth]{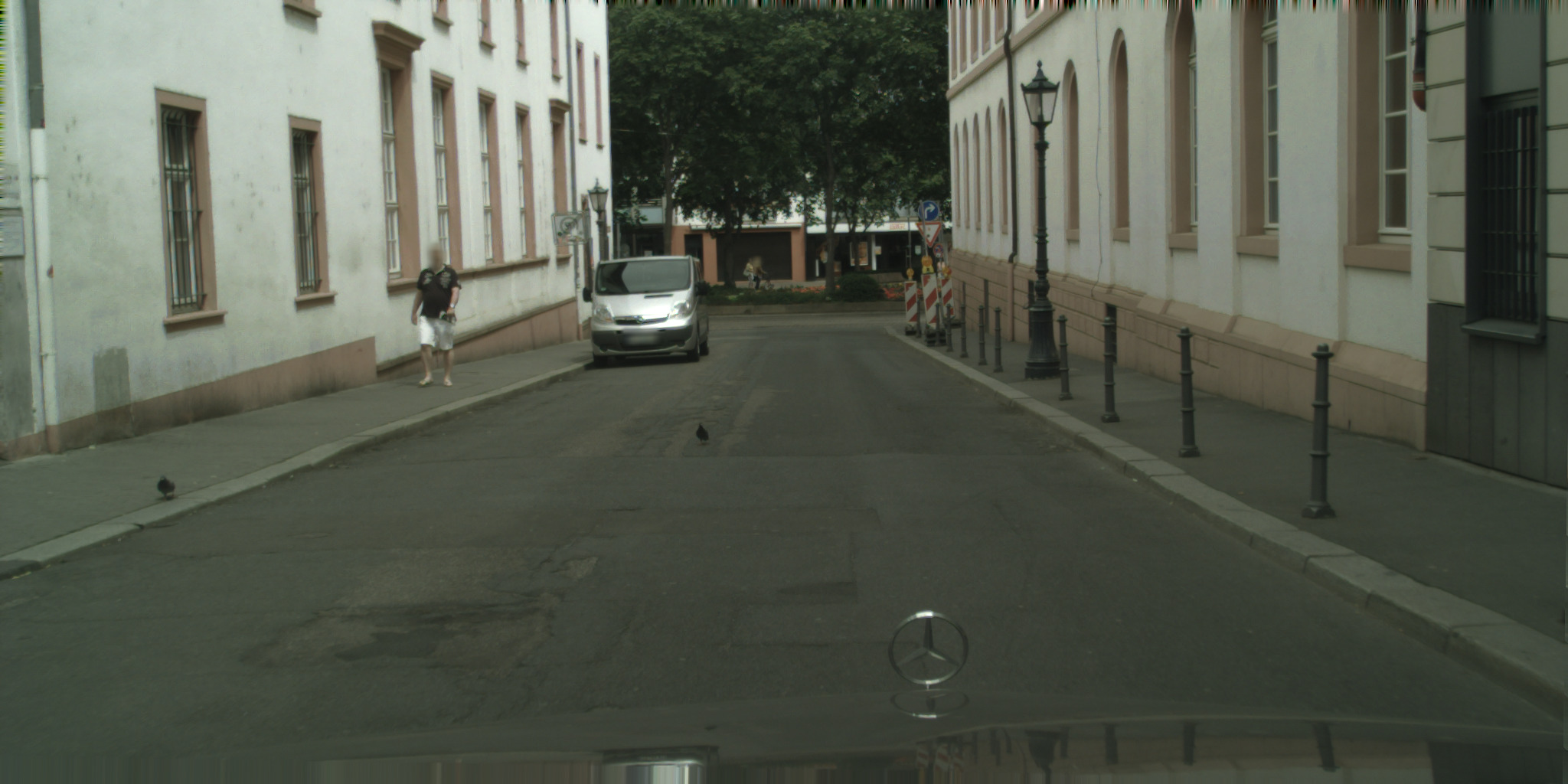}
\includegraphics[width=\textwidth]{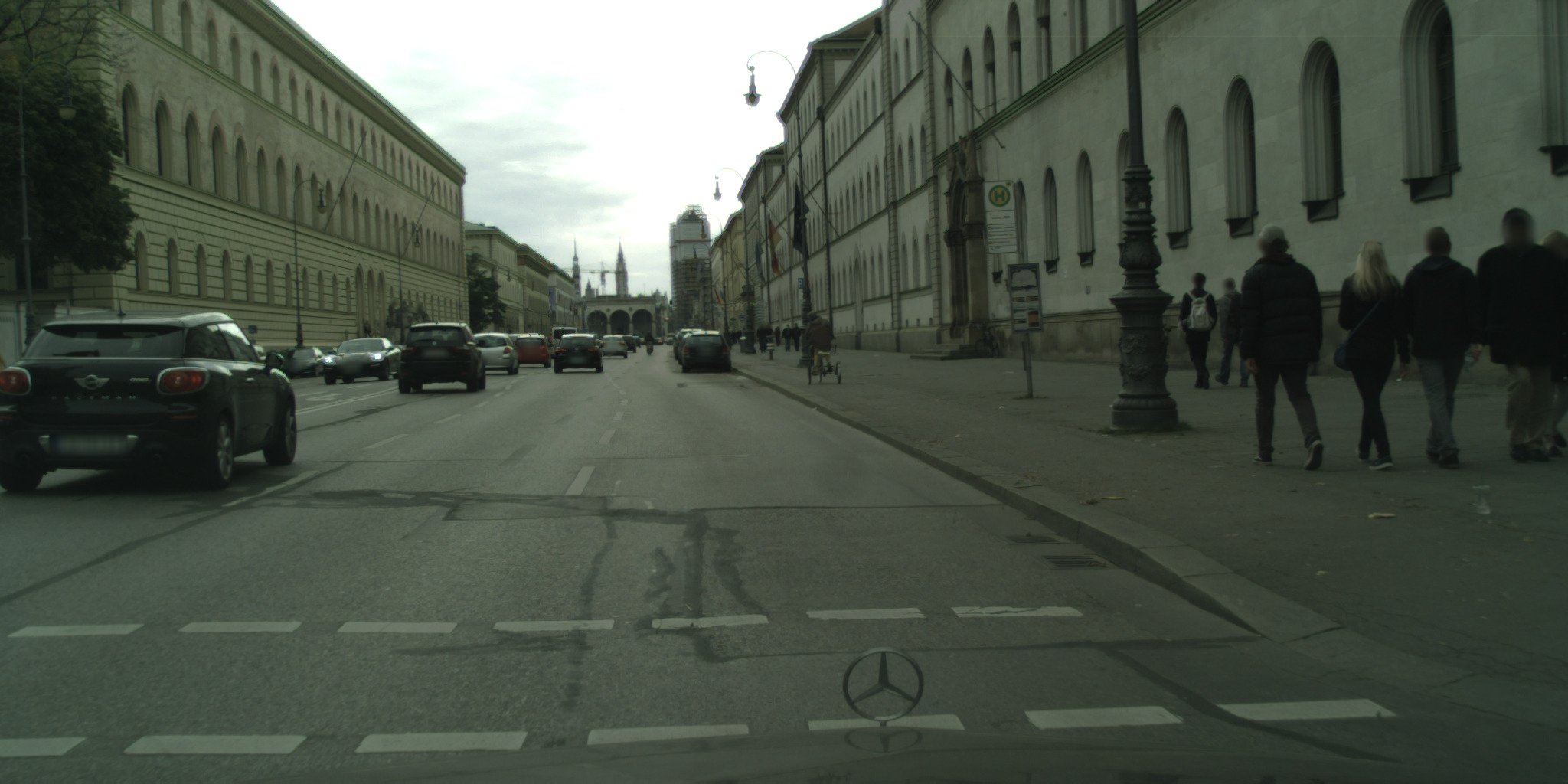}
\includegraphics[width=\textwidth]{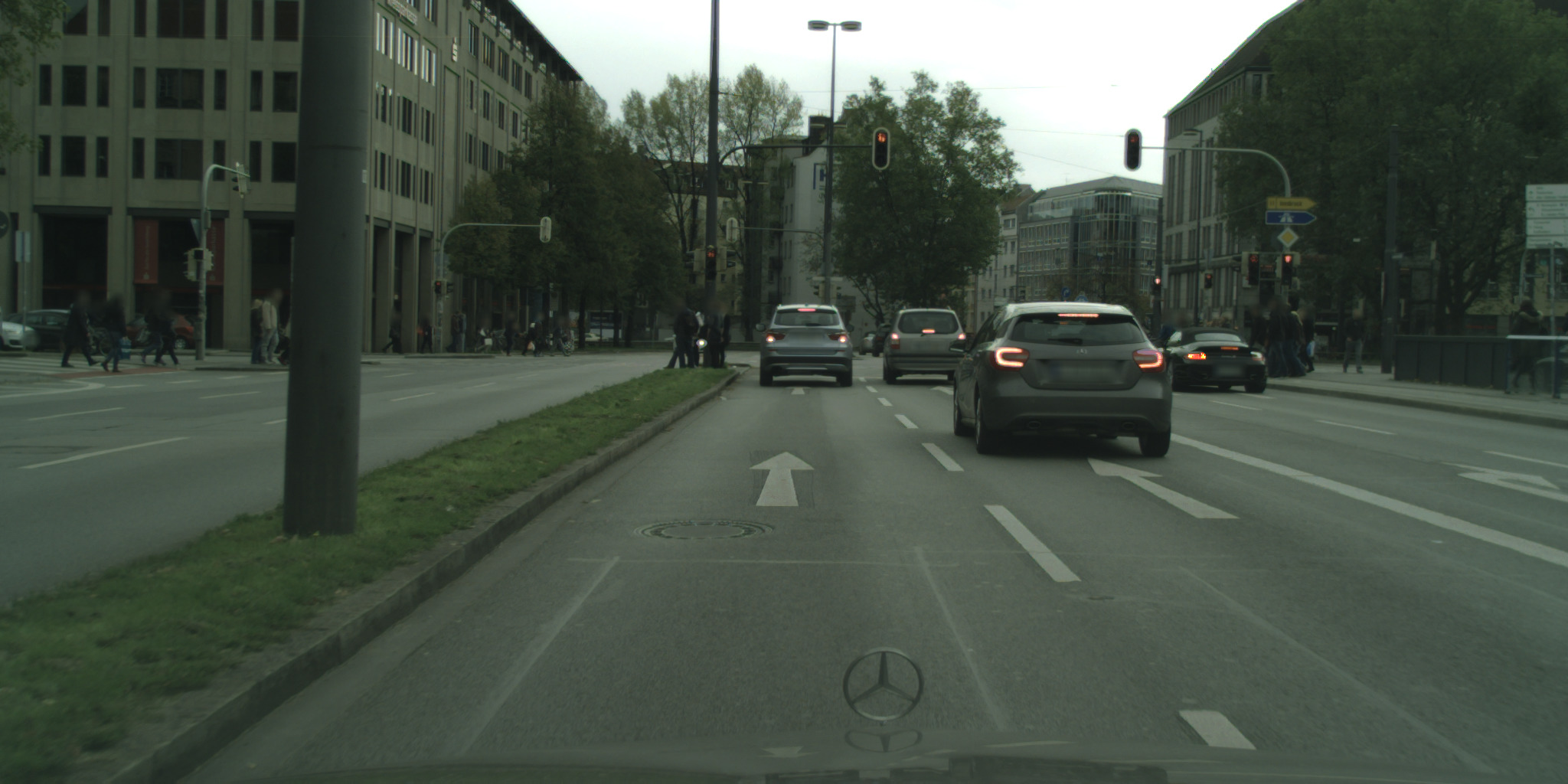}
\includegraphics[width=\textwidth]{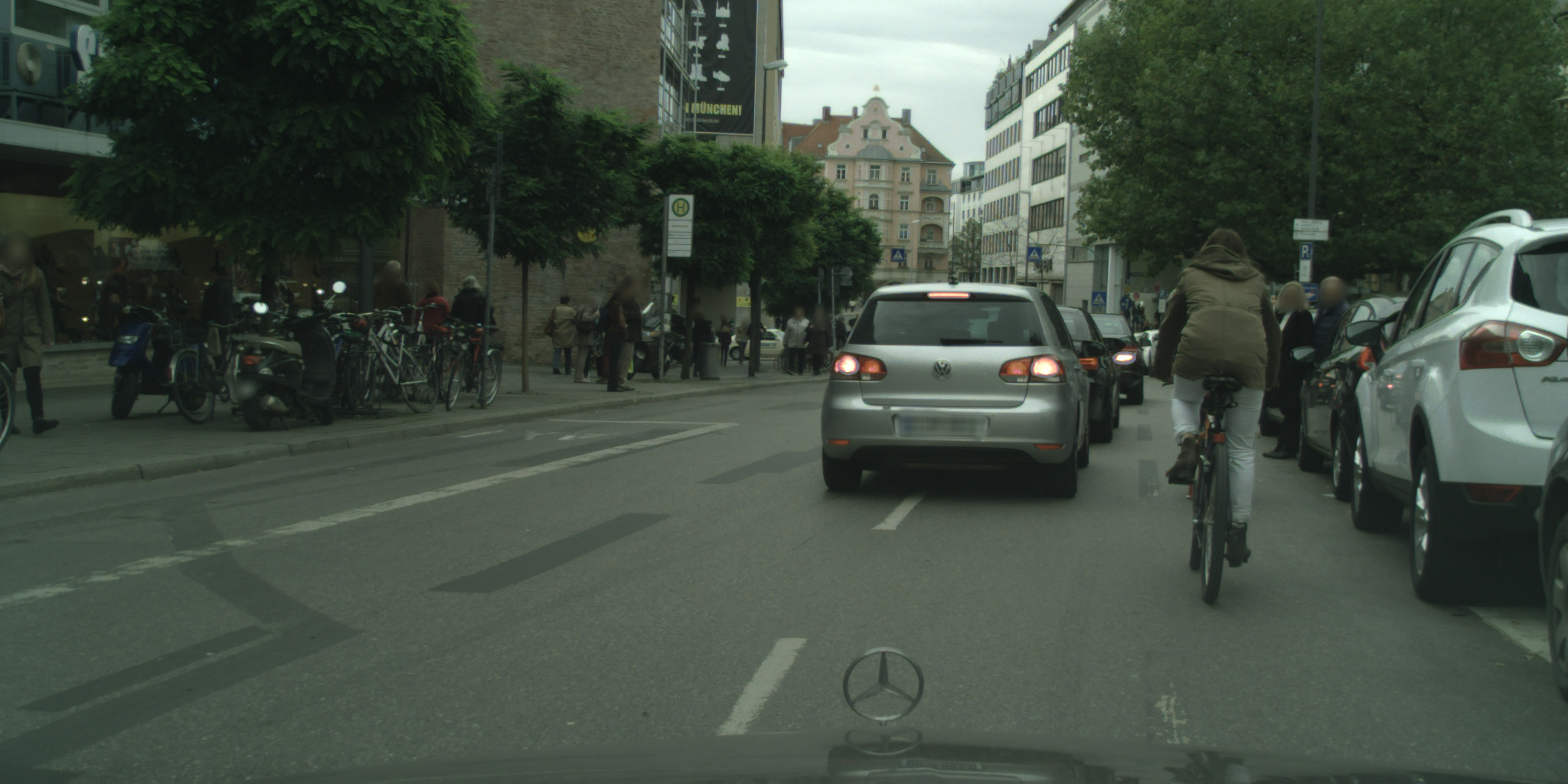}
\caption{Original}
\end{subfigure}
\begin{subfigure}[t]{0.3333333333333333\textwidth}
\centering
\includegraphics[width=\textwidth]{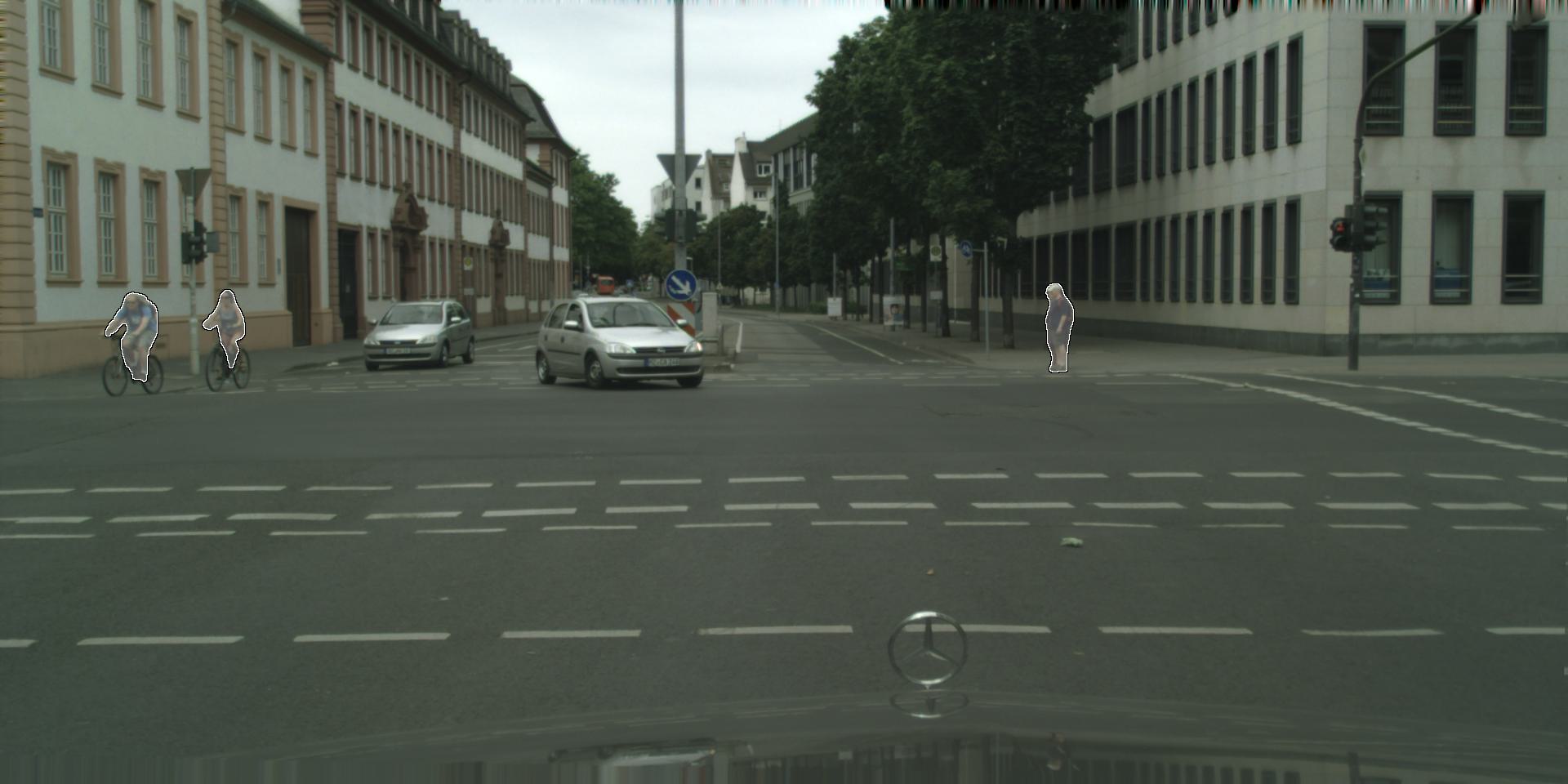}
\includegraphics[width=\textwidth]{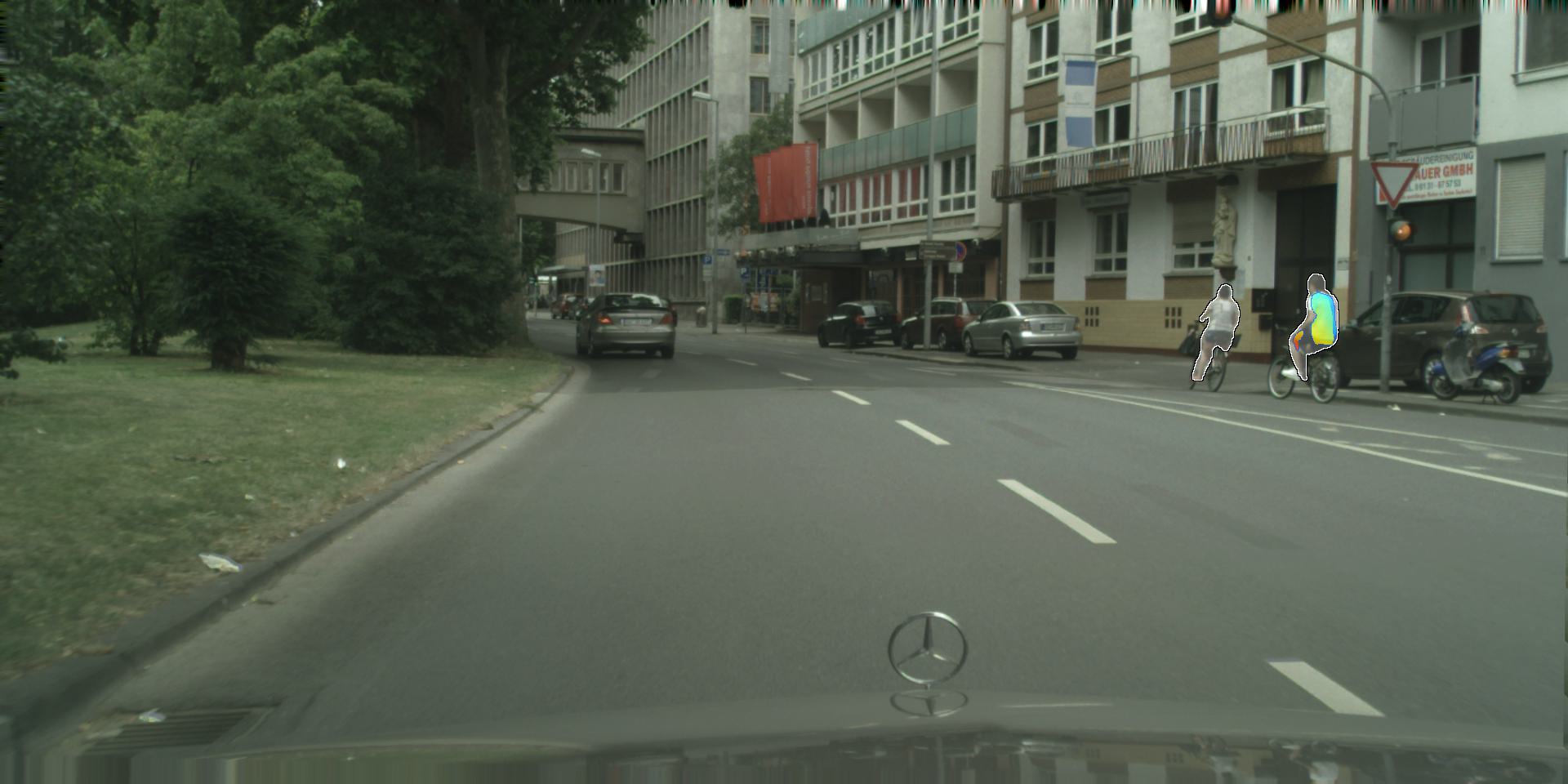}
\includegraphics[width=\textwidth]{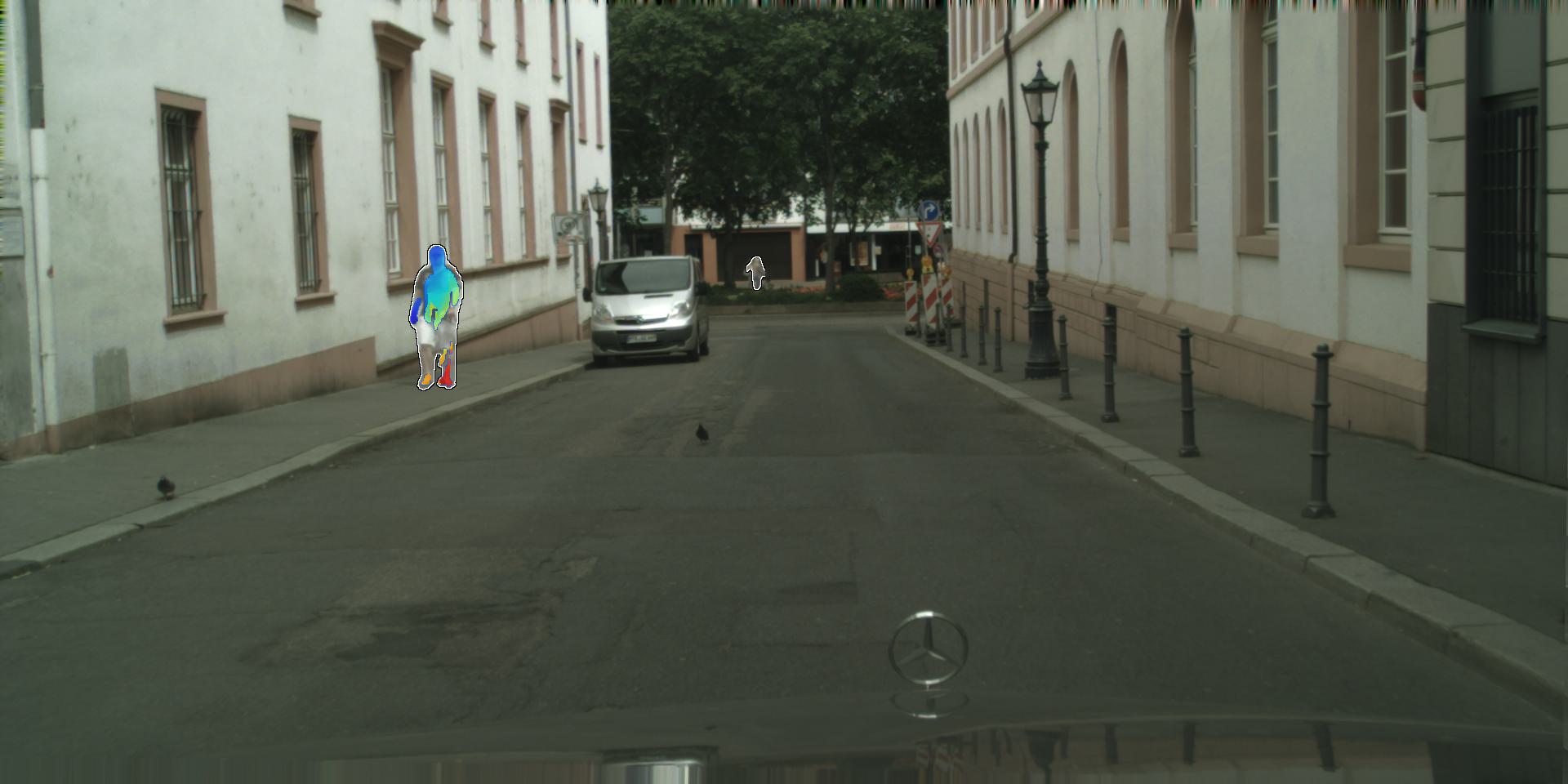}
\includegraphics[width=\textwidth]{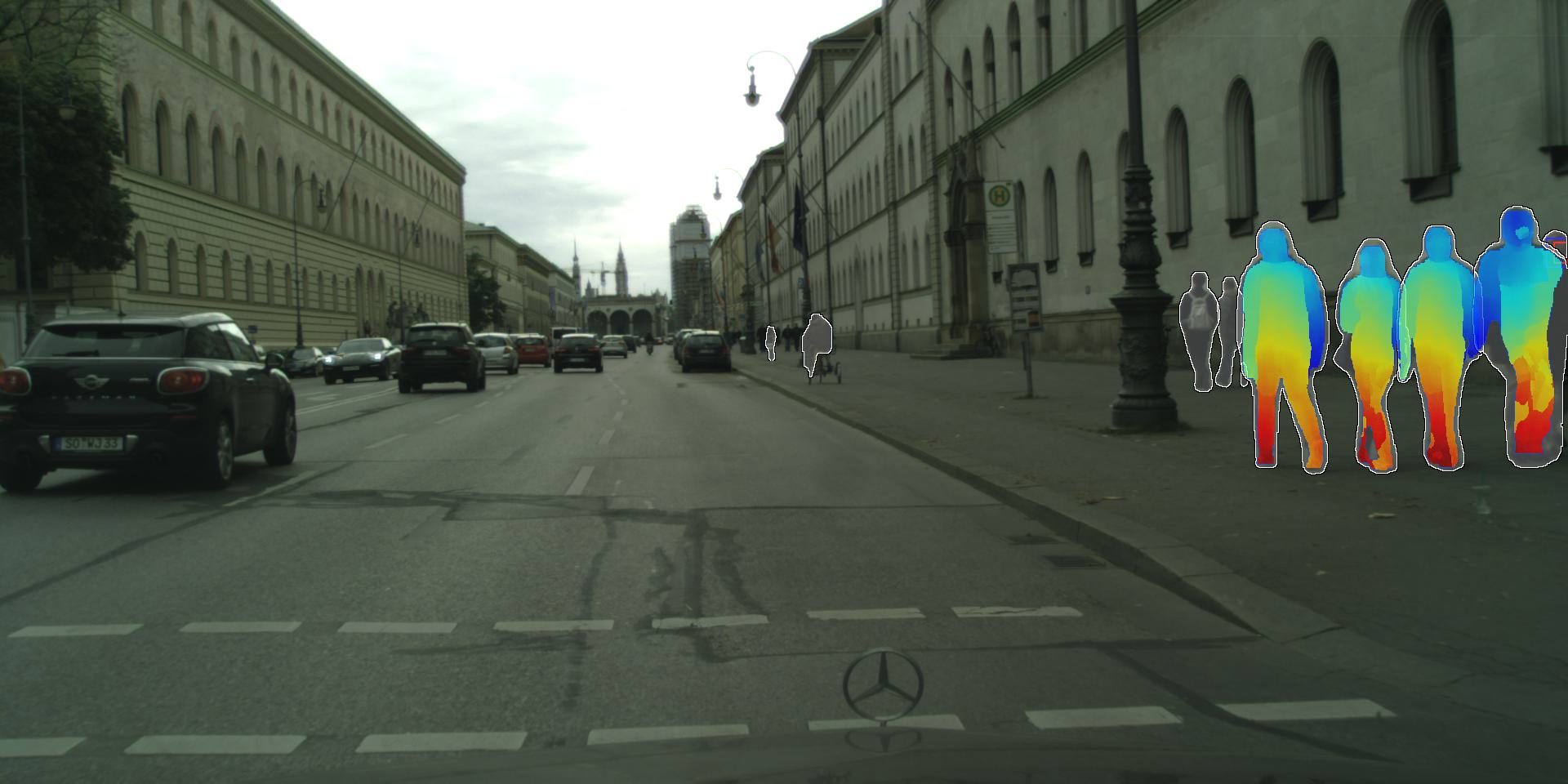}
\includegraphics[width=\textwidth]{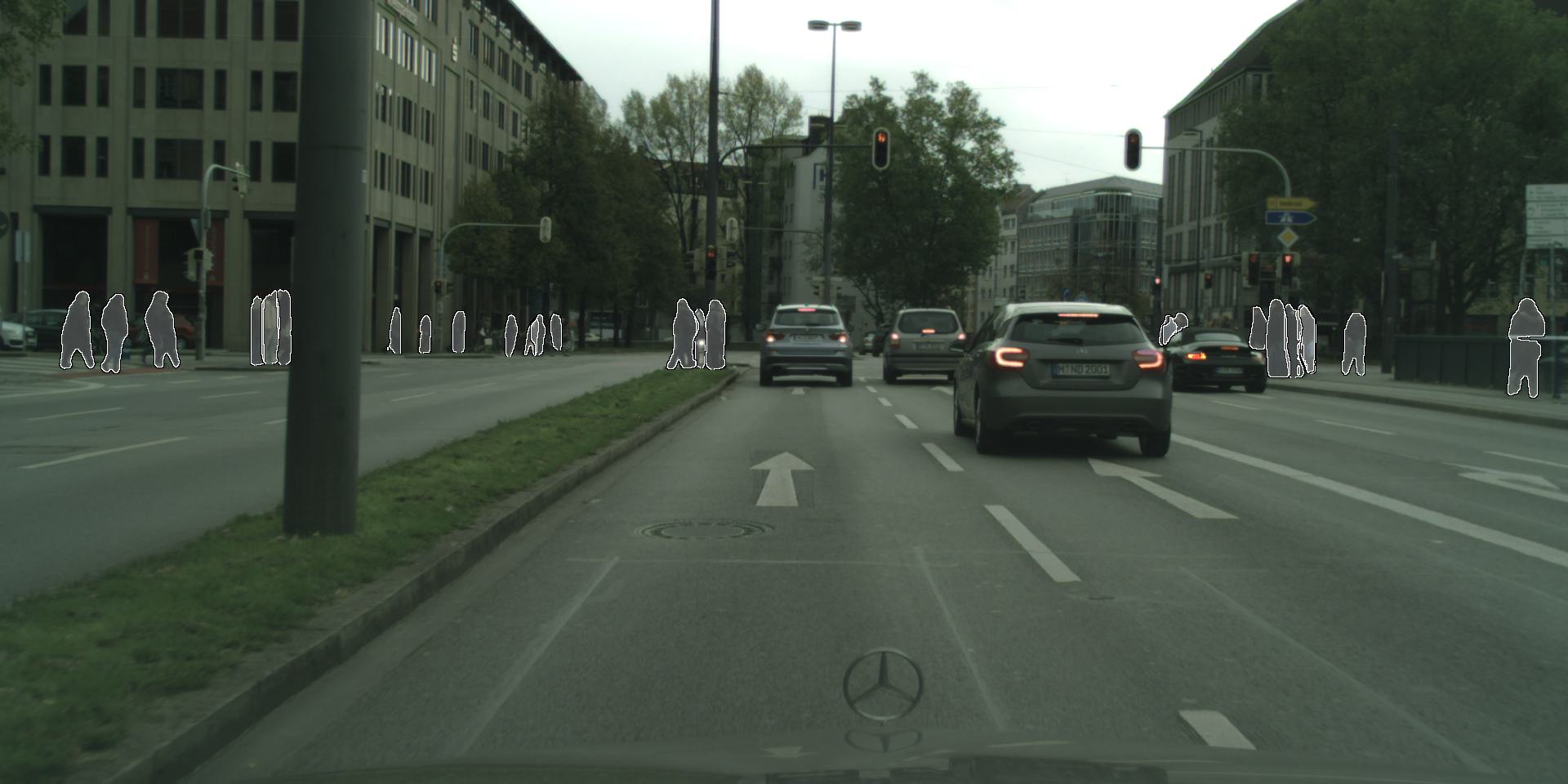}
\includegraphics[width=\textwidth]{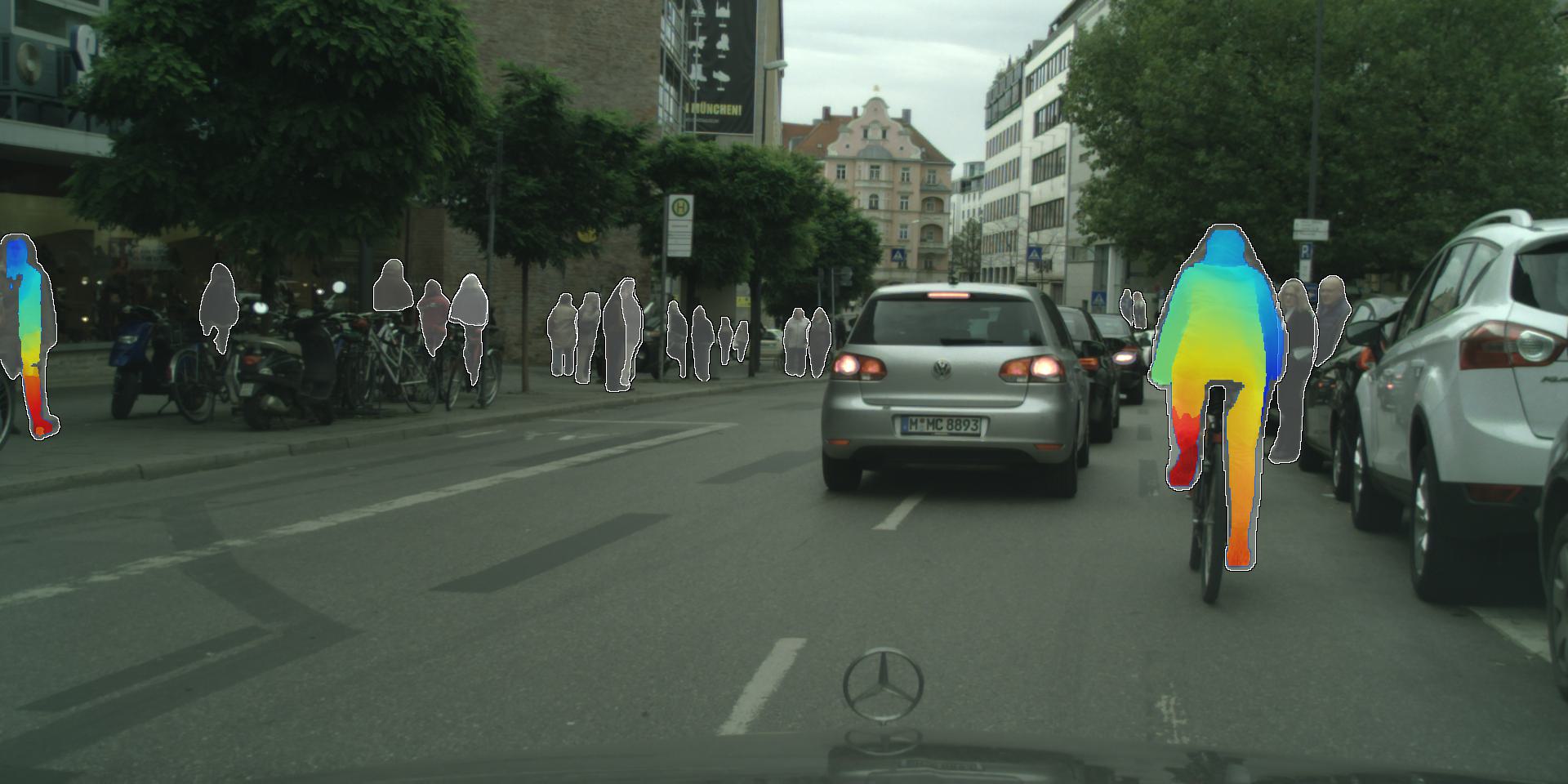}
\caption{Detections}
\end{subfigure}
\begin{subfigure}[t]{0.3333333333333333\textwidth}
\centering
\includegraphics[width=\textwidth]{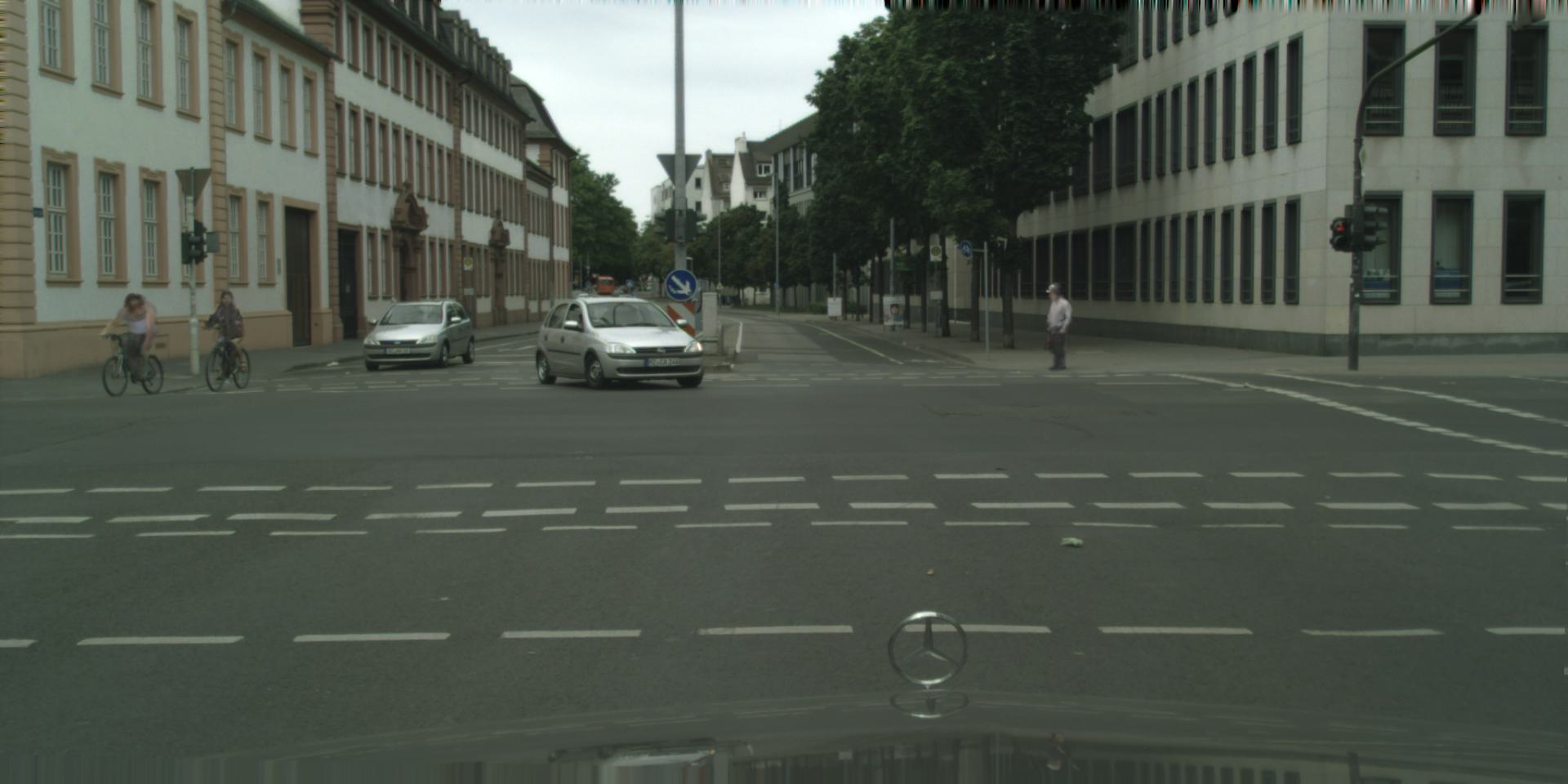}
\includegraphics[width=\textwidth]{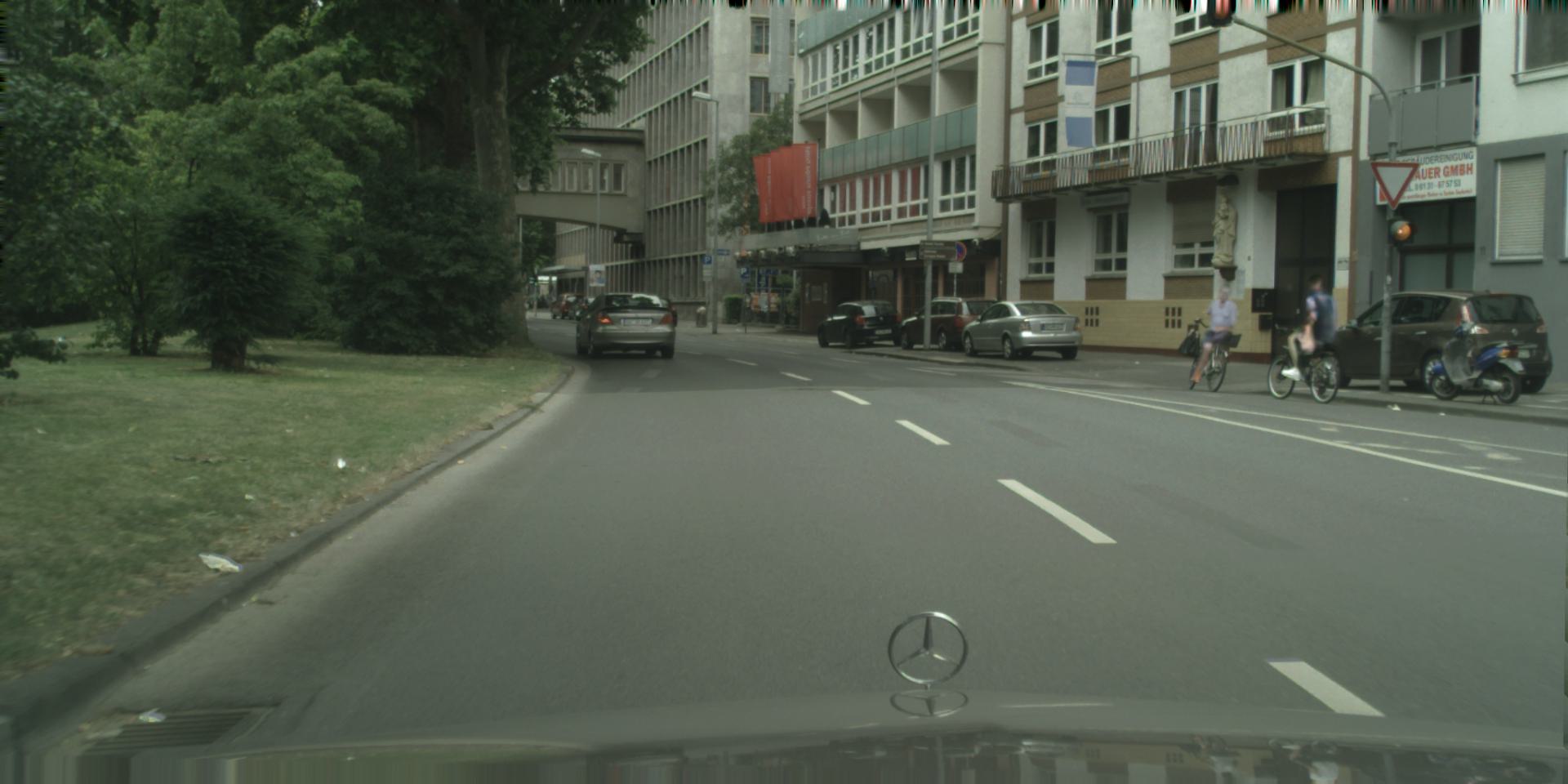}
\includegraphics[width=\textwidth]{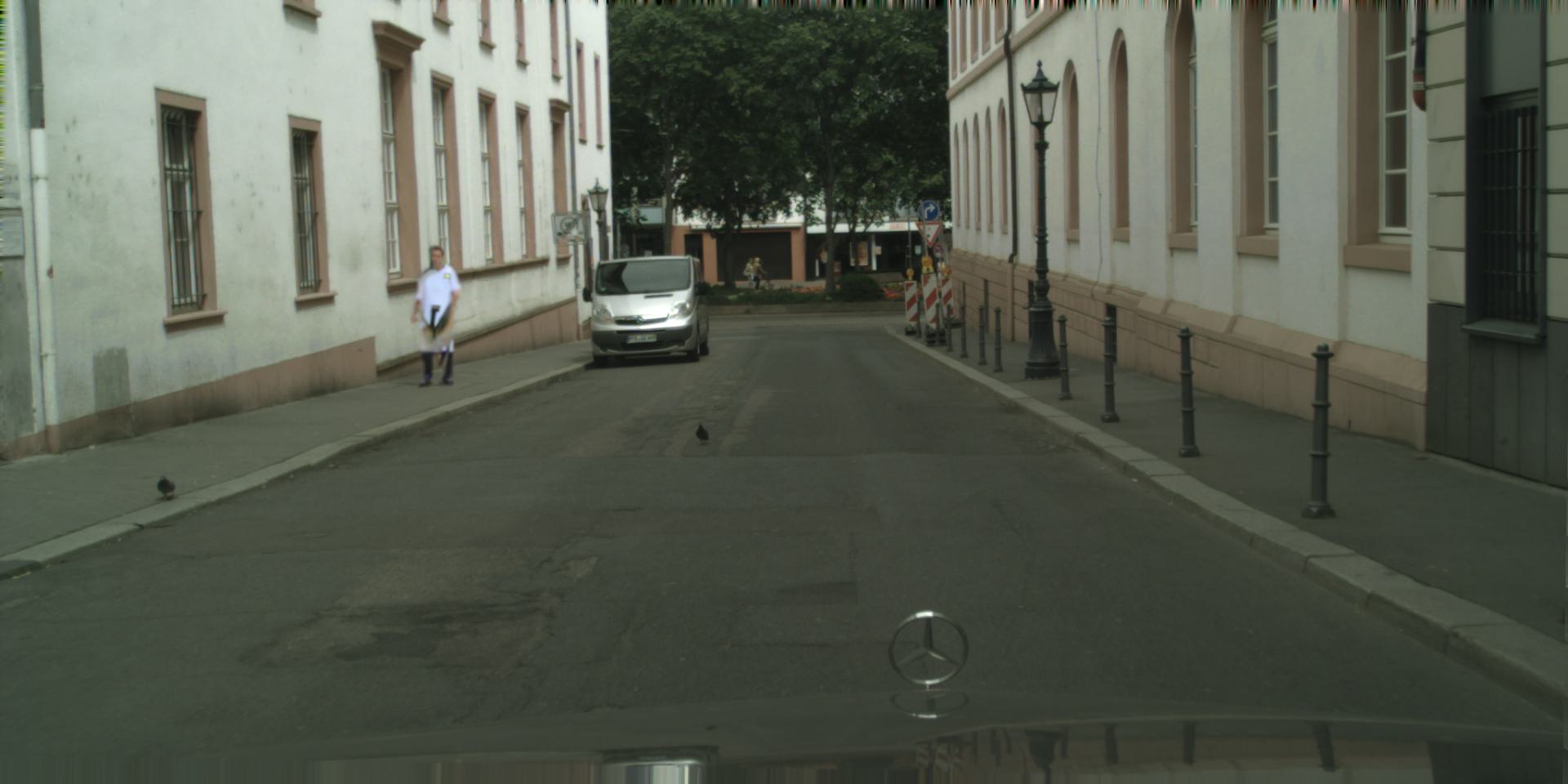}
\includegraphics[width=\textwidth]{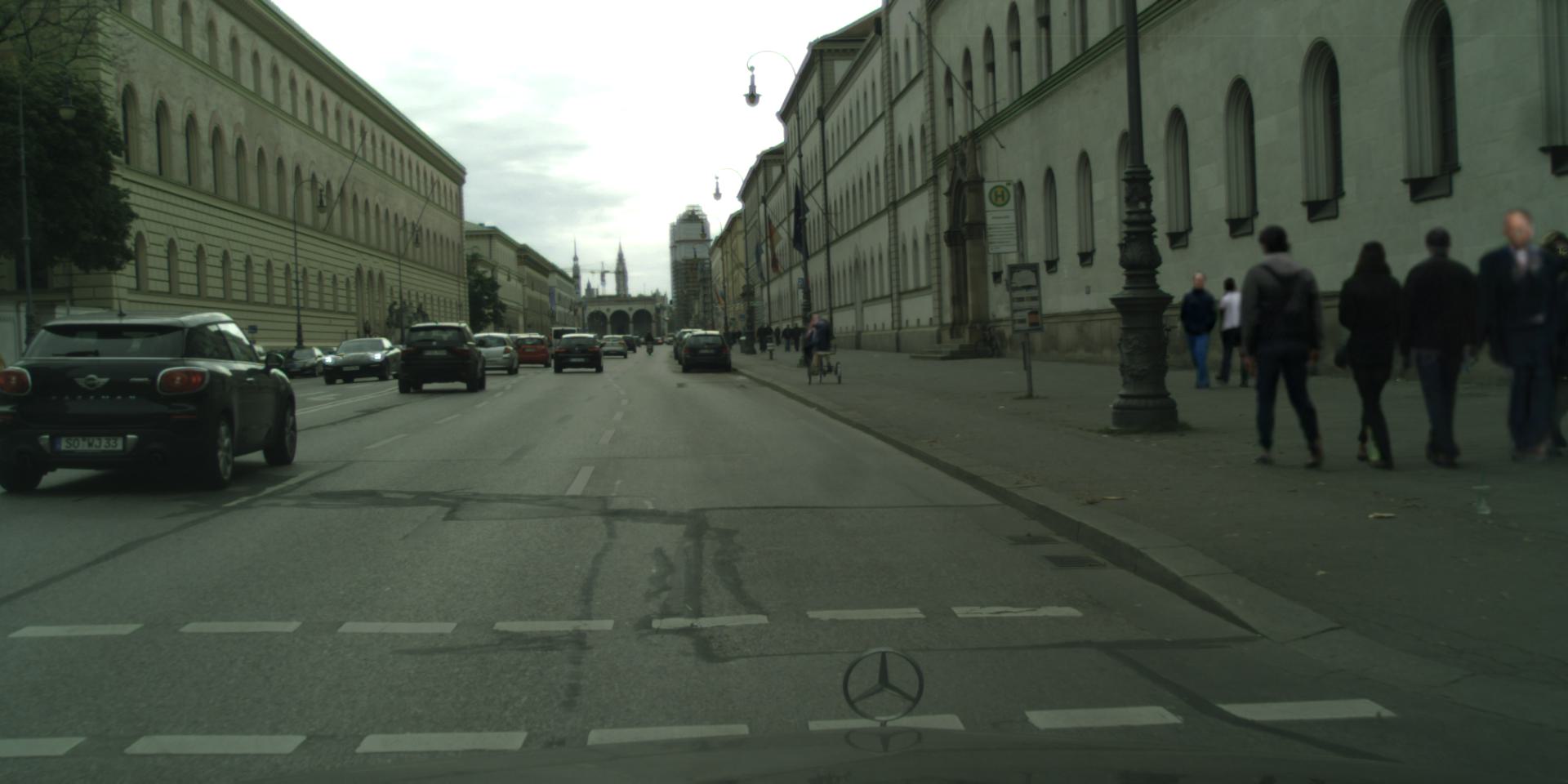}
\includegraphics[width=\textwidth]{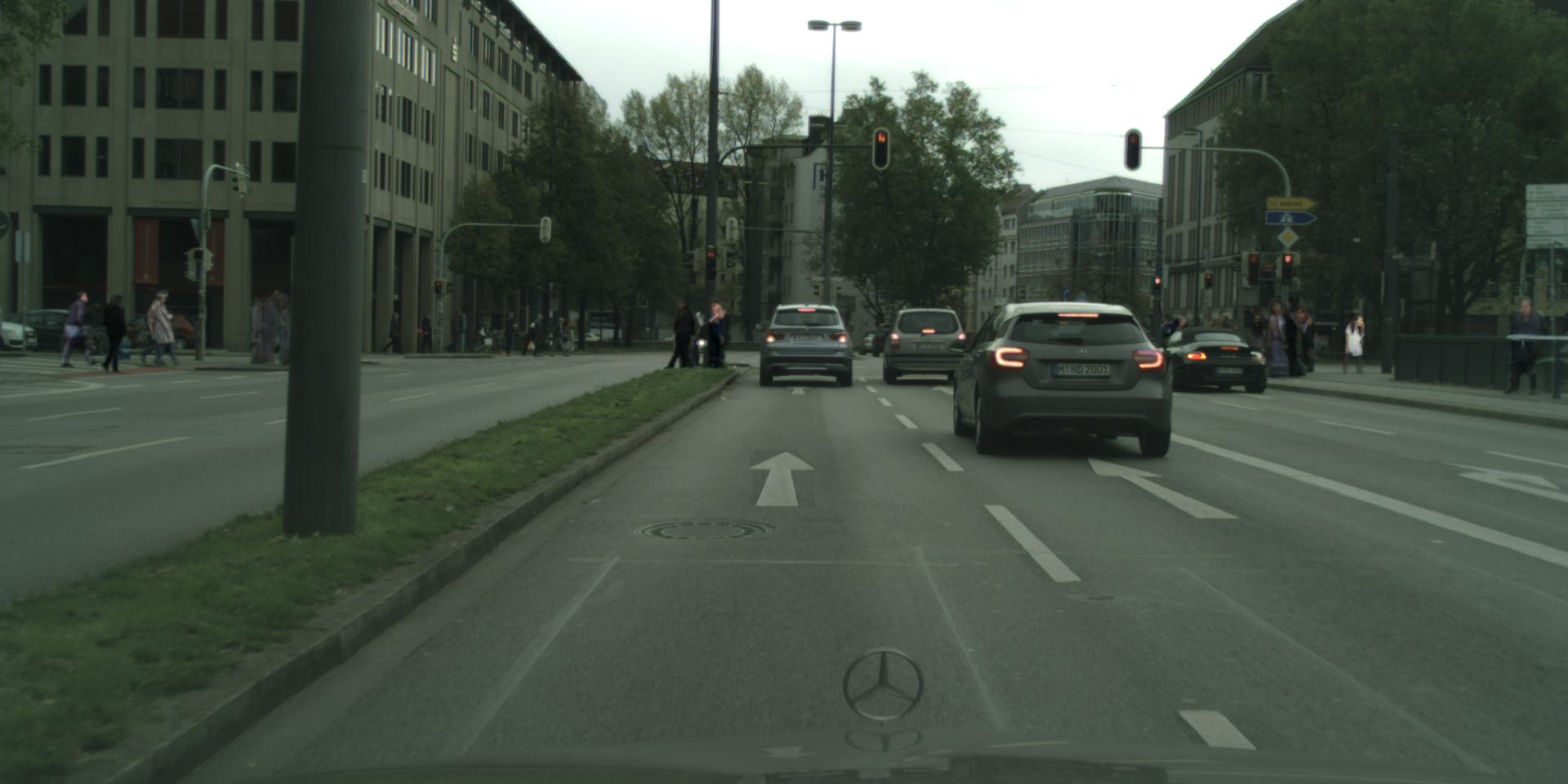}
\includegraphics[width=\textwidth]{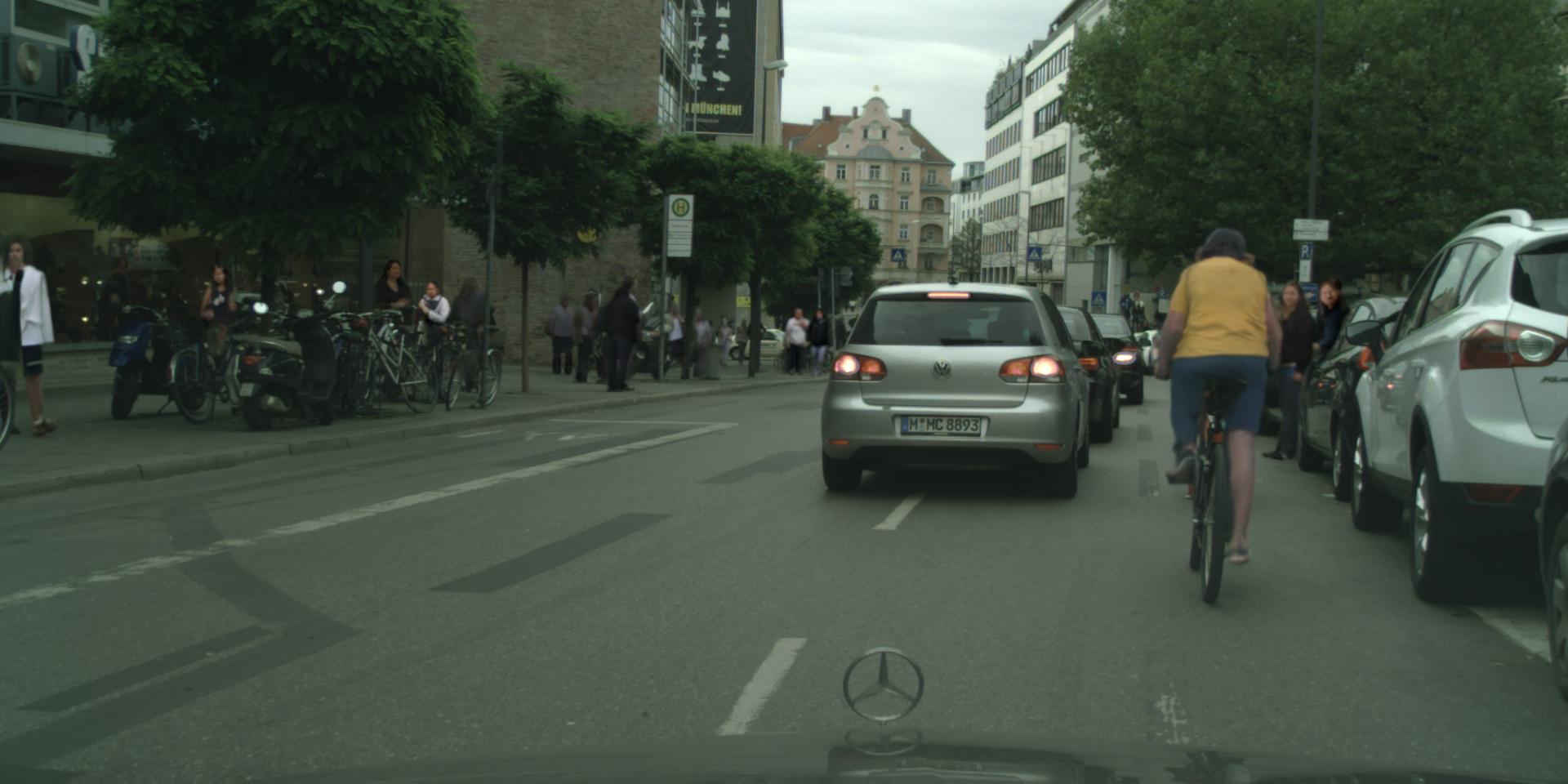}
\caption{Anonymized}
\end{subfigure}
\caption{Randomly generated images from the Cityscapes \cite{cityscapes} test set. (a) shows the original image (with blurred face to follow Cityscapes licensing), (b) the detections and (c) anonymized image}
\label{fig:citscapes_random2}
\end{figure*}

\begin{figure*}[t]
\centering
\begin{subfigure}[t]{0.3333333333333333\textwidth}
\centering
\includegraphics[width=\textwidth]{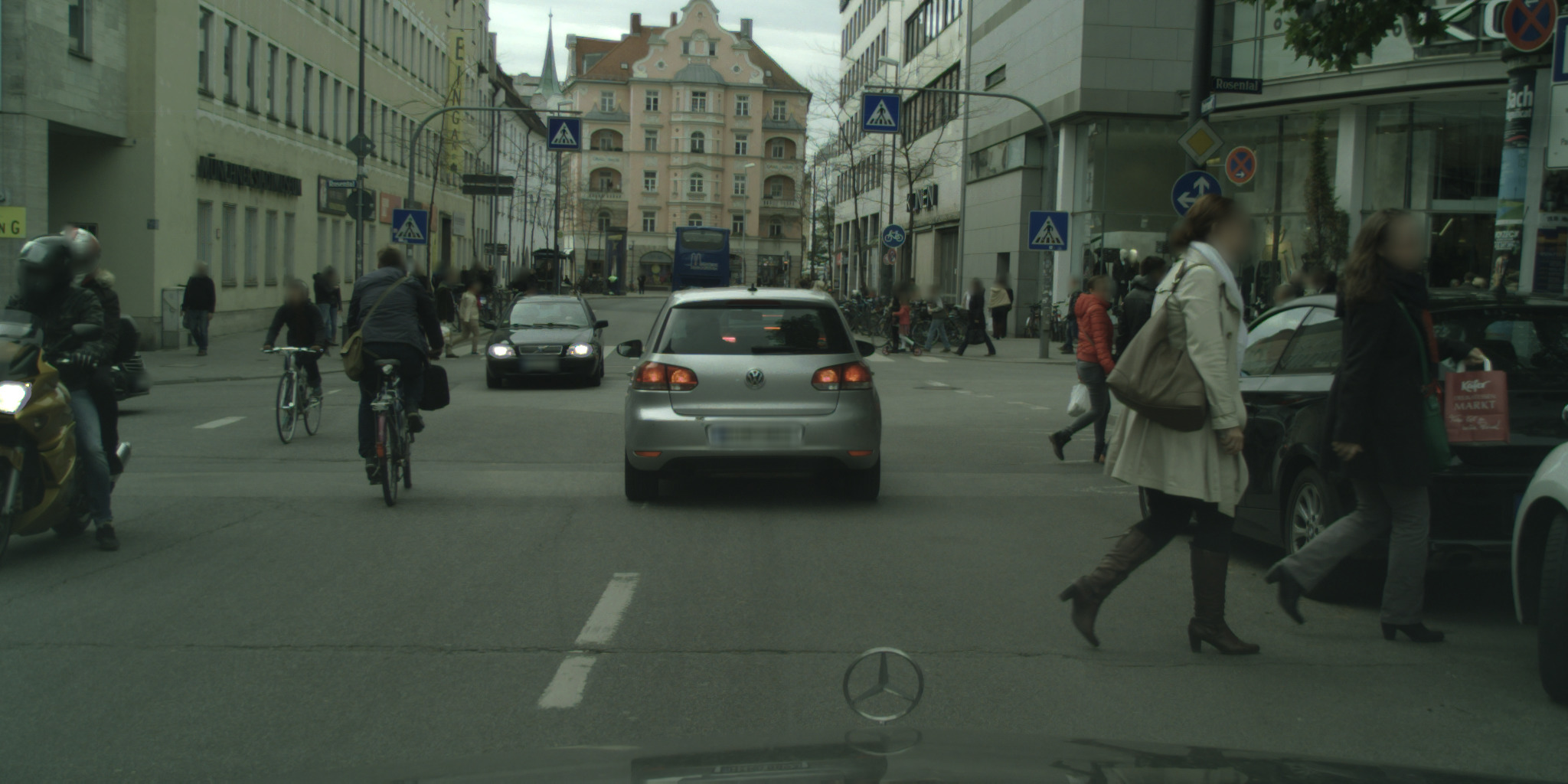}
\includegraphics[width=\textwidth]{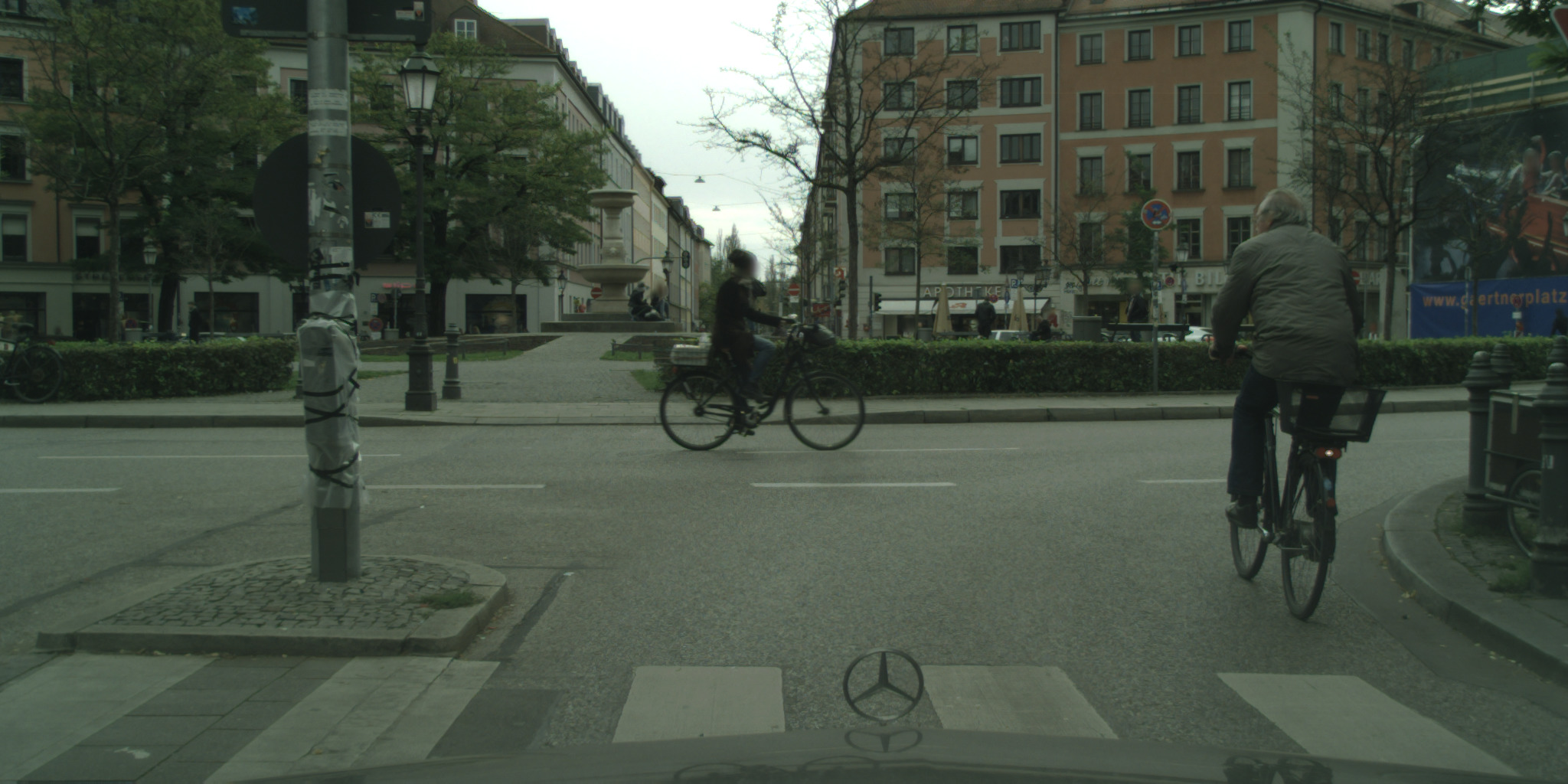}
\includegraphics[width=\textwidth]{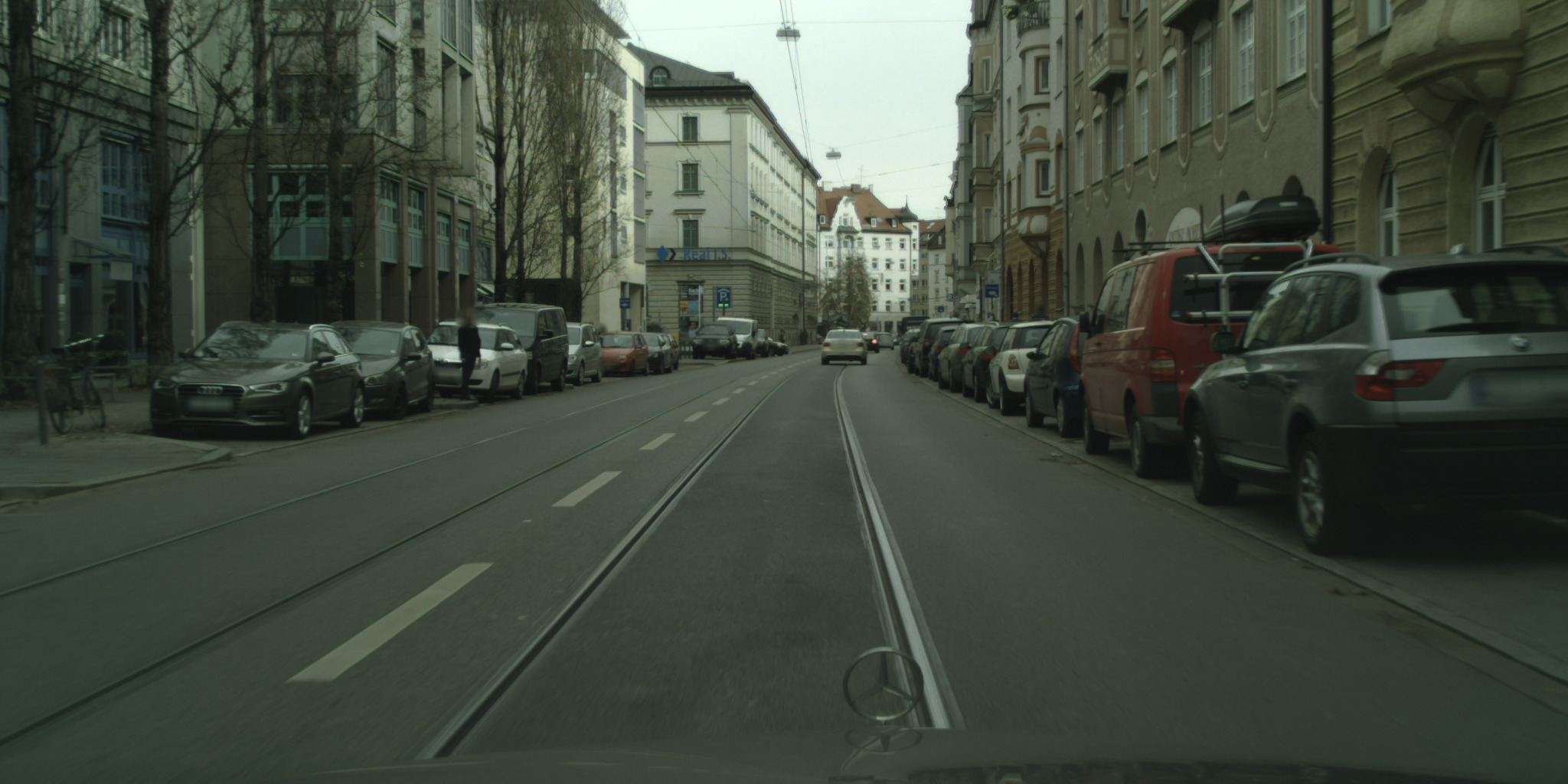}
\includegraphics[width=\textwidth]{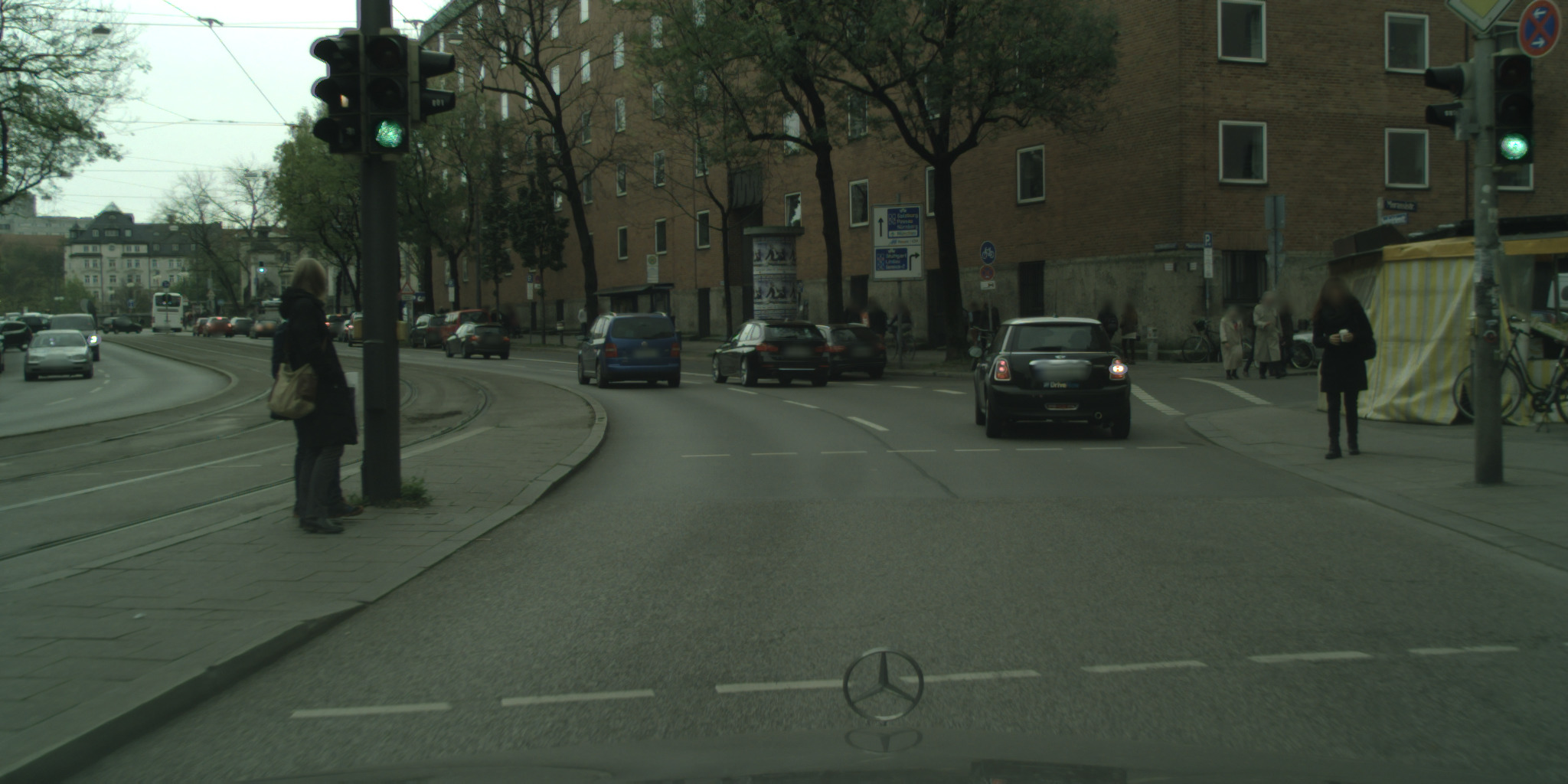}
\includegraphics[width=\textwidth]{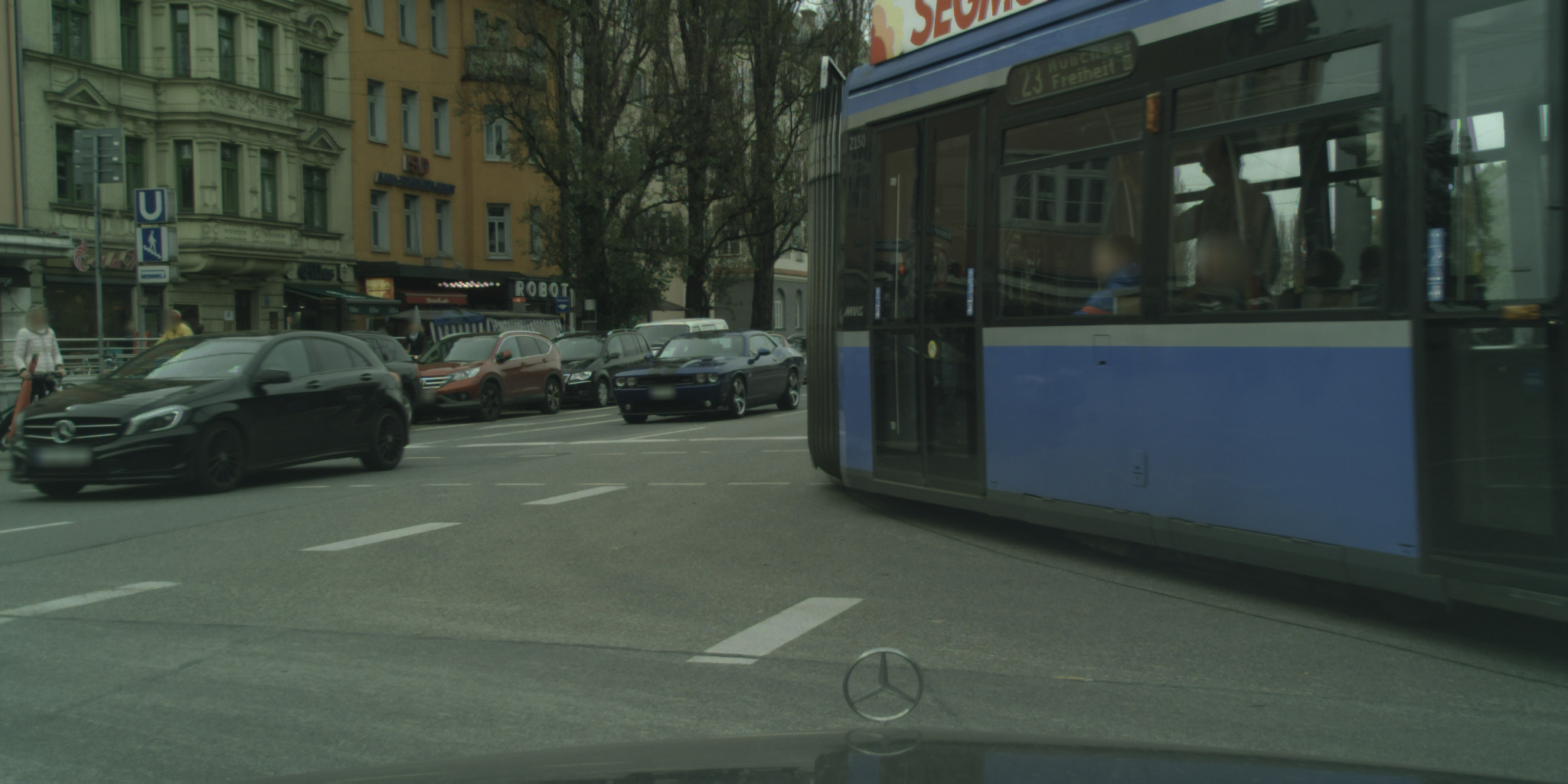}
\caption{Original}
\end{subfigure}
\begin{subfigure}[t]{0.3333333333333333\textwidth}
\centering
\includegraphics[width=\textwidth]{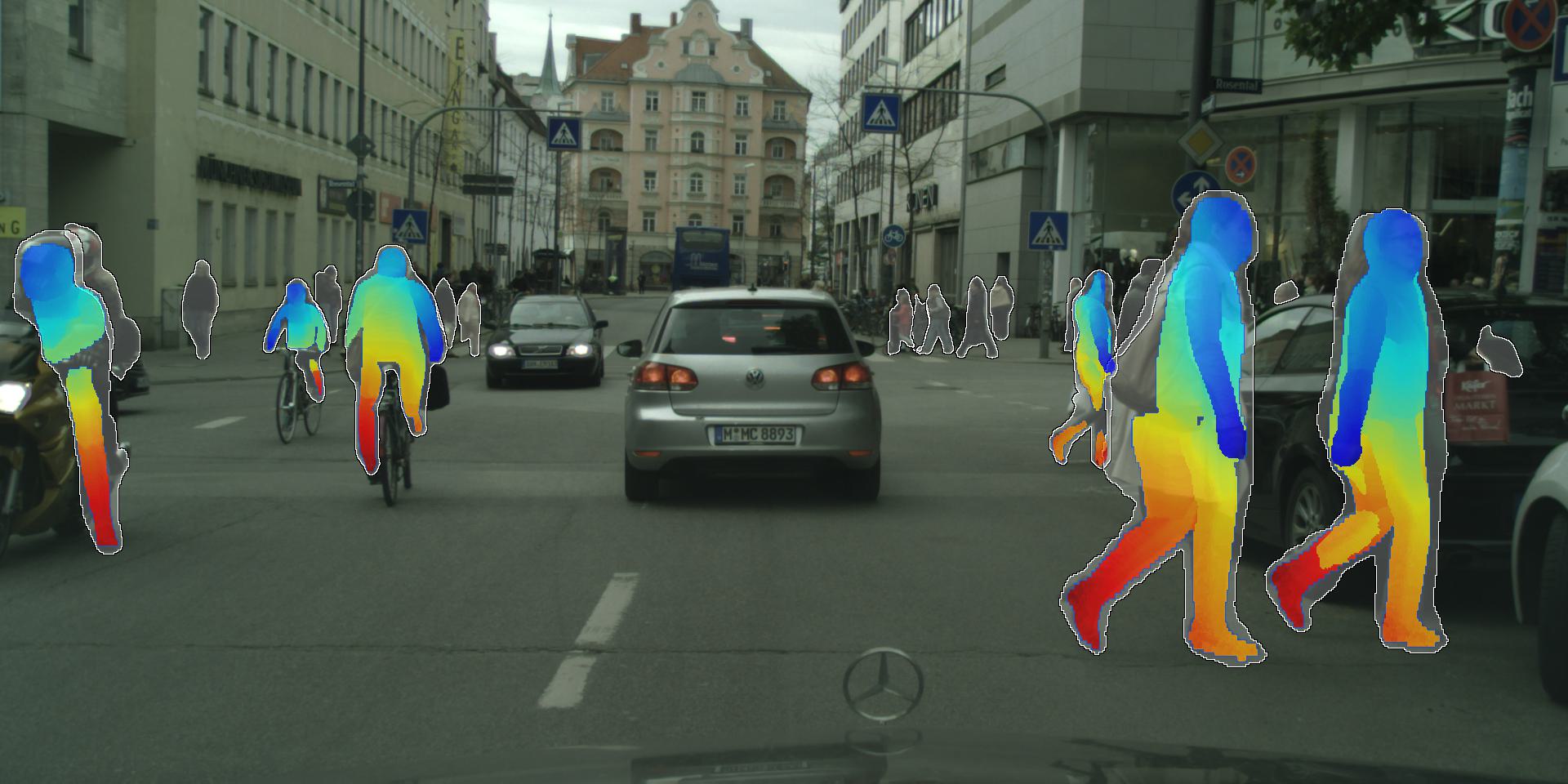}
\includegraphics[width=\textwidth]{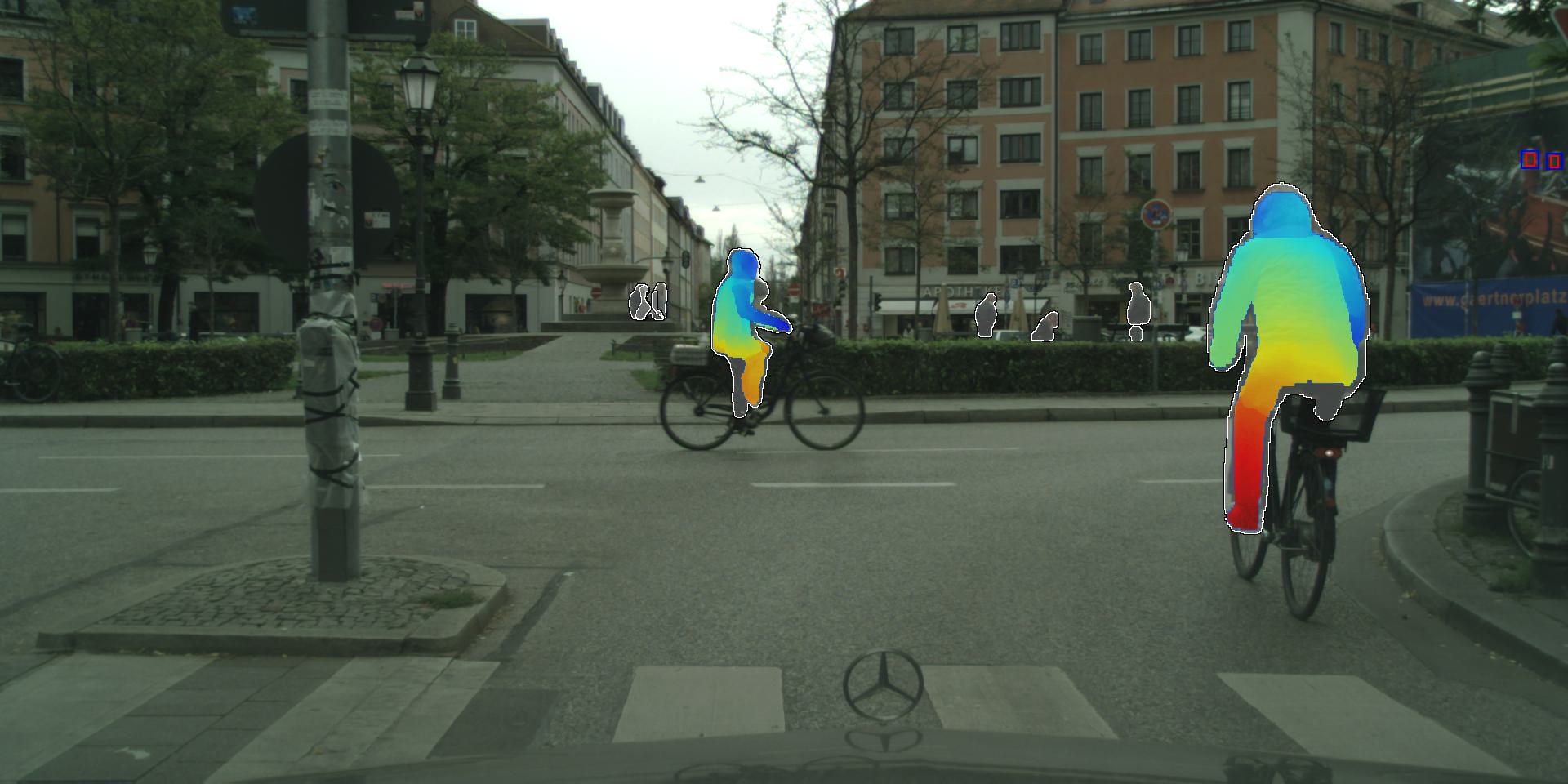}
\includegraphics[width=\textwidth]{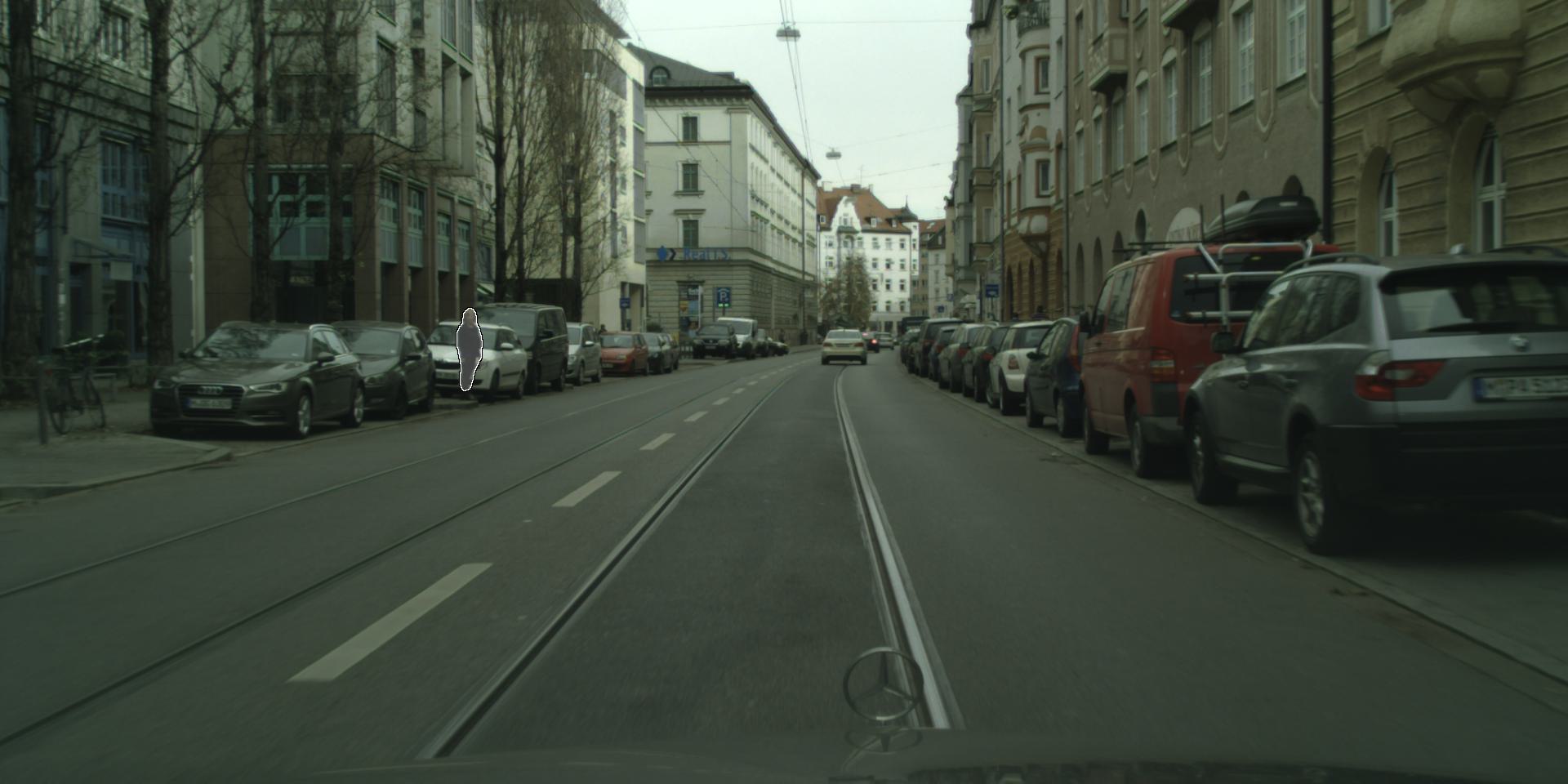}
\includegraphics[width=\textwidth]{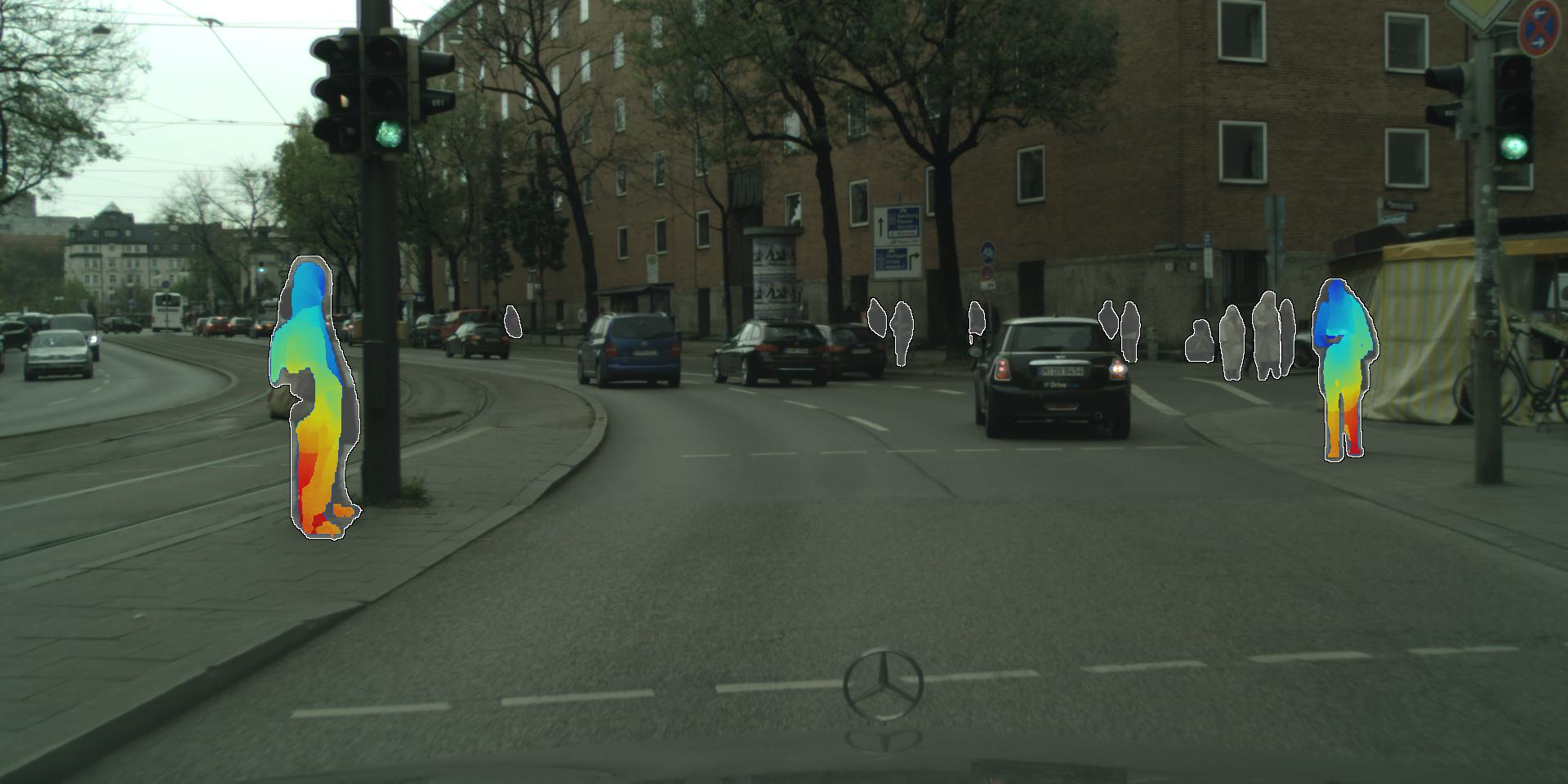}
\includegraphics[width=\textwidth]{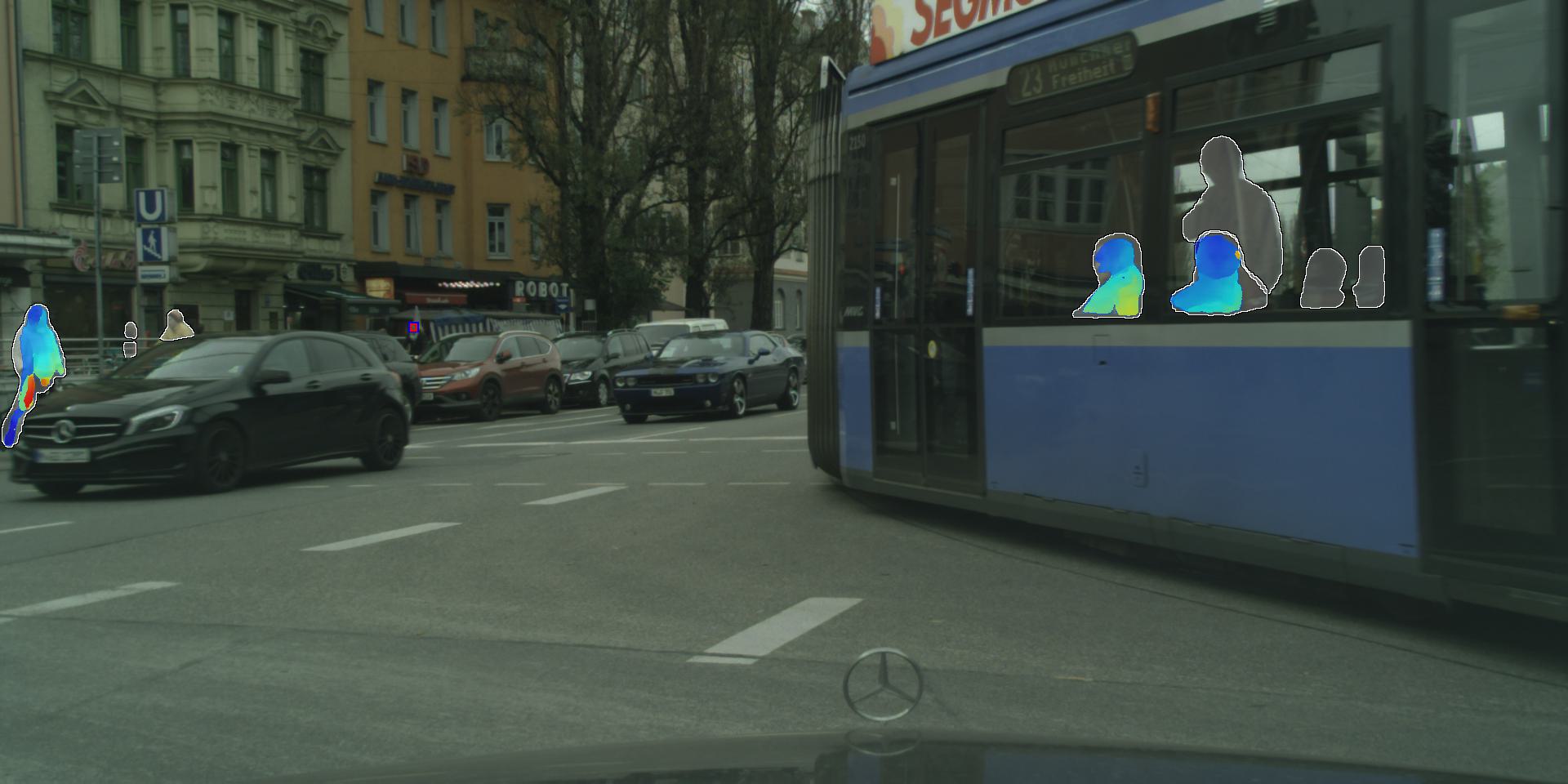}
\caption{Detections}
\end{subfigure}
\begin{subfigure}[t]{0.3333333333333333\textwidth}
\centering
\includegraphics[width=\textwidth]{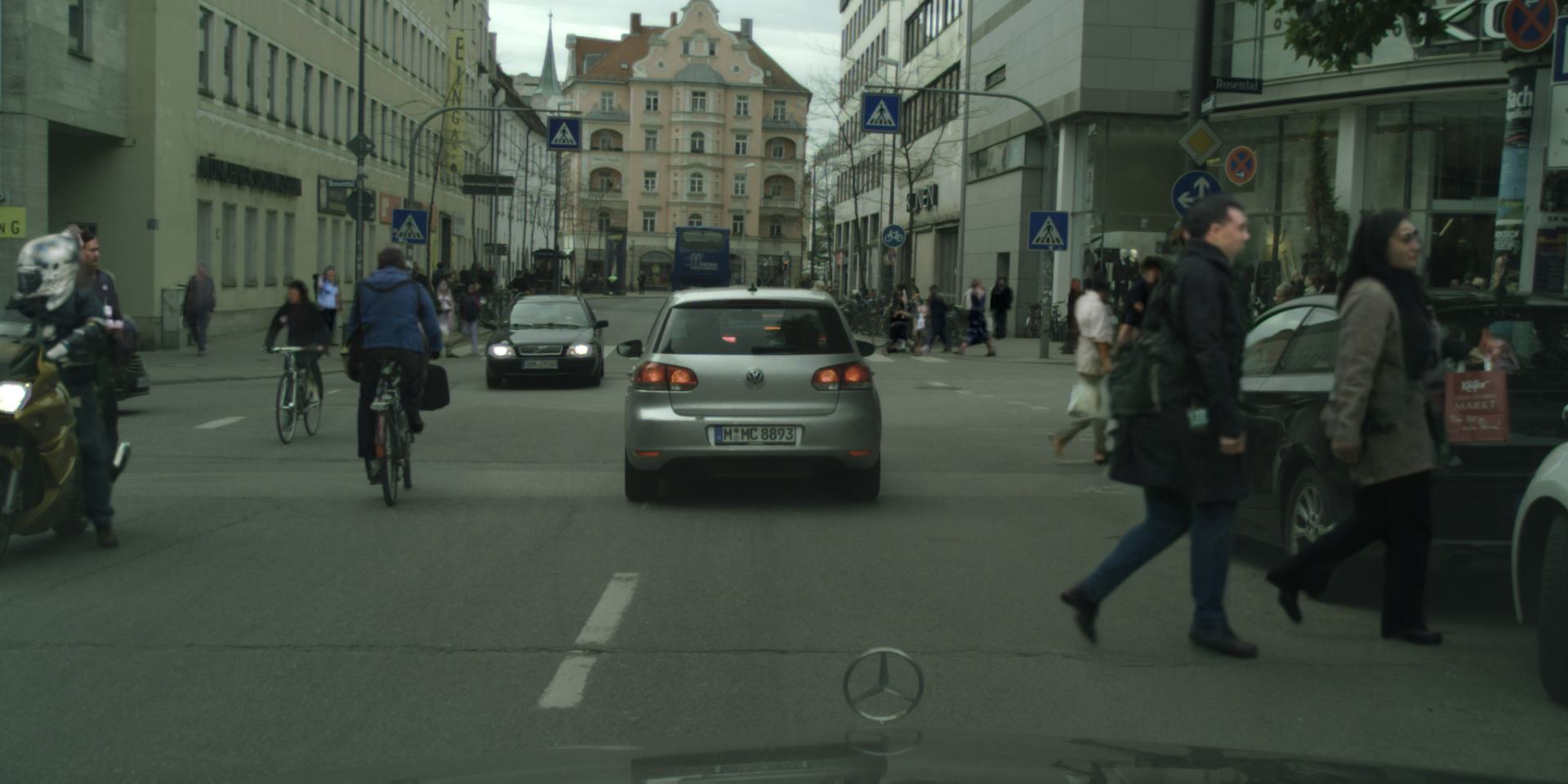}
\includegraphics[width=\textwidth]{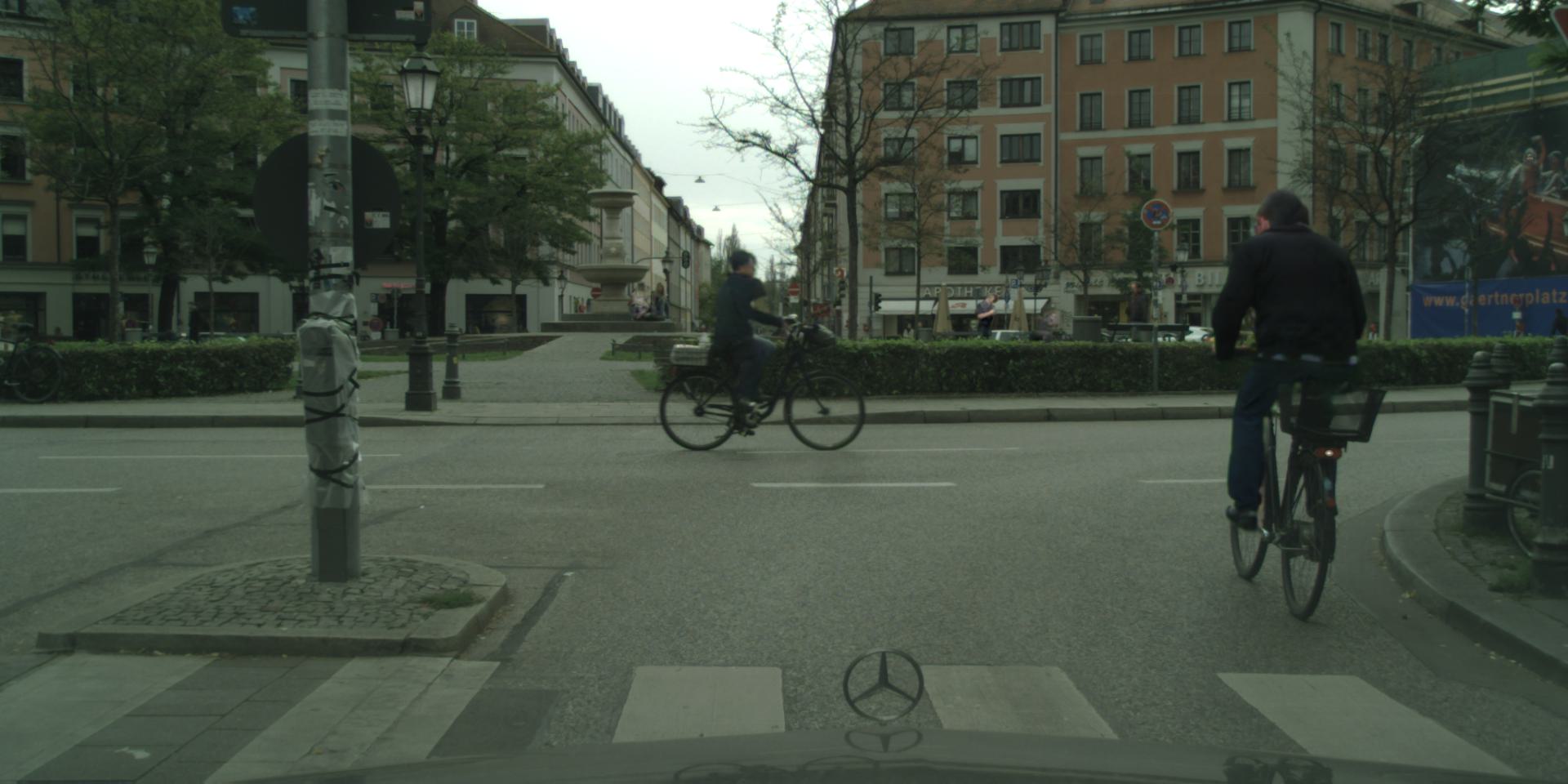}
\includegraphics[width=\textwidth]{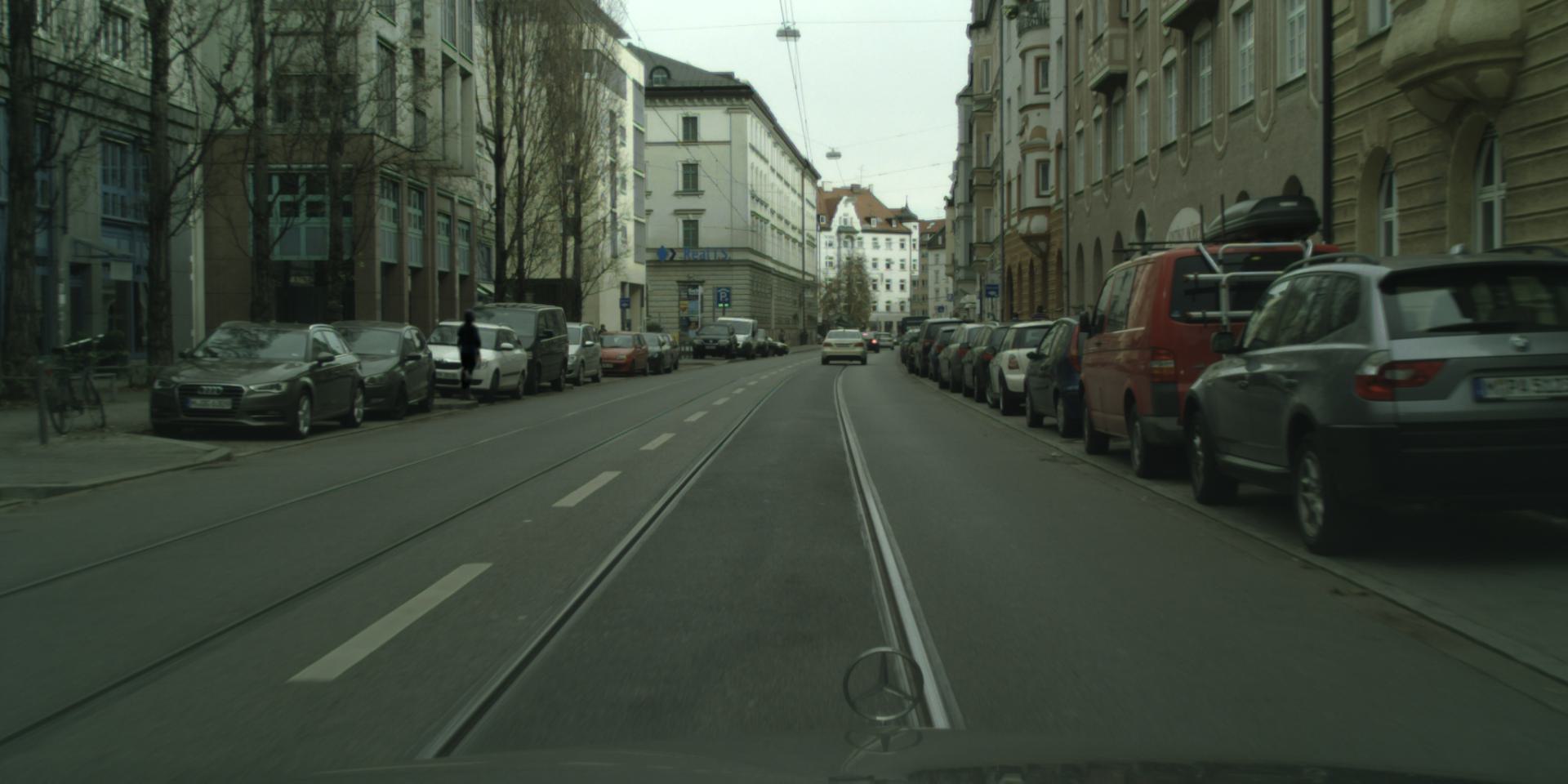}
\includegraphics[width=\textwidth]{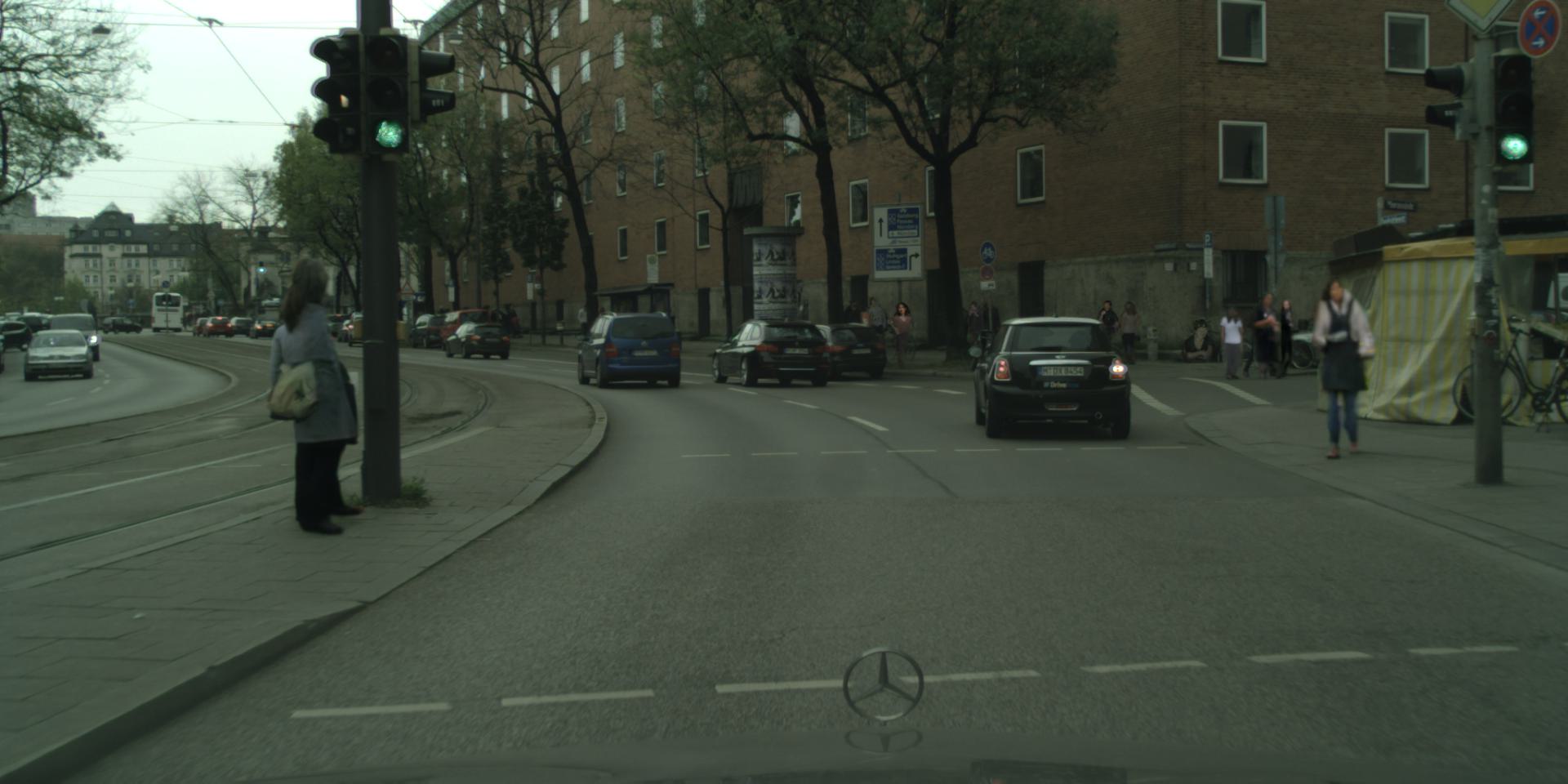}
\includegraphics[width=\textwidth]{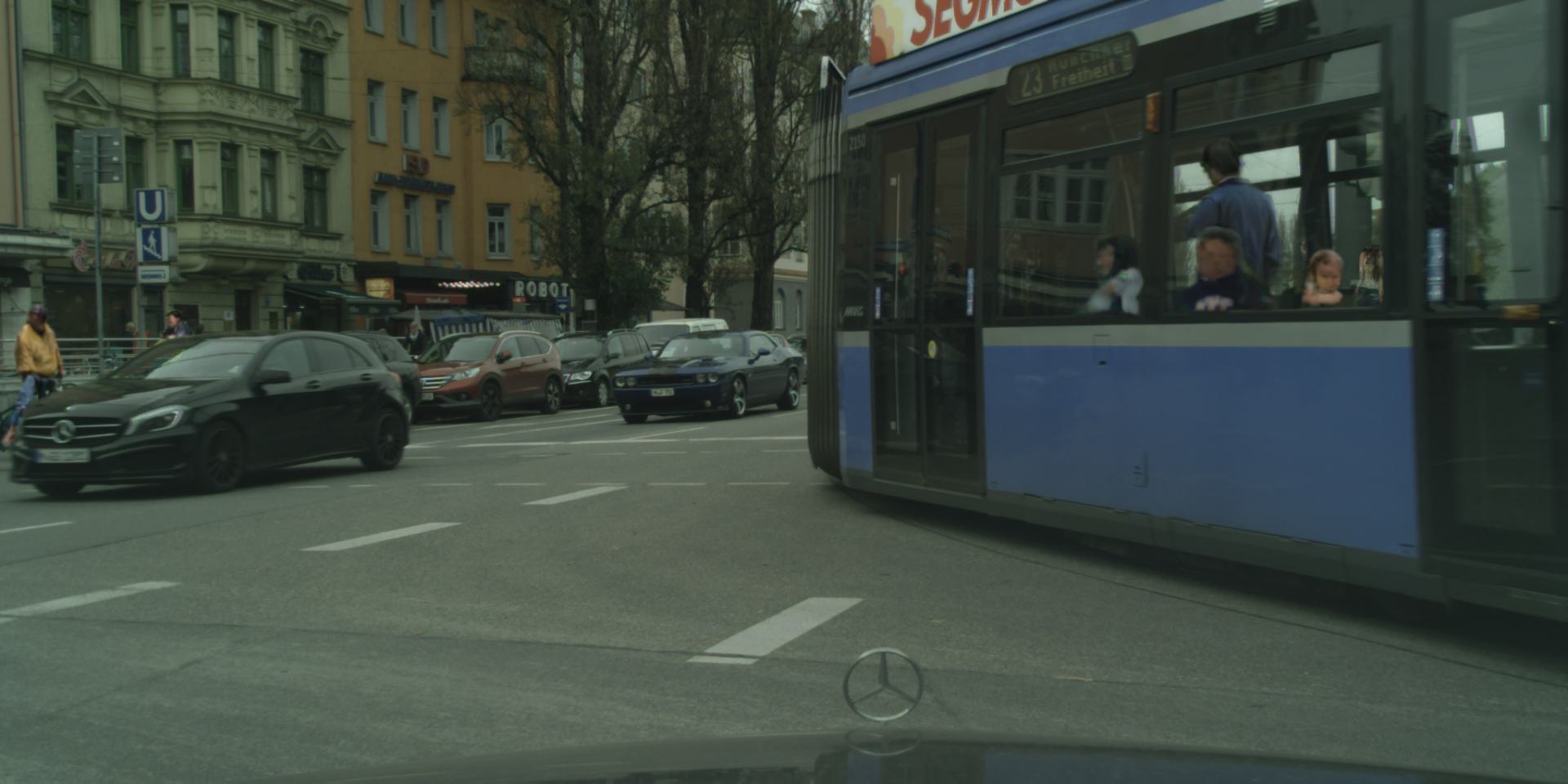}
\caption{Anonymized}
\end{subfigure}
\caption{Randomly generated images from the Cityscapes \cite{cityscapes} test set. (a) shows the original image (with blurred face to follow Cityscapes licensing), (b) the detections and (c) anonymized image}
\label{fig:citscapes_random3}
\end{figure*}

\paragraph{Cityscapes}
Randomly selected images from the Cityscapes \cite{cityscapes} dataset are shown in \Cref{fig:citscapes_random1}, \Cref{fig:citscapes_random2}, and \Cref{fig:citscapes_random3}.

\begin{figure*}[t]
\centering
\begin{subfigure}[t]{0.3333333333333333\textwidth}
\centering
\includegraphics[height=0.16666666\textheight]{generated_images/random_coco/input/000000000785.jpg}
\includegraphics[height=0.16666666\textheight]{generated_images/random_coco/input/000000144003.jpg}
\includegraphics[height=0.16666666\textheight]{generated_images/random_coco/input/000000153011.jpg}
\includegraphics[height=0.16666666\textheight]{generated_images/random_coco/input/000000164602.jpg}
\includegraphics[height=0.16666666\textheight]{generated_images/random_coco/input/000000186449.jpg}
\includegraphics[height=0.16666666\textheight]{generated_images/random_coco/input/000000249786.jpg}
\caption{Original}
\end{subfigure}
\begin{subfigure}[t]{0.3333333333333333\textwidth}
\centering
\includegraphics[height=0.16666666\textheight]{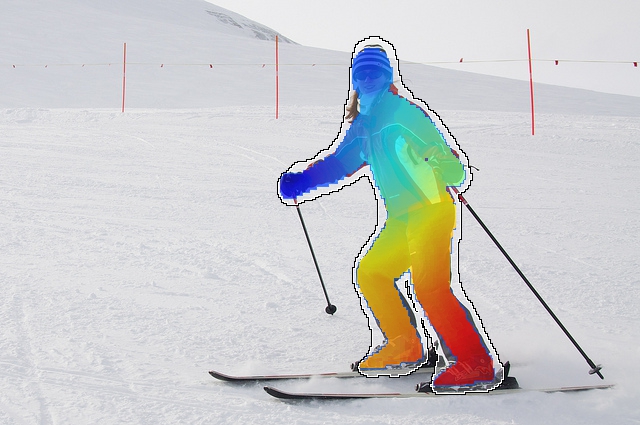}
\includegraphics[height=0.16666666\textheight]{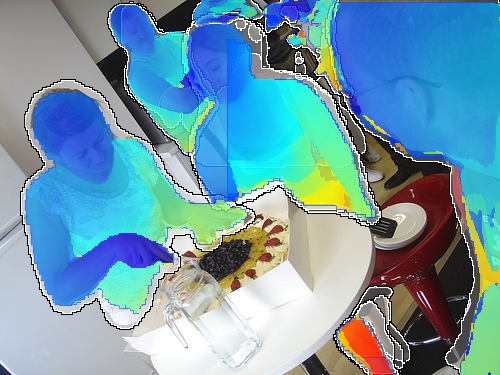}
\includegraphics[height=0.16666666\textheight]{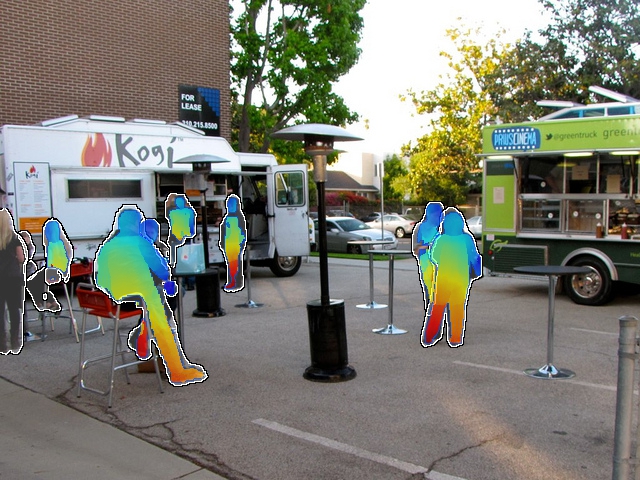}
\includegraphics[height=0.16666666\textheight]{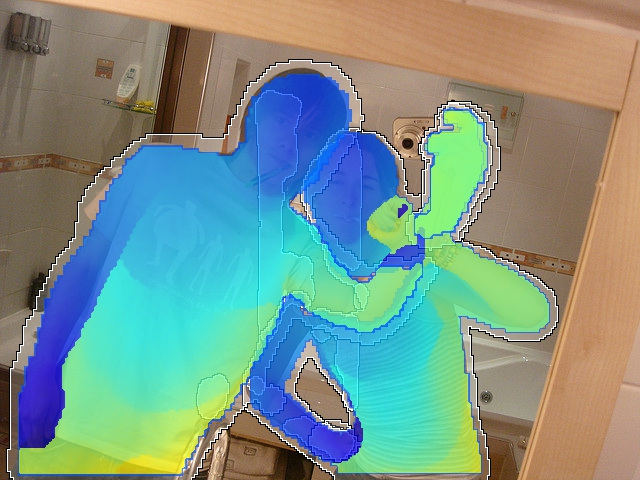}
\includegraphics[height=0.16666666\textheight]{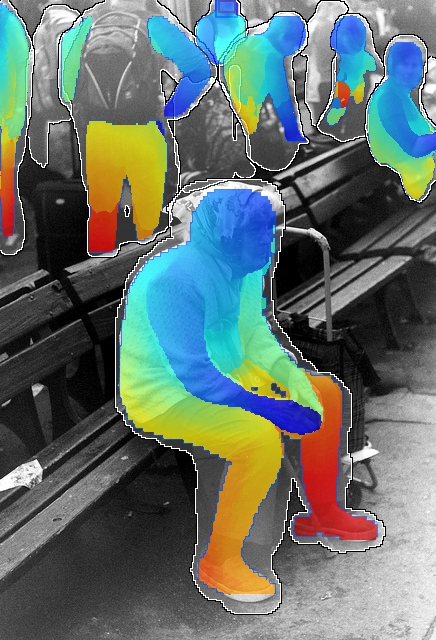}
\includegraphics[height=0.16666666\textheight]{generated_images/random_coco/detections/000000249786.jpg}
\caption{Detections}
\end{subfigure}
\begin{subfigure}[t]{0.3333333333333333\textwidth}
\centering
\includegraphics[height=0.16666666\textheight]{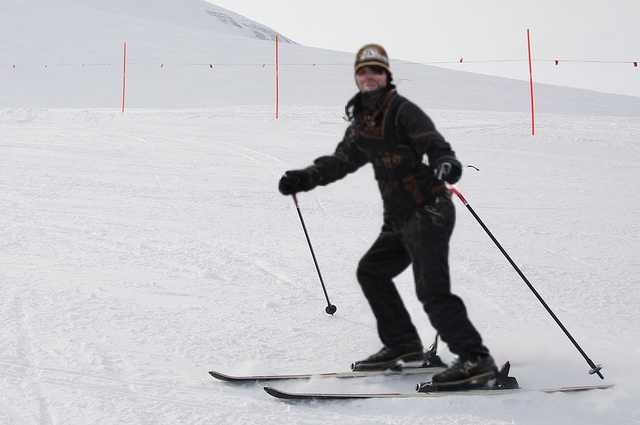}
\includegraphics[height=0.16666666\textheight]{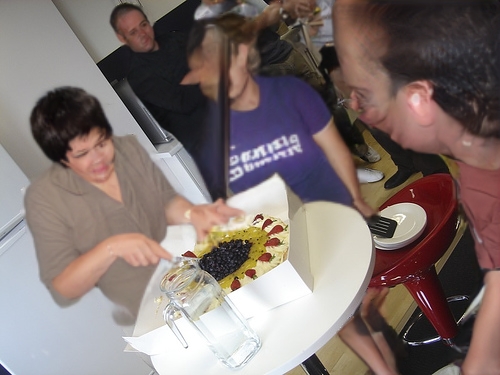}
\includegraphics[height=0.16666666\textheight]{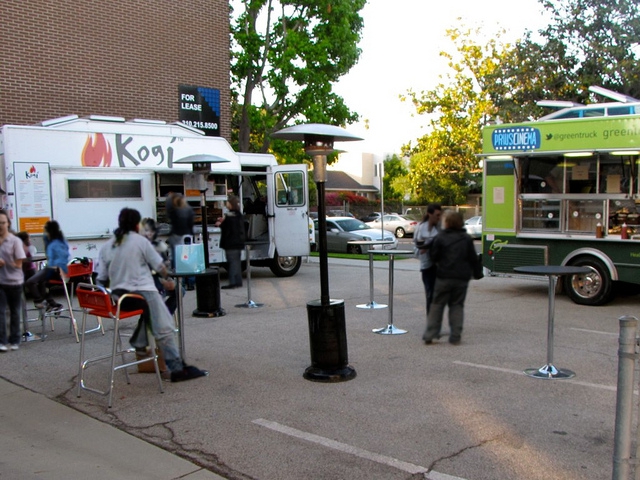}
\includegraphics[height=0.16666666\textheight]{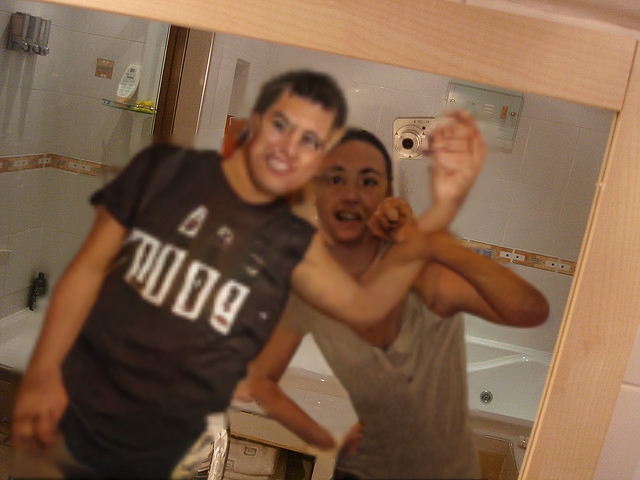}
\includegraphics[height=0.16666666\textheight]{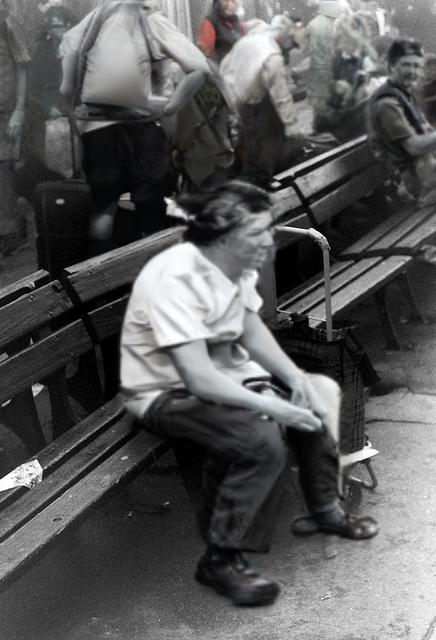}
\includegraphics[height=0.16666666\textheight]{generated_images/random_coco/anonymized/000000249786.jpg}
\caption{Anonymized}
\end{subfigure}
\caption{Randomly generated images from COCO \cite{Lin2014COCO} val2017. (a) shows the original image, (b) the detections and (c) anonymized image}
\label{fig:random_coco1}
\end{figure*}

\begin{figure*}[t]
\centering
\begin{subfigure}[t]{0.3333333333333333\textwidth}
\centering
\includegraphics[height=0.15\textheight]{generated_images/random_coco/input/000000273493.jpg}
\includegraphics[height=0.15\textheight]{generated_images/random_coco/input/000000276024.jpg}
\includegraphics[height=0.15\textheight]{generated_images/random_coco/input/000000286660.jpg}
\includegraphics[height=0.15\textheight]{generated_images/random_coco/input/000000291791.jpg}
\includegraphics[height=0.15\textheight]{generated_images/random_coco/input/000000357737.jpg}
\includegraphics[height=0.15\textheight]{generated_images/random_coco/input/000000439522.jpg}
\caption{Original}
\end{subfigure}
\begin{subfigure}[t]{0.3333333333333333\textwidth}
\centering
\includegraphics[height=0.15\textheight]{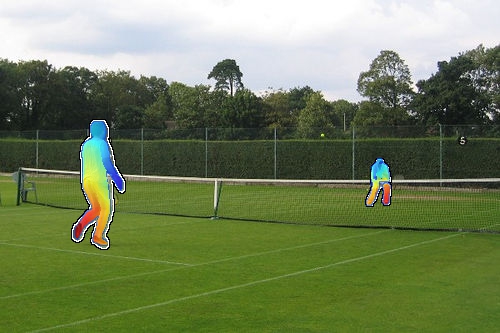}
\includegraphics[height=0.15\textheight]{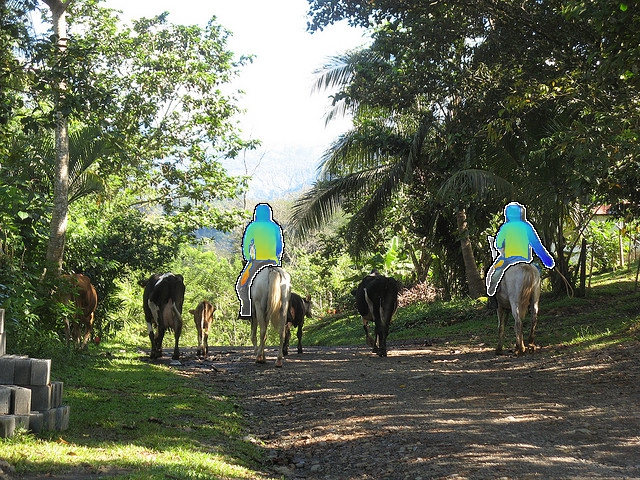}
\includegraphics[height=0.15\textheight]{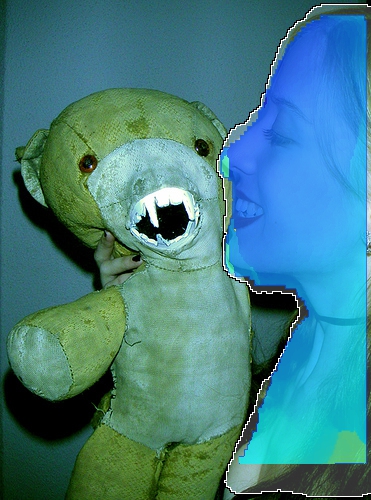}
\includegraphics[height=0.15\textheight]{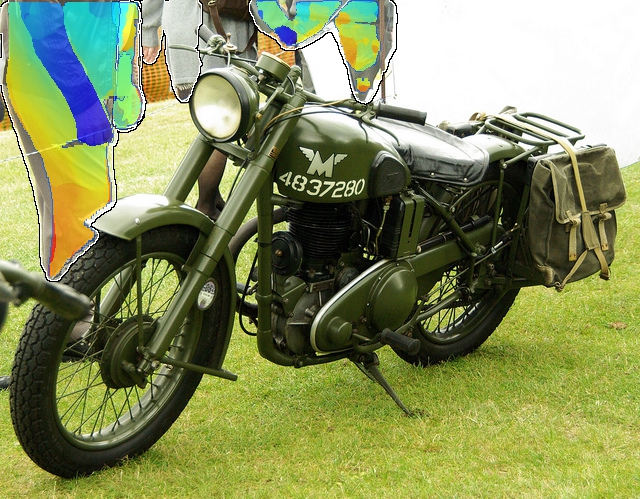}
\includegraphics[height=0.15\textheight]{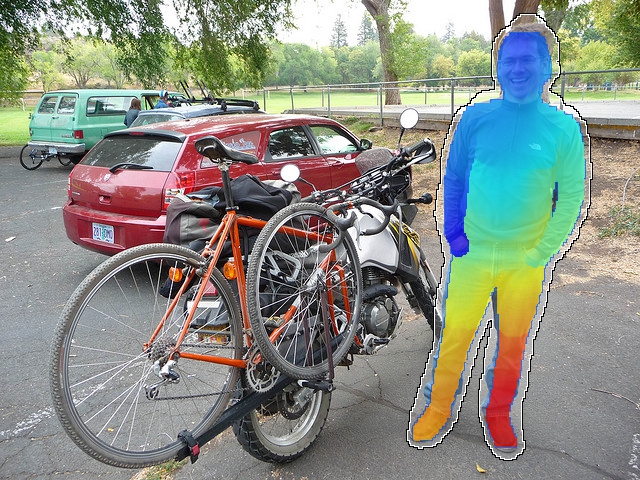}
\includegraphics[height=0.15\textheight]{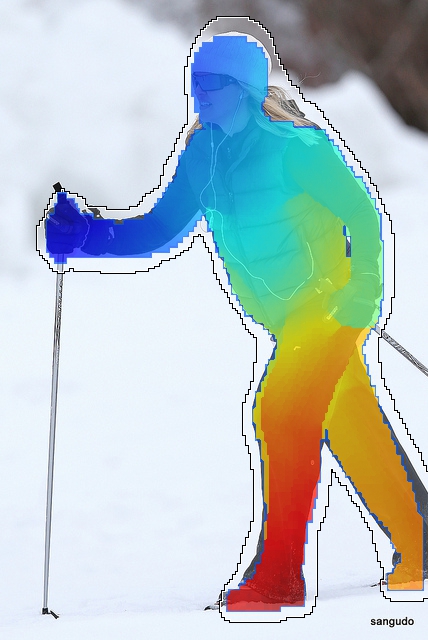}
\caption{Detections}
\end{subfigure}
\begin{subfigure}[t]{0.3333333333333333\textwidth}
\centering
\includegraphics[height=0.15\textheight]{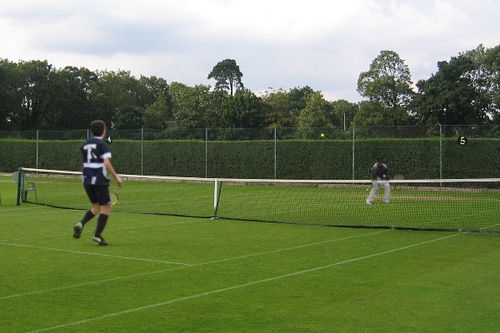}
\includegraphics[height=0.15\textheight]{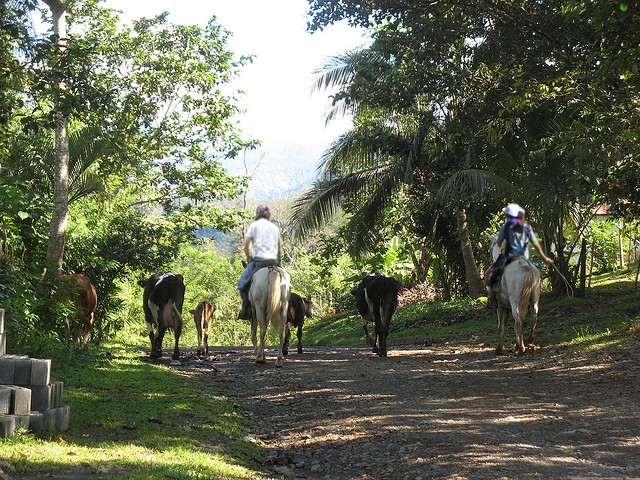}
\includegraphics[height=0.15\textheight]{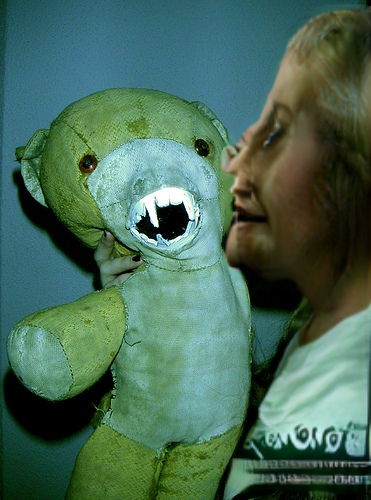}
\includegraphics[height=0.15\textheight]{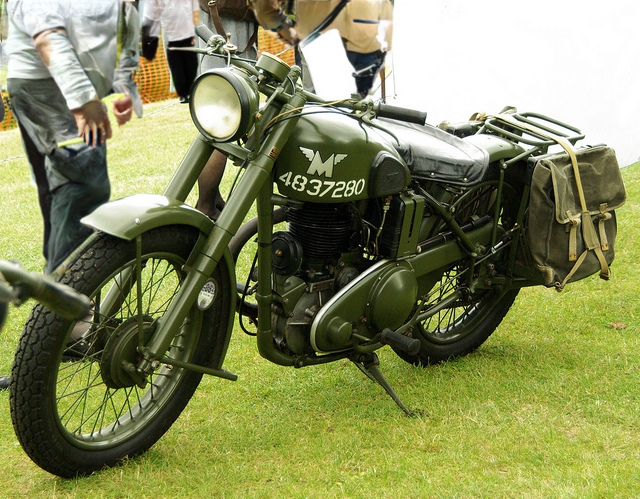}
\includegraphics[height=0.15\textheight]{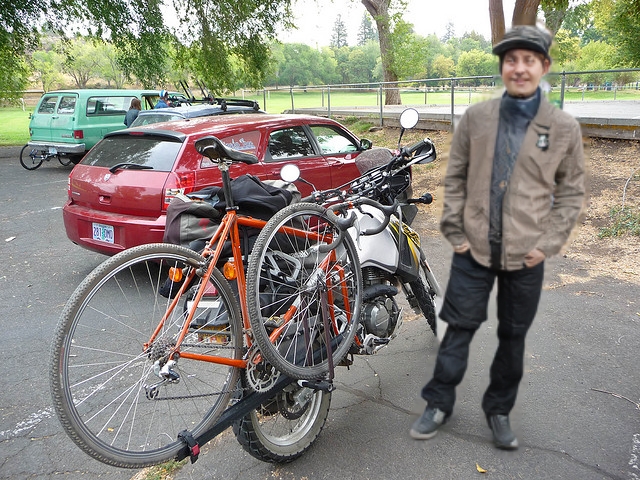}
\includegraphics[height=0.15\textheight]{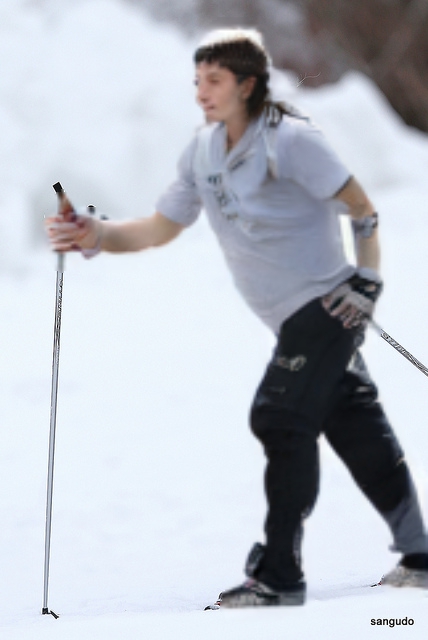}
\caption{Anonymized}
\end{subfigure}
\caption{Randomly generated images from COCO \cite{Lin2014COCO} val2017. (a) shows the original image, (b) the detections and (c) anonymized image}
\label{fig:random_coco2}
\end{figure*}

\begin{figure*}[t]
\centering
\begin{subfigure}[t]{0.3333333333333333\textwidth}
\centering
\includegraphics[height=0.15\textheight]{generated_images/random_coco/input/000000440336.jpg}
\includegraphics[height=0.15\textheight]{generated_images/random_coco/input/000000455157.jpg}
\includegraphics[height=0.15\textheight]{generated_images/random_coco/input/000000465179.jpg}
\includegraphics[height=0.15\textheight]{generated_images/random_coco/input/000000476215.jpg}
\includegraphics[height=0.15\textheight]{generated_images/random_coco/input/000000500565.jpg}
\includegraphics[height=0.15\textheight]{generated_images/random_coco/input/000000504415.jpg}
\caption{Original}
\end{subfigure}
\begin{subfigure}[t]{0.3333333333333333\textwidth}
\centering
\includegraphics[height=0.15\textheight]{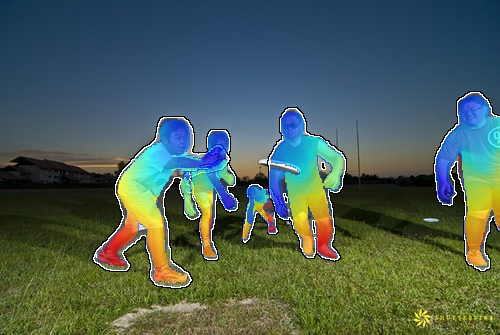}
\includegraphics[height=0.15\textheight]{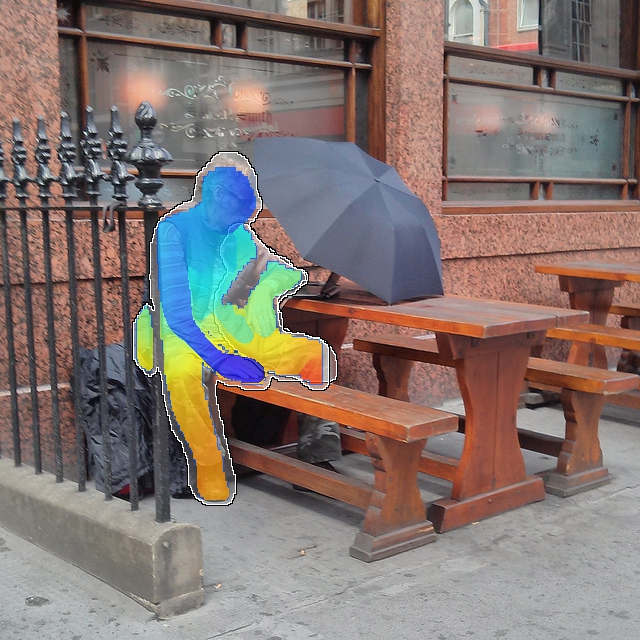}
\includegraphics[height=0.15\textheight]{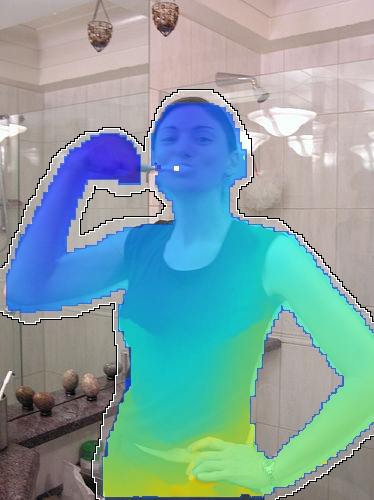}
\includegraphics[height=0.15\textheight]{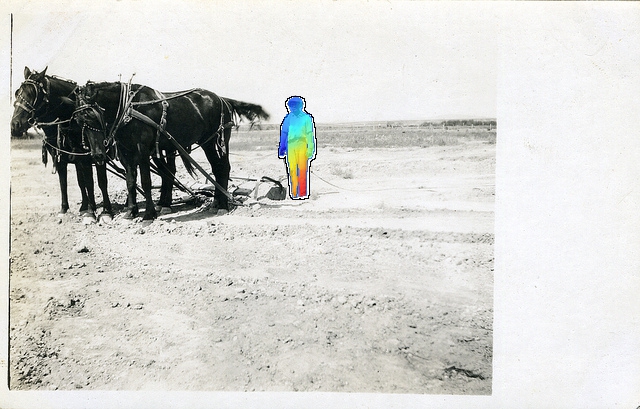}
\includegraphics[height=0.15\textheight]{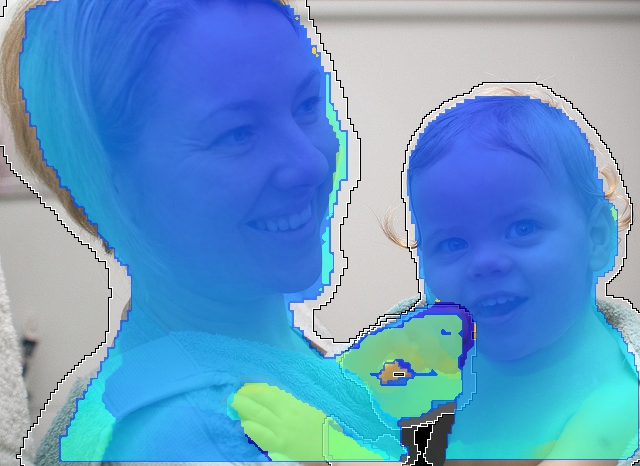}
\includegraphics[height=0.15\textheight]{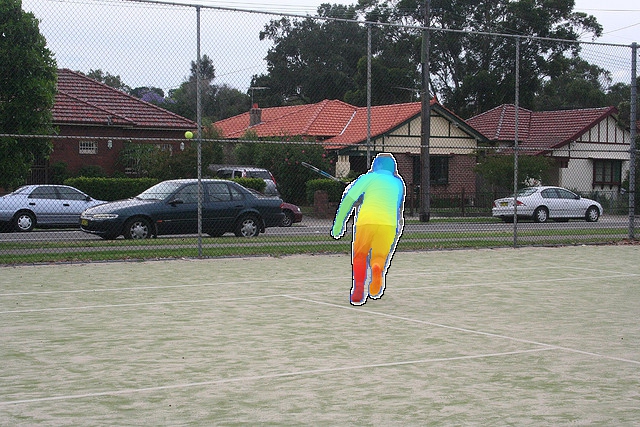}
\caption{Detections}
\end{subfigure}
\begin{subfigure}[t]{0.3333333333333333\textwidth}
\centering
\includegraphics[height=0.15\textheight]{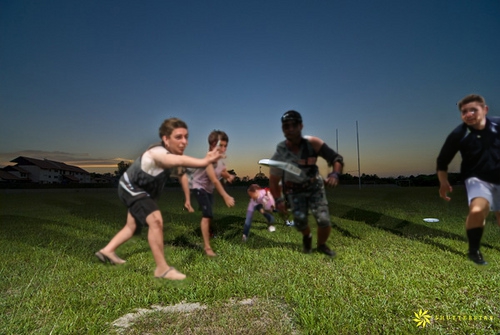}
\includegraphics[height=0.15\textheight]{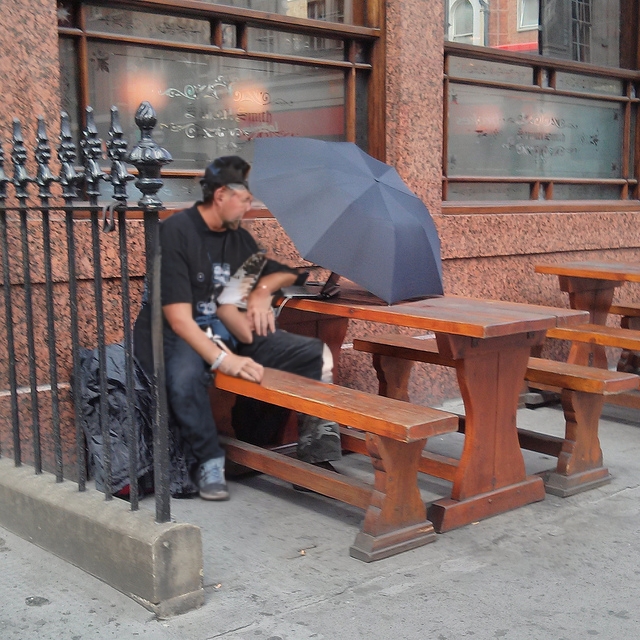}
\includegraphics[height=0.15\textheight]{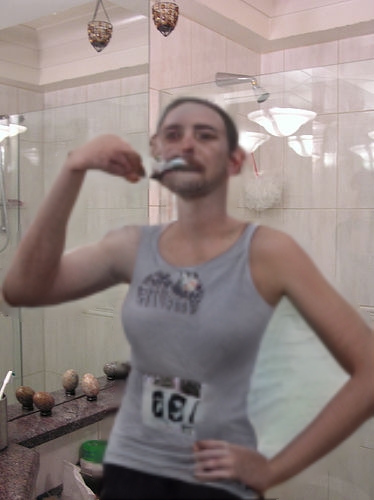}
\includegraphics[height=0.15\textheight]{generated_images/random_coco/anonymized/000000476215.jpg}
\includegraphics[height=0.15\textheight]{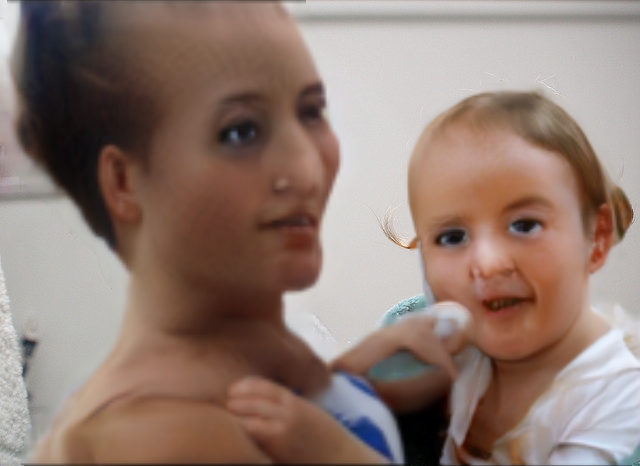}
\includegraphics[height=0.15\textheight]{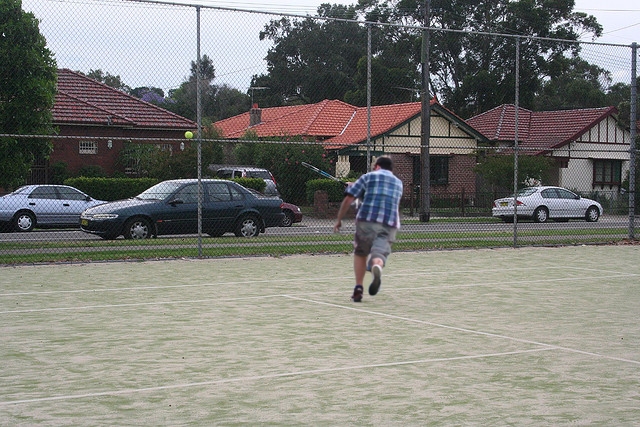}
\caption{Anonymized}
\end{subfigure}
\caption{Randomly generated images from COCO \cite{Lin2014COCO} val2017. (a) shows the original image, (b) the detections and (c) anonymized image}
\label{fig:random_coco3}
\end{figure*}

\paragraph{COCO}
Randomly selected images from the COCO \cite{Lin2014COCO} dataset are shown in \Cref{fig:random_coco1}, \Cref{fig:random_coco2}, \Cref{fig:random_coco3}.

\begin{figure*}[t]
\centering
\begin{subfigure}[t]{0.14285714285714285\textwidth}
\centering
\includegraphics[height=0.15\textheight]{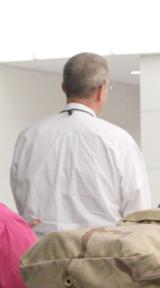}
\includegraphics[height=0.15\textheight]{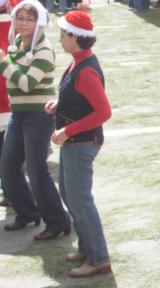}
\includegraphics[height=0.15\textheight]{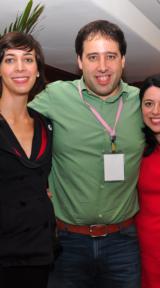}
\includegraphics[height=0.15\textheight]{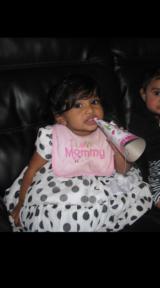}
\includegraphics[height=0.15\textheight]{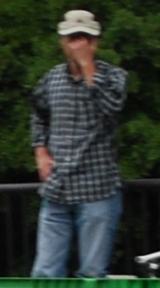}
\includegraphics[height=0.15\textheight]{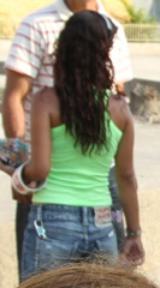}
\caption{}
\end{subfigure}
\begin{subfigure}[t]{0.14285714285714285\textwidth}
\centering
\includegraphics[height=0.15\textheight]{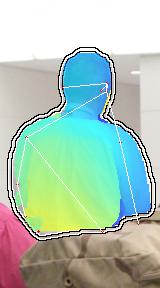}
\includegraphics[height=0.15\textheight]{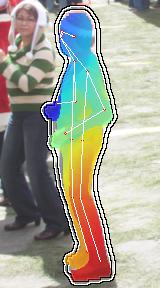}
\includegraphics[height=0.15\textheight]{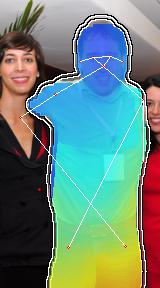}
\includegraphics[height=0.15\textheight]{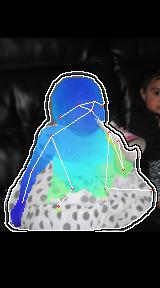}
\includegraphics[height=0.15\textheight]{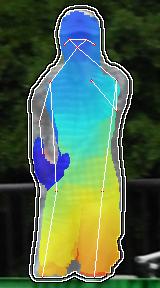}
\includegraphics[height=0.15\textheight]{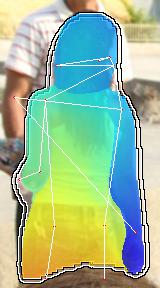}
\caption{}
\end{subfigure}
\begin{subfigure}[t]{0.14285714285714285\textwidth}
\centering
\includegraphics[height=0.15\textheight]{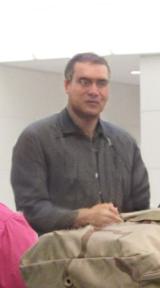}
\includegraphics[height=0.15\textheight]{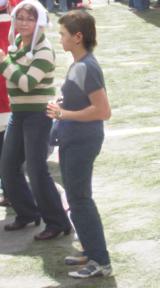}
\includegraphics[height=0.15\textheight]{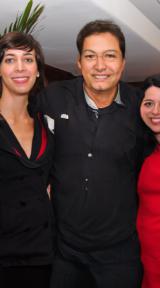}
\includegraphics[height=0.15\textheight]{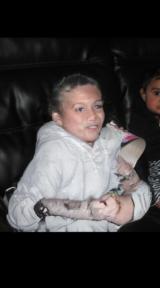}
\includegraphics[height=0.15\textheight]{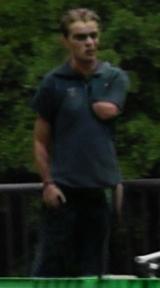}
\includegraphics[height=0.15\textheight]{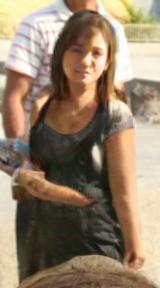}
\caption{}
\end{subfigure}
\begin{subfigure}[t]{0.14285714285714285\textwidth}
\centering
\includegraphics[height=0.15\textheight]{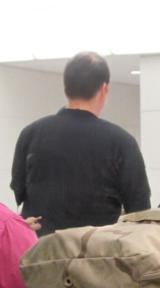}
\includegraphics[height=0.15\textheight]{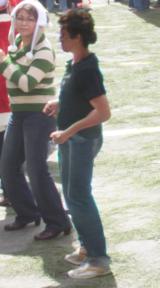}
\includegraphics[height=0.15\textheight]{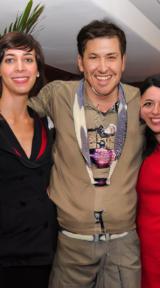}
\includegraphics[height=0.15\textheight]{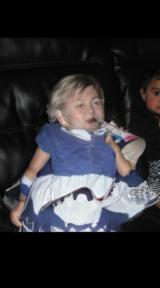}
\includegraphics[height=0.15\textheight]{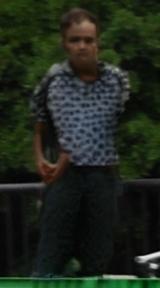}
\includegraphics[height=0.15\textheight]{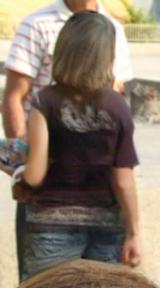}
\caption{}
\end{subfigure}
\begin{subfigure}[t]{0.14285714285714285\textwidth}
\centering
\includegraphics[height=0.15\textheight]{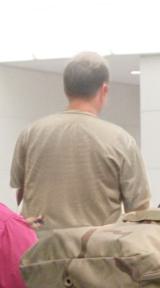}
\includegraphics[height=0.15\textheight]{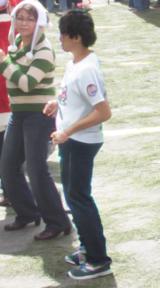}
\includegraphics[height=0.15\textheight]{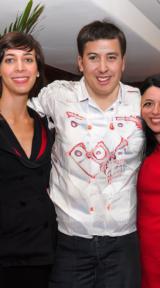}
\includegraphics[height=0.15\textheight]{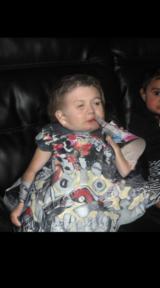}
\includegraphics[height=0.15\textheight]{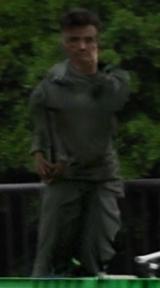}
\includegraphics[height=0.15\textheight]{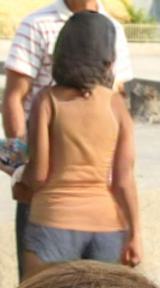}
\caption{}
\end{subfigure}
\begin{subfigure}[t]{0.14285714285714285\textwidth}
\centering
\includegraphics[height=0.15\textheight]{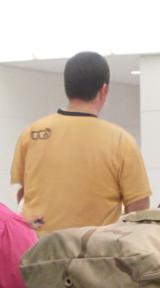}
\includegraphics[height=0.15\textheight]{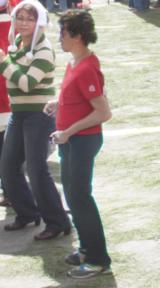}
\includegraphics[height=0.15\textheight]{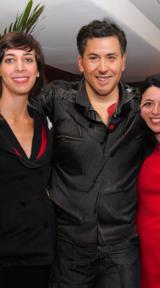}
\includegraphics[height=0.15\textheight]{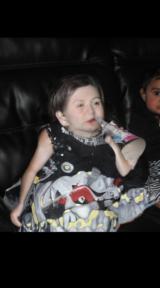}
\includegraphics[height=0.15\textheight]{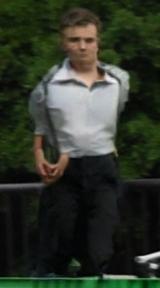}
\includegraphics[height=0.15\textheight]{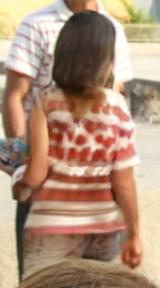}
\caption{}
\end{subfigure}
\begin{subfigure}[t]{0.14285714285714285\textwidth}
\centering
\includegraphics[height=0.15\textheight]{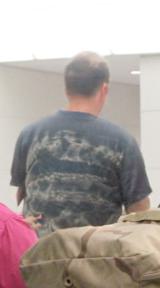}
\includegraphics[height=0.15\textheight]{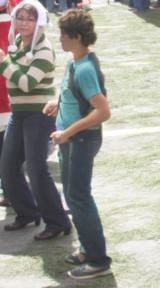}
\includegraphics[height=0.15\textheight]{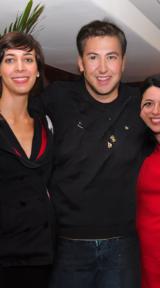}
\includegraphics[height=0.15\textheight]{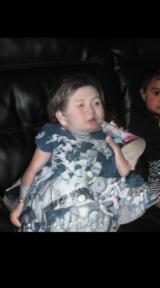}
\includegraphics[height=0.15\textheight]{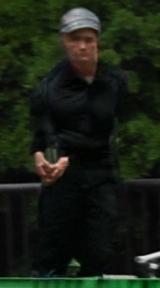}
\includegraphics[height=0.15\textheight]{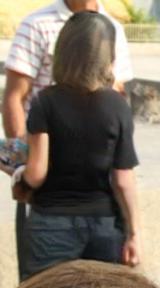}
\caption{}
\end{subfigure}
\caption{Randomly generated images from the FDH dataset. (a) shows the original image, (b) is the conditional input, (c) is the generated result for the unconditional generator, and (c-g) are from the CSE-guided generator.}
\label{fig:random_fdh1}
\end{figure*}

\begin{figure*}[t]
\centering
\begin{subfigure}[t]{0.14285714285714285\textwidth}
\centering
\includegraphics[height=0.15\textheight]{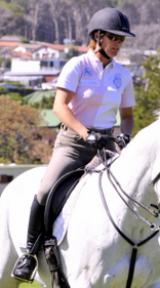}
\includegraphics[height=0.15\textheight]{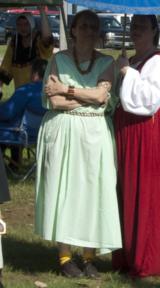}
\includegraphics[height=0.15\textheight]{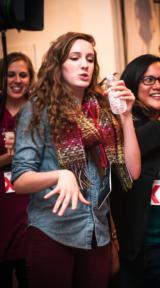}
\includegraphics[height=0.15\textheight]{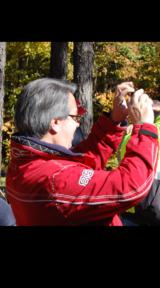}
\includegraphics[height=0.15\textheight]{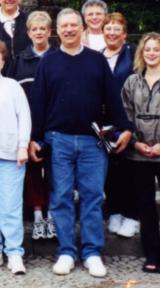}
\includegraphics[height=0.15\textheight]{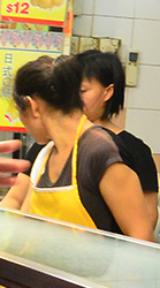}
\caption{}
\end{subfigure}
\begin{subfigure}[t]{0.14285714285714285\textwidth}
\centering
\includegraphics[height=0.15\textheight]{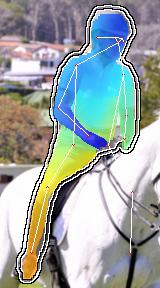}
\includegraphics[height=0.15\textheight]{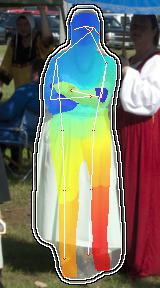}
\includegraphics[height=0.15\textheight]{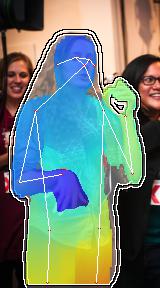}
\includegraphics[height=0.15\textheight]{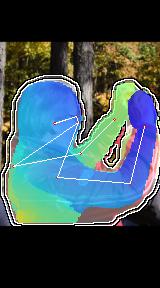}
\includegraphics[height=0.15\textheight]{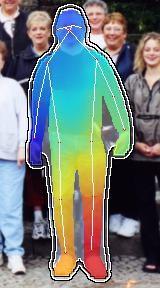}
\includegraphics[height=0.15\textheight]{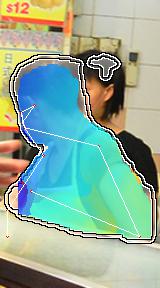}
\caption{}
\end{subfigure}
\begin{subfigure}[t]{0.14285714285714285\textwidth}
\centering
\includegraphics[height=0.15\textheight]{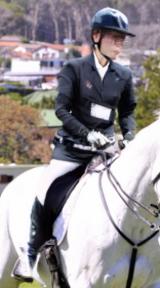}
\includegraphics[height=0.15\textheight]{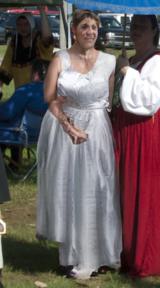}
\includegraphics[height=0.15\textheight]{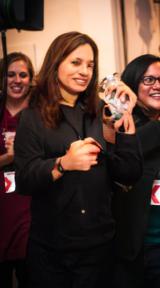}
\includegraphics[height=0.15\textheight]{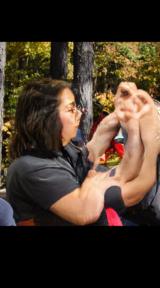}
\includegraphics[height=0.15\textheight]{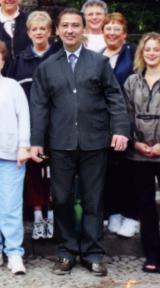}
\includegraphics[height=0.15\textheight]{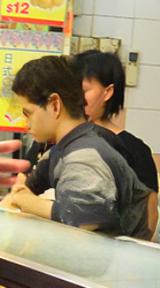}
\caption{}
\end{subfigure}
\begin{subfigure}[t]{0.14285714285714285\textwidth}
\centering
\includegraphics[height=0.15\textheight]{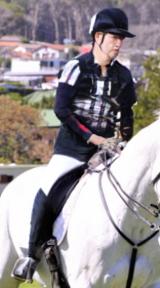}
\includegraphics[height=0.15\textheight]{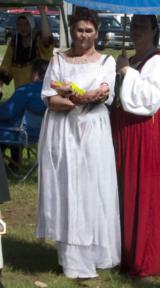}
\includegraphics[height=0.15\textheight]{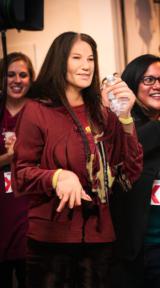}
\includegraphics[height=0.15\textheight]{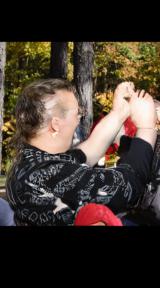}
\includegraphics[height=0.15\textheight]{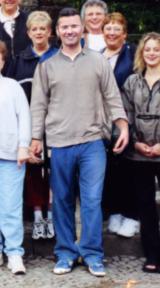}
\includegraphics[height=0.15\textheight]{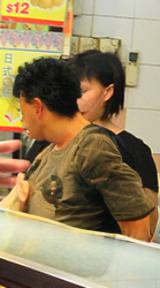}
\caption{}
\end{subfigure}
\begin{subfigure}[t]{0.14285714285714285\textwidth}
\centering
\includegraphics[height=0.15\textheight]{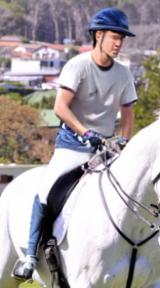}
\includegraphics[height=0.15\textheight]{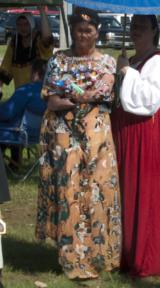}
\includegraphics[height=0.15\textheight]{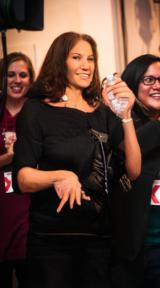}
\includegraphics[height=0.15\textheight]{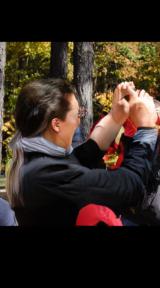}
\includegraphics[height=0.15\textheight]{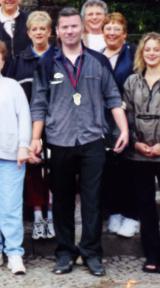}
\includegraphics[height=0.15\textheight]{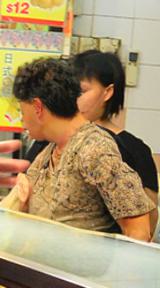}
\caption{}
\end{subfigure}
\begin{subfigure}[t]{0.14285714285714285\textwidth}
\centering
\includegraphics[height=0.15\textheight]{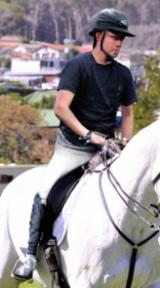}
\includegraphics[height=0.15\textheight]{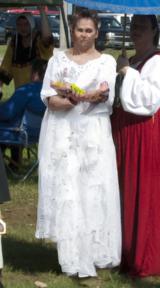}
\includegraphics[height=0.15\textheight]{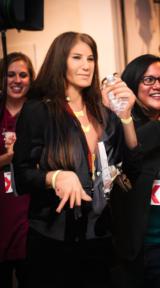}
\includegraphics[height=0.15\textheight]{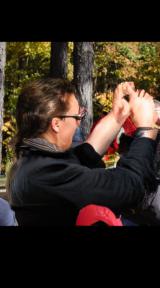}
\includegraphics[height=0.15\textheight]{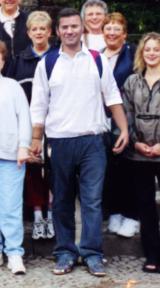}
\includegraphics[height=0.15\textheight]{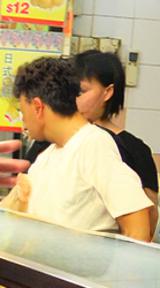}
\caption{}
\end{subfigure}
\begin{subfigure}[t]{0.14285714285714285\textwidth}
\centering
\includegraphics[height=0.15\textheight]{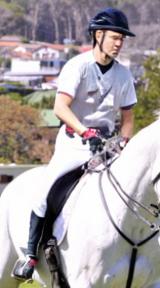}
\includegraphics[height=0.15\textheight]{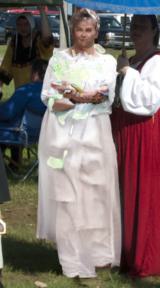}
\includegraphics[height=0.15\textheight]{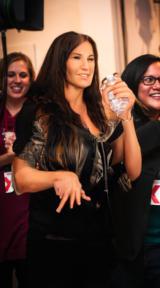}
\includegraphics[height=0.15\textheight]{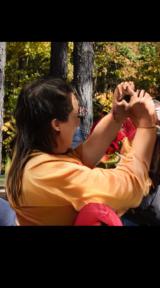}
\includegraphics[height=0.15\textheight]{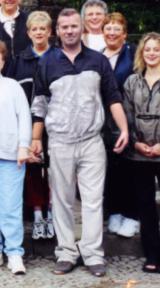}
\includegraphics[height=0.15\textheight]{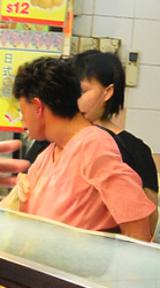}
\caption{}
\end{subfigure}
\caption{Randomly generated images from the FDH dataset. (a) shows the original image, (b) is the conditional input, (c) is the generated result for the unconditional generator, and (c-g) are from the CSE-guided generator.}
\label{fig:random_fdh2}
\end{figure*}

\begin{figure*}[t]
\centering
\begin{subfigure}[t]{0.14285714285714285\textwidth}
\centering
\includegraphics[height=0.15\textheight]{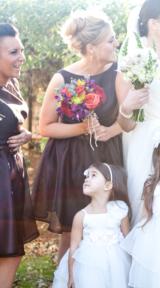}
\includegraphics[height=0.15\textheight]{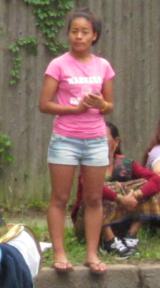}
\includegraphics[height=0.15\textheight]{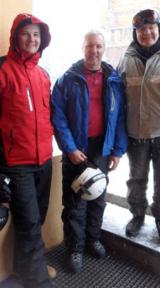}
\includegraphics[height=0.15\textheight]{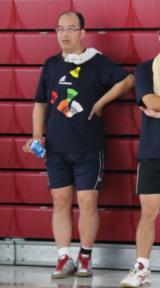}
\includegraphics[height=0.15\textheight]{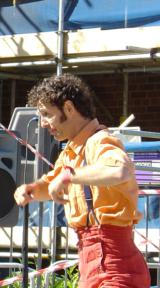}
\caption{}
\end{subfigure}
\begin{subfigure}[t]{0.14285714285714285\textwidth}
\centering
\includegraphics[height=0.15\textheight]{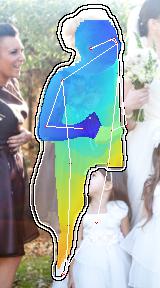}
\includegraphics[height=0.15\textheight]{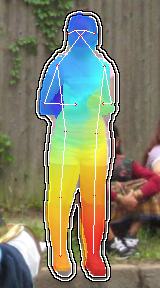}
\includegraphics[height=0.15\textheight]{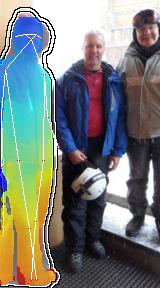}
\includegraphics[height=0.15\textheight]{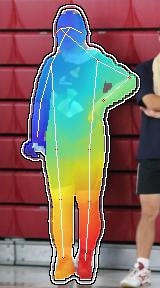}
\includegraphics[height=0.15\textheight]{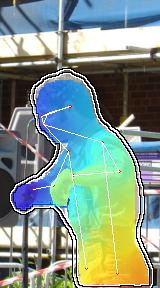}
\caption{}
\end{subfigure}
\begin{subfigure}[t]{0.14285714285714285\textwidth}
\centering
\includegraphics[height=0.15\textheight]{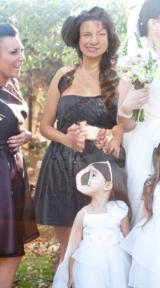}
\includegraphics[height=0.15\textheight]{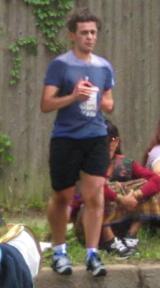}
\includegraphics[height=0.15\textheight]{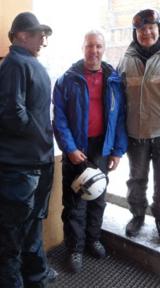}
\includegraphics[height=0.15\textheight]{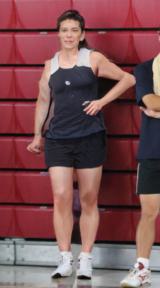}
\includegraphics[height=0.15\textheight]{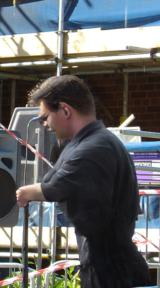}
\caption{}
\end{subfigure}
\begin{subfigure}[t]{0.14285714285714285\textwidth}
\centering
\includegraphics[height=0.15\textheight]{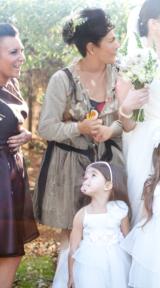}
\includegraphics[height=0.15\textheight]{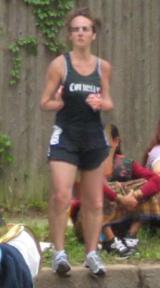}
\includegraphics[height=0.15\textheight]{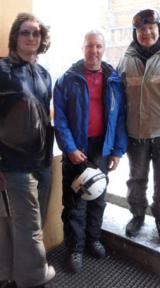}
\includegraphics[height=0.15\textheight]{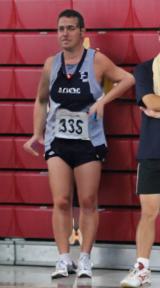}
\includegraphics[height=0.15\textheight]{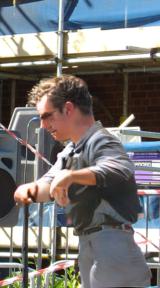}
\caption{}
\end{subfigure}
\begin{subfigure}[t]{0.14285714285714285\textwidth}
\centering
\includegraphics[height=0.15\textheight]{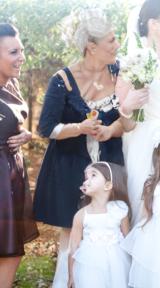}
\includegraphics[height=0.15\textheight]{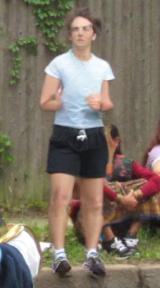}
\includegraphics[height=0.15\textheight]{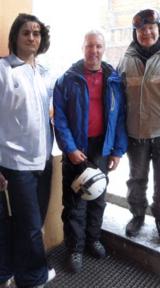}
\includegraphics[height=0.15\textheight]{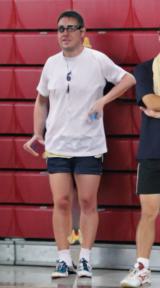}
\includegraphics[height=0.15\textheight]{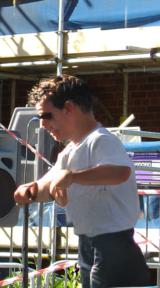}
\caption{}
\end{subfigure}
\begin{subfigure}[t]{0.14285714285714285\textwidth}
\centering
\includegraphics[height=0.15\textheight]{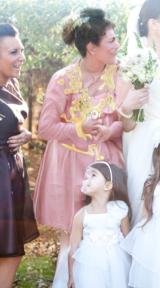}
\includegraphics[height=0.15\textheight]{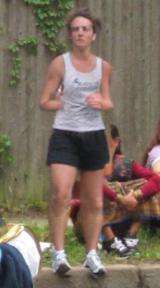}
\includegraphics[height=0.15\textheight]{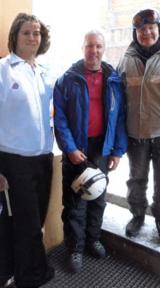}
\includegraphics[height=0.15\textheight]{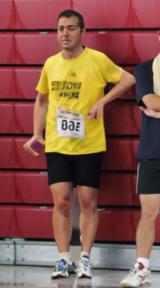}
\includegraphics[height=0.15\textheight]{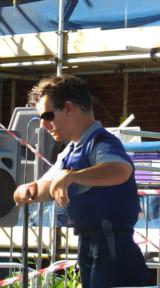}
\caption{}
\end{subfigure}
\begin{subfigure}[t]{0.14285714285714285\textwidth}
\centering
\includegraphics[height=0.15\textheight]{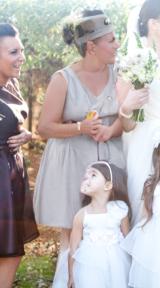}
\includegraphics[height=0.15\textheight]{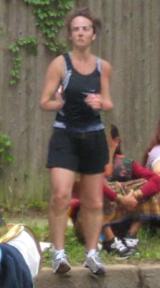}
\includegraphics[height=0.15\textheight]{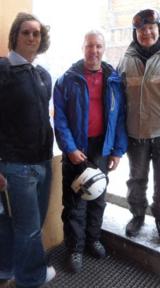}
\includegraphics[height=0.15\textheight]{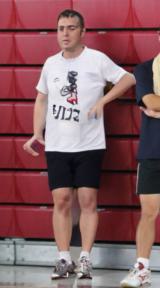}
\includegraphics[height=0.15\textheight]{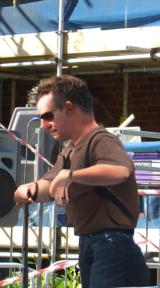}
\caption{}
\end{subfigure}
\caption{Randomly generated images from the FDH dataset. (a) shows the original image, (b) is the conditional input, (c) is the generated result for the unconditional generator, and (c-g) are from the CSE-guided generator.}
\label{fig:random_fdh3}
\end{figure*}

\paragraph{FDH}
Randomly selected images from the FDH \cite{Lin2014COCO} dataset are shown in \Cref{fig:random_fdh1}, \Cref{fig:random_fdh2}, and \Cref{fig:random_fdh3}

\begin{figure*}[t]
\centering
\begin{subfigure}[t]{0.16666666666666666\textwidth}
\centering
\includegraphics[width=\textwidth]{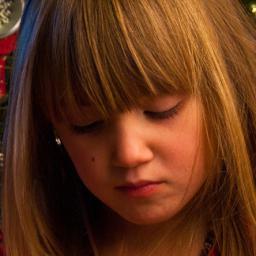}
\includegraphics[width=\textwidth]{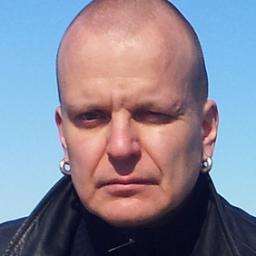}
\includegraphics[width=\textwidth]{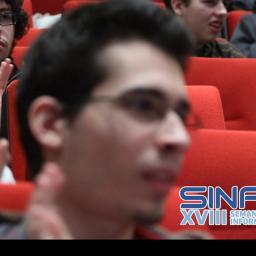}
\includegraphics[width=\textwidth]{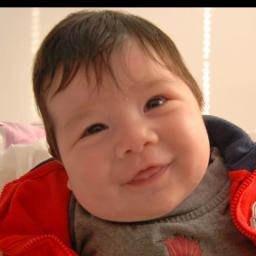}
\includegraphics[width=\textwidth]{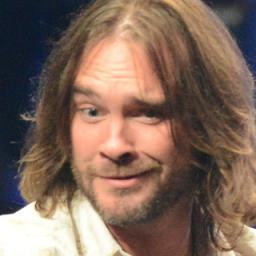}
\includegraphics[width=\textwidth]{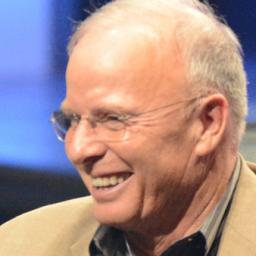}
\caption{}
\end{subfigure}
\begin{subfigure}[t]{0.16666666666666666\textwidth}
\centering
\includegraphics[width=\textwidth]{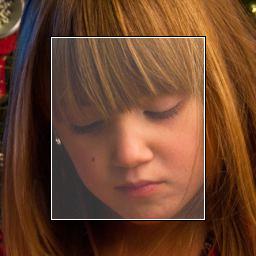}
\includegraphics[width=\textwidth]{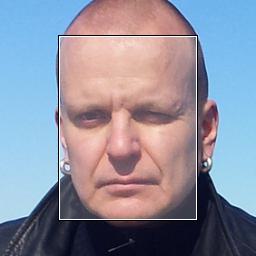}
\includegraphics[width=\textwidth]{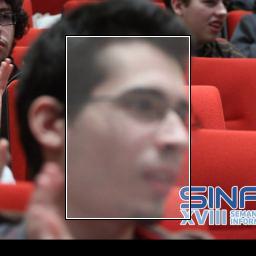}
\includegraphics[width=\textwidth]{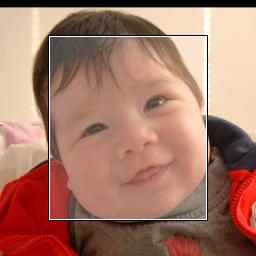}
\includegraphics[width=\textwidth]{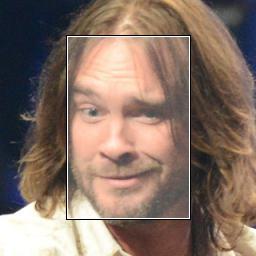}
\includegraphics[width=\textwidth]{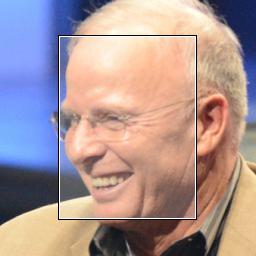}
\caption{}
\end{subfigure}
\begin{subfigure}[t]{0.16666666666666666\textwidth}
\centering
\includegraphics[width=\textwidth]{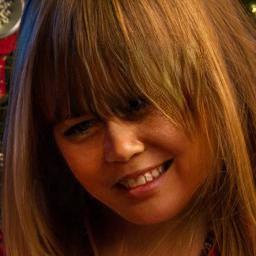}
\includegraphics[width=\textwidth]{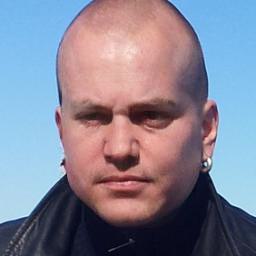}
\includegraphics[width=\textwidth]{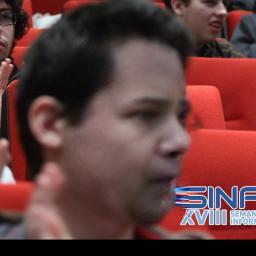}
\includegraphics[width=\textwidth]{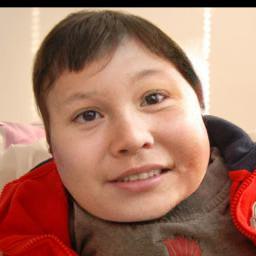}
\includegraphics[width=\textwidth]{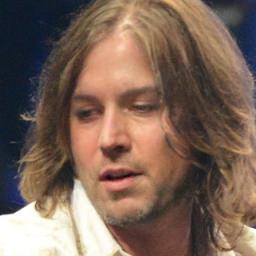}
\includegraphics[width=\textwidth]{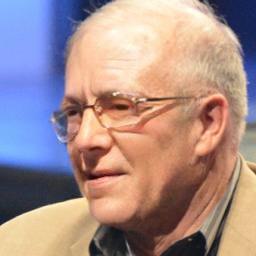}
\caption{}
\end{subfigure}
\begin{subfigure}[t]{0.16666666666666666\textwidth}
\centering
\includegraphics[width=\textwidth]{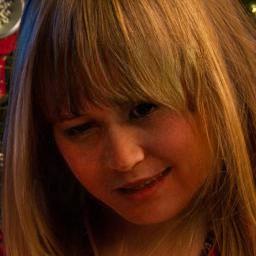}
\includegraphics[width=\textwidth]{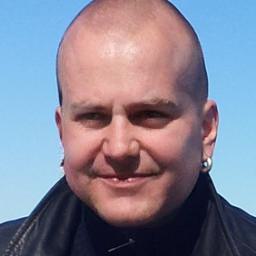}
\includegraphics[width=\textwidth]{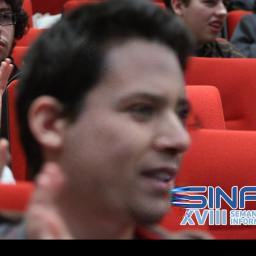}
\includegraphics[width=\textwidth]{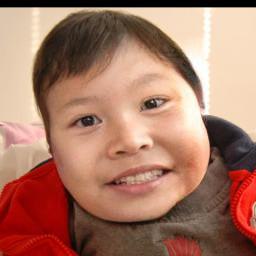}
\includegraphics[width=\textwidth]{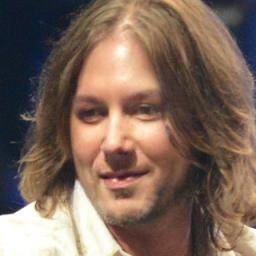}
\includegraphics[width=\textwidth]{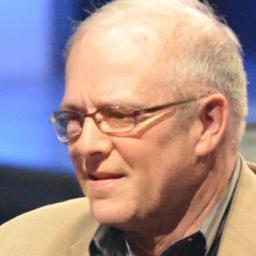}
\caption{}
\end{subfigure}
\begin{subfigure}[t]{0.16666666666666666\textwidth}
\centering
\includegraphics[width=\textwidth]{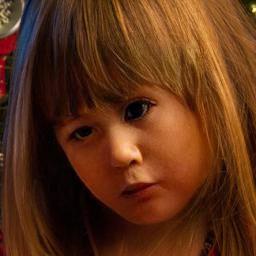}
\includegraphics[width=\textwidth]{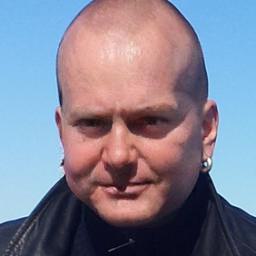}
\includegraphics[width=\textwidth]{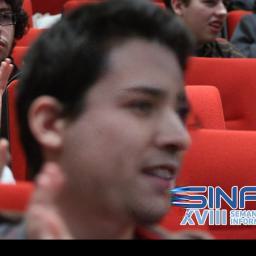}
\includegraphics[width=\textwidth]{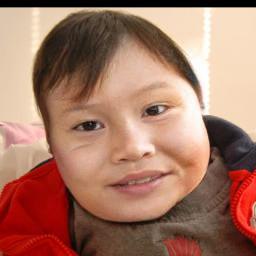}
\includegraphics[width=\textwidth]{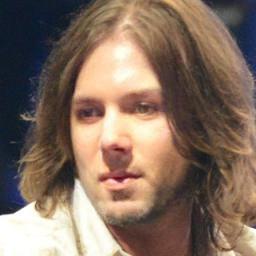}
\includegraphics[width=\textwidth]{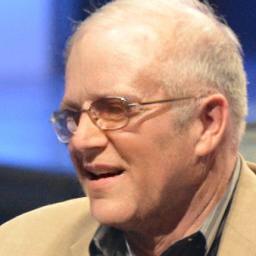}
\caption{}
\end{subfigure}
\begin{subfigure}[t]{0.16666666666666666\textwidth}
\centering
\includegraphics[width=\textwidth]{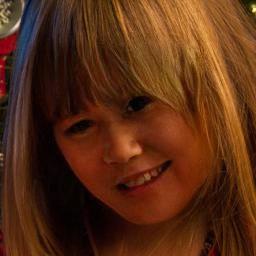}
\includegraphics[width=\textwidth]{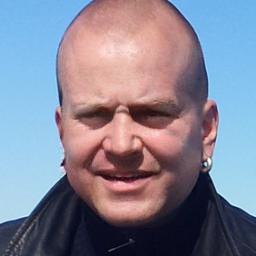}
\includegraphics[width=\textwidth]{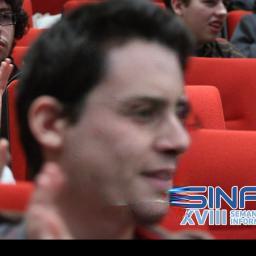}
\includegraphics[width=\textwidth]{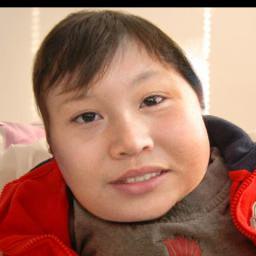}
\includegraphics[width=\textwidth]{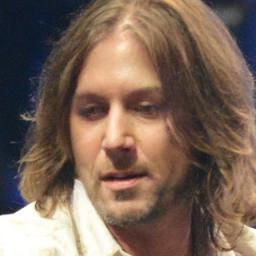}
\includegraphics[width=\textwidth]{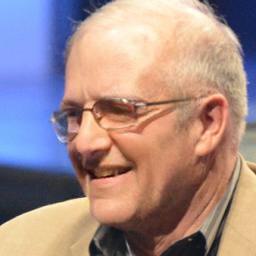}
\caption{}
\end{subfigure}
\caption{Randomly generated images from the FDF dataset. (a) shows the original image, (b) is the conditional input, and (b-f) are generated images.
No latent truncation is applied for the generated images.}
\label{fig:random_fdf1}
\end{figure*}

\begin{figure*}[t]
\centering
\begin{subfigure}[t]{0.16666666666666666\textwidth}
\centering
\includegraphics[width=\textwidth]{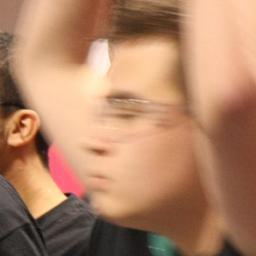}
\includegraphics[width=\textwidth]{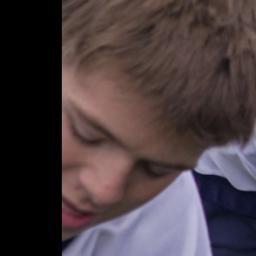}
\includegraphics[width=\textwidth]{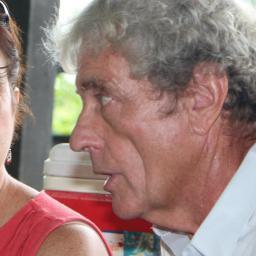}
\includegraphics[width=\textwidth]{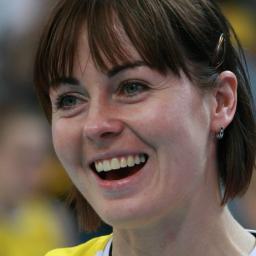}
\includegraphics[width=\textwidth]{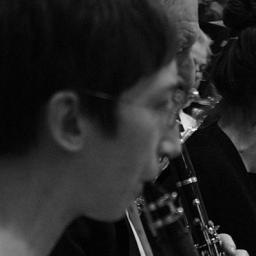}
\includegraphics[width=\textwidth]{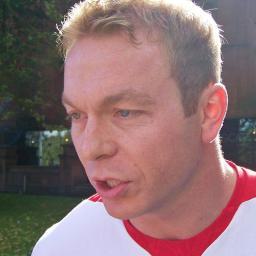}
\caption{}
\end{subfigure}
\begin{subfigure}[t]{0.16666666666666666\textwidth}
\centering
\includegraphics[width=\textwidth]{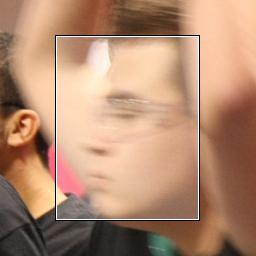}
\includegraphics[width=\textwidth]{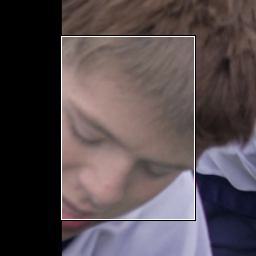}
\includegraphics[width=\textwidth]{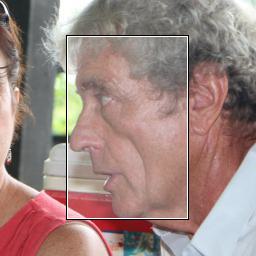}
\includegraphics[width=\textwidth]{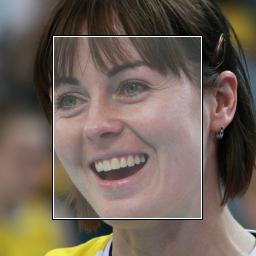}
\includegraphics[width=\textwidth]{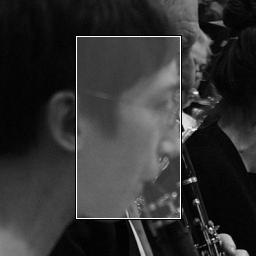}
\includegraphics[width=\textwidth]{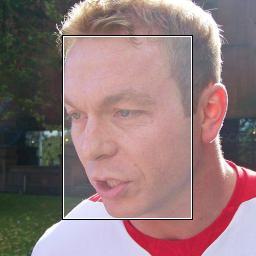}
\caption{}
\end{subfigure}
\begin{subfigure}[t]{0.16666666666666666\textwidth}
\centering
\includegraphics[width=\textwidth]{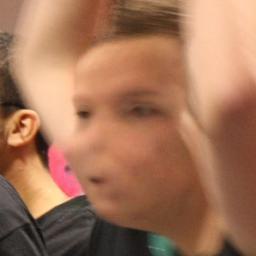}
\includegraphics[width=\textwidth]{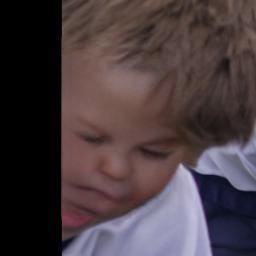}
\includegraphics[width=\textwidth]{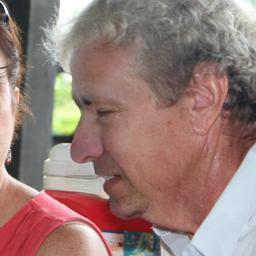}
\includegraphics[width=\textwidth]{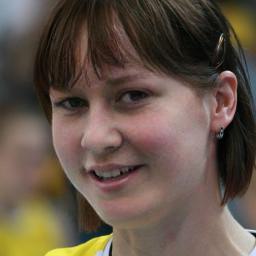}
\includegraphics[width=\textwidth]{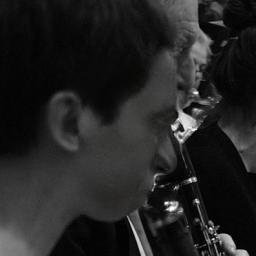}
\includegraphics[width=\textwidth]{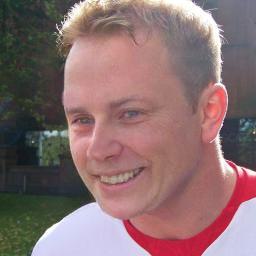}
\caption{}
\end{subfigure}
\begin{subfigure}[t]{0.16666666666666666\textwidth}
\centering
\includegraphics[width=\textwidth]{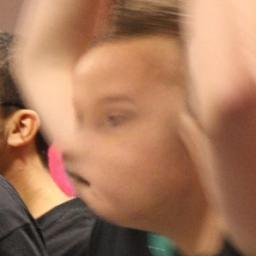}
\includegraphics[width=\textwidth]{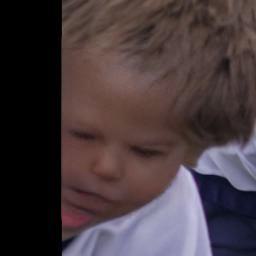}
\includegraphics[width=\textwidth]{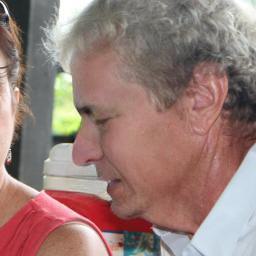}
\includegraphics[width=\textwidth]{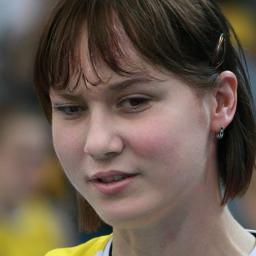}
\includegraphics[width=\textwidth]{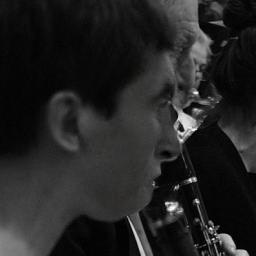}
\includegraphics[width=\textwidth]{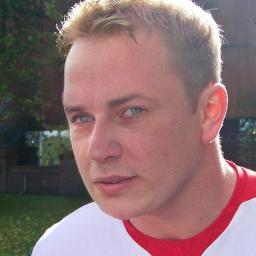}
\caption{}
\end{subfigure}
\begin{subfigure}[t]{0.16666666666666666\textwidth}
\centering
\includegraphics[width=\textwidth]{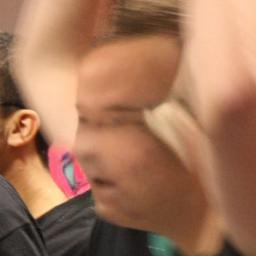}
\includegraphics[width=\textwidth]{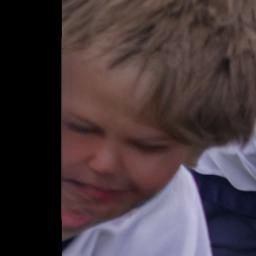}
\includegraphics[width=\textwidth]{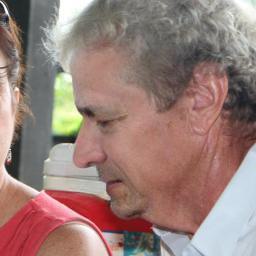}
\includegraphics[width=\textwidth]{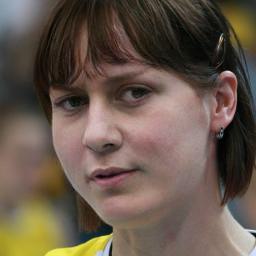}
\includegraphics[width=\textwidth]{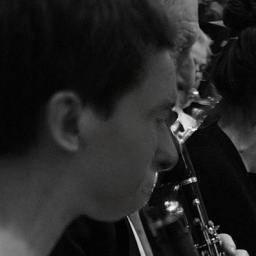}
\includegraphics[width=\textwidth]{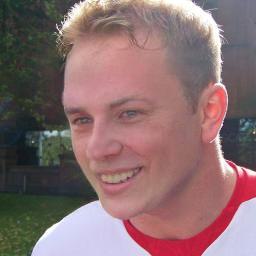}
\caption{}
\end{subfigure}
\begin{subfigure}[t]{0.16666666666666666\textwidth}
\centering
\includegraphics[width=\textwidth]{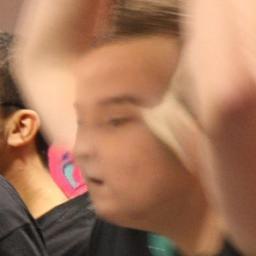}
\includegraphics[width=\textwidth]{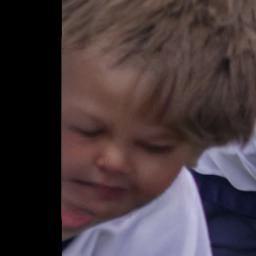}
\includegraphics[width=\textwidth]{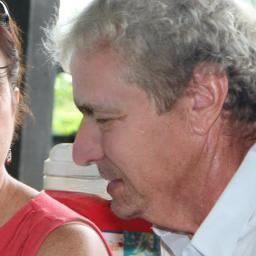}
\includegraphics[width=\textwidth]{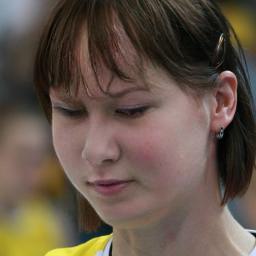}
\includegraphics[width=\textwidth]{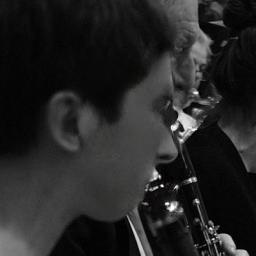}
\includegraphics[width=\textwidth]{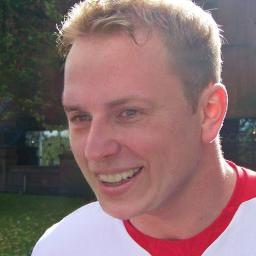}
\caption{}
\end{subfigure}
\caption{Randomly generated images from the FDF dataset. (a) shows the original image, (b) is the conditional input, and (b-f) are generated images.
No latent truncation is applied for the generated images.}
\label{fig:random_fdf2}
\end{figure*}

\begin{figure*}[t]
\centering
\begin{subfigure}[t]{0.16666666666666666\textwidth}
\centering
\includegraphics[width=\textwidth]{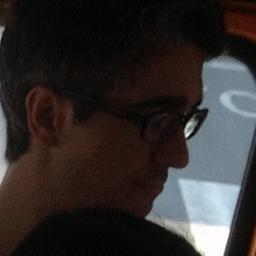}
\includegraphics[width=\textwidth]{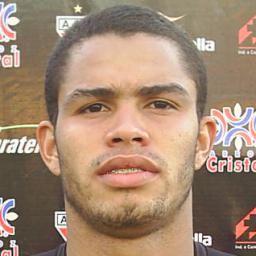}
\includegraphics[width=\textwidth]{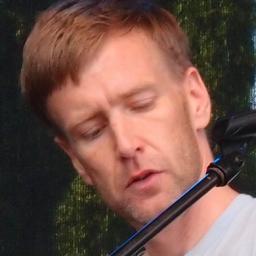}
\includegraphics[width=\textwidth]{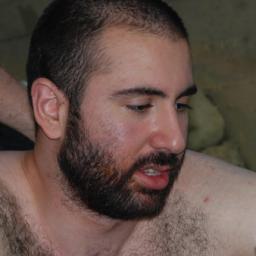}
\caption{}
\end{subfigure}
\begin{subfigure}[t]{0.16666666666666666\textwidth}
\centering
\includegraphics[width=\textwidth]{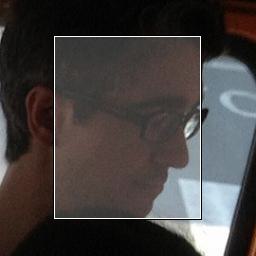}
\includegraphics[width=\textwidth]{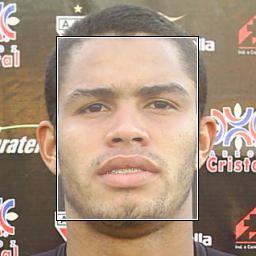}
\includegraphics[width=\textwidth]{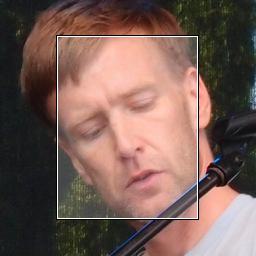}
\includegraphics[width=\textwidth]{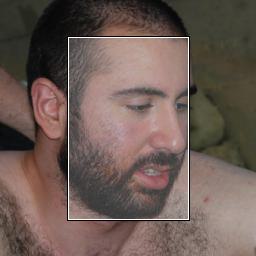}
\caption{}
\end{subfigure}
\begin{subfigure}[t]{0.16666666666666666\textwidth}
\centering
\includegraphics[width=\textwidth]{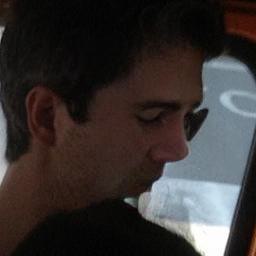}
\includegraphics[width=\textwidth]{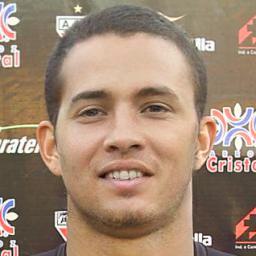}
\includegraphics[width=\textwidth]{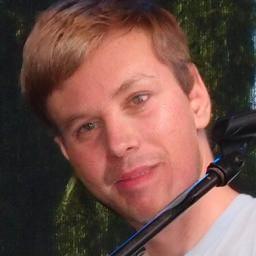}
\includegraphics[width=\textwidth]{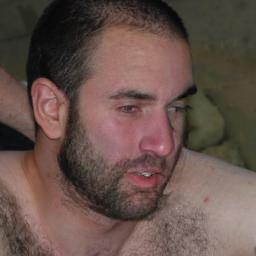}
\caption{}
\end{subfigure}
\begin{subfigure}[t]{0.16666666666666666\textwidth}
\centering
\includegraphics[width=\textwidth]{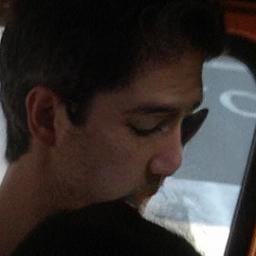}
\includegraphics[width=\textwidth]{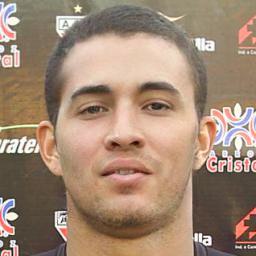}
\includegraphics[width=\textwidth]{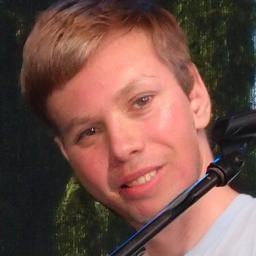}
\includegraphics[width=\textwidth]{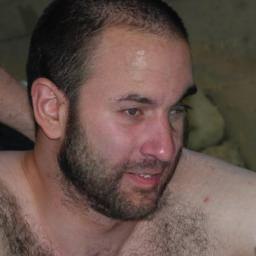}
\caption{}
\end{subfigure}
\begin{subfigure}[t]{0.16666666666666666\textwidth}
\centering
\includegraphics[width=\textwidth]{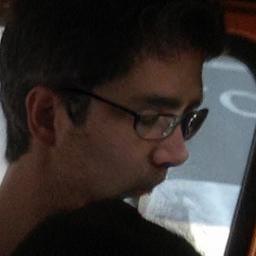}
\includegraphics[width=\textwidth]{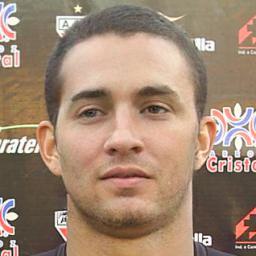}
\includegraphics[width=\textwidth]{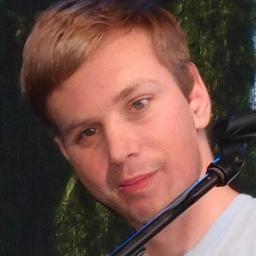}
\includegraphics[width=\textwidth]{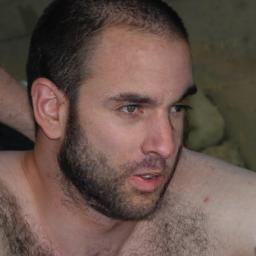}
\caption{}
\end{subfigure}
\begin{subfigure}[t]{0.16666666666666666\textwidth}
\centering
\includegraphics[width=\textwidth]{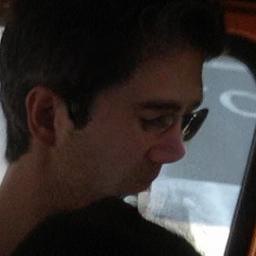}
\includegraphics[width=\textwidth]{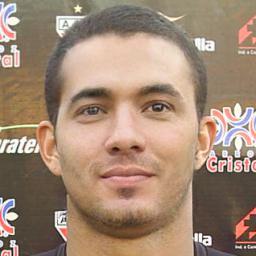}
\includegraphics[width=\textwidth]{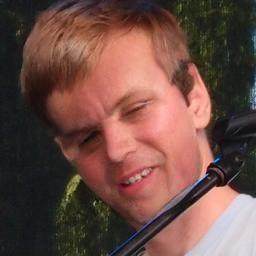}
\includegraphics[width=\textwidth]{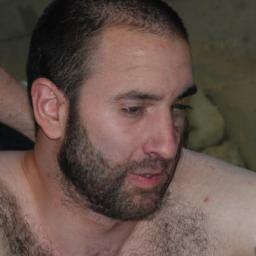}
\caption{}
\end{subfigure}
\caption{
    Randomly generated images from the FDF dataset. (a) shows the original image, (b) is the conditional input, and (b-f) are generated images.
    No latent truncation is applied for the generated images.}
\label{fig:random_fdf3}
\end{figure*}

\paragraph{FDF256}
Randomly selected images from the FDF256 \cite{Lin2014COCO} dataset are shown in \Cref{fig:random_fdf1}, \Cref{fig:random_fdf2}, and \Cref{fig:random_fdf3}

\newpage
{\small
\bibliographystyle{ieee_fullname}
\bibliography{bib_files/latex-wacv2023}}

\begin{thebibliography}{10}\itemsep=-1pt

\bibitem{Balakrishnan2018}
Guha Balakrishnan, Amy Zhao, Adrian~V Dalca, Fredo Durand, and John Guttag.
\newblock {Synthesizing Images of Humans in Unseen Poses}.
\newblock In {\em 2018 IEEE/CVF Conference on Computer Vision and Pattern
  Recognition (CVPR)}, pages 8340--8348, 2018.

\bibitem{Brkic2017}
Karla Brkic, Ivan Sikiric, Tomislav Hrkac, and Zoran Kalafatic.
\newblock {I Know That Person: Generative Full Body and Face De-identification
  of People in Images}.
\newblock In {\em 2017 IEEE Conference on Computer Vision and Pattern
  Recognition Workshops (CVPRW)}, pages 1319--1328, 2017.

\bibitem{Chan2019}
Caroline Chan, Shiry Ginosar, Tinghui Zhou, and Alexei Efros.
\newblock {Everybody Dance Now}.
\newblock In {\em 2019 IEEE/CVF International Conference on Computer Vision
  (ICCV)}, pages 5932--5941, 2019.

\bibitem{cityscapes}
Marius Cordts, Mohamed Omran, Sebastian Ramos, Timo Rehfeld, Markus Enzweiler,
  Rodrigo Benenson, Uwe Franke, Stefan Roth, and Bernt Schiele.
\newblock {The Cityscapes Dataset for Semantic Urban Scene Understanding}.
\newblock In {\em 2016 IEEE Conference on Computer Vision and Pattern
  Recognition (CVPR)}, pages 3213--3223, 2016.

\bibitem{Deng2010ImageNet}
Jia Deng, Wei Dong, Richard Socher, Li-Jia Li, {Kai Li}, and {Li Fei-Fei}.
\newblock {ImageNet: A large-scale hierarchical image database}.
\newblock In {\em 2009 IEEE Conference on Computer Vision and Pattern
  Recognition (CVPR)}, pages 248--255, 2009.

\bibitem{Dolhansky2020}
Brian Dolhansky, Joanna Bitton, Ben Pflaum, Jikuo Lu, Russ Howes, Menglin Wang,
  and Cristian~Canton Ferrer.
\newblock {The DeepFake Detection Challenge (DFDC) Dataset}.
\newblock 2020.

\bibitem{Gafni2019}
Oran Gafni, Lior Wolf, and Yaniv Taigman.
\newblock {Live Face De-Identification in Video}.
\newblock In {\em 2019 IEEE/CVF International Conference on Computer Vision
  (ICCV)}, pages 9377--9386, 2019.

\bibitem{Gross2006}
Ralph Gross, Latanya Sweeney, F. de~la Torre, and Simon Baker.
\newblock {Model-Based Face De-Identification}.
\newblock In {\em 2006 Conference on Computer Vision and Pattern Recognition
  Workshop (CVPRW)}, pages 161--161, 2006.

\bibitem{Han2018}
Xintong Han, Zuxuan Wu, Zhe Wu, Ruichi Yu, and Larry~S Davis.
\newblock {VITON: An Image-Based Virtual Try-on Network}.
\newblock In {\em 2018 IEEE/CVF Conference on Computer Vision and Pattern
  Recognition (CVPR)}, pages 7543--7552, 2018.

\bibitem{He2017}
Kaiming He, Georgia Gkioxari, Piotr Dollar, and Ross Girshick.
\newblock {Mask R-CNN}.
\newblock In {\em 2017 IEEE International Conference on Computer Vision
  (ICCV)}, pages 2980--2988. IEEE, oct 2017.

\bibitem{He2016resnet}
Kaiming He, Xiangyu Zhang, Shaoqing Ren, and Jian Sun.
\newblock {Deep Residual Learning for Image Recognition}.
\newblock In {\em 2016 IEEE Conference on Computer Vision and Pattern
  Recognition (CVPR)}, pages 770--778, 2016.

\bibitem{Heusel2017FID}
Martin Heusel, Hubert Ramsauer, Thomas Unterthiner, Bernhard Nessler, and Sepp
  Hochreiter.
\newblock {Gans trained by a two time-scale update rule converge to a local
  nash equilibrium}.
\newblock In {\em Advances in Neural Information Processing Systems}, pages
  6626--6637, 2017.

\bibitem{Hukkelas2019}
H{\aa}kon Hukkel{\aa}s, Rudolf Mester, and Frank Lindseth.
\newblock {DeepPrivacy: A Generative Adversarial Network for Face
  Anonymization}.
\newblock In George Bebis, Richard Boyle, Bahram Parvin, Darko Koracin, Daniela
  Ushizima, Sek Chai, Shinjiro Sueda, Xin Lin, Aidong Lu, Daniel Thalmann,
  Chaoli Wang, and Panpan Xu, editors, {\em Advances in Visual Computing},
  pages 565--578. Springer International Publishing, 2019.

\bibitem{Hukkelas2022}
H{\aa}kon Hukkel{\aa}s, Morten Smebye, Rudolf Mester, and Frank Lindseth.
\newblock Realistic full-body anonymization with surface-guided gans.
\newblock {\em arXiv preprint arXiv:2201.02193}, 2022.

\bibitem{Abarca2021handbook}
Arun {Jain, Anil and Flynn, Patrick and Ross}.
\newblock {\em {Handbook of Biometrics}}.
\newblock Springer US, Boston, MA, 2008.

\bibitem{Jourabloo2015}
Amin Jourabloo, Xi Yin, and Xiaoming Liu.
\newblock {Attribute preserved face de-identification}.
\newblock {\em 2015 International Conference on Biometrics (ICB)}, pages
  278--285, 2015.

\bibitem{Karras2018}
Tero Karras, Samuli Laine, and Timo Aila.
\newblock {A Style-Based Generator Architecture for Generative Adversarial
  Networks}.
\newblock In {\em 2019 IEEE/CVF Conference on Computer Vision and Pattern
  Recognition (CVPR)}, pages 4396--4405, 2019.

\bibitem{Karras2019b}
Tero Karras, Samuli Laine, Miika Aittala, Janne Hellsten, Jaakko Lehtinen, and
  Timo Aila.
\newblock {Analyzing and Improving the Image Quality of StyleGAN}.
\newblock In {\em 2020 IEEE/CVF Conference on Computer Vision and Pattern
  Recognition (CVPR)}, pages 8107--8116, 2020.

\bibitem{Kingma2014Adam}
Diederik~P Kingma and Jimmy Ba.
\newblock {Adam: A Method for Stochastic Optimization}.
\newblock In {\em International Conference on Learning Representations}, 2015.

\bibitem{Kocasari2022stylemc}
Umut Kocasari, Alara Dirik, Mert Tiftikci, and Pinar Yanardag.
\newblock {StyleMC: Multi-Channel Based Fast Text-Guided Image Generation and
  Manipulation}.
\newblock In {\em 2022 IEEE/CVF Winter Conference on Applications of Computer
  Vision (WACV)}, pages 3441--3450, 2022.

\bibitem{kynkaanniemi2022role}
Tuomas Kynk{\"{a}}{\"{a}}nniemi, Tero Karras, Miika Aittala, Timo Aila, and
  Jaakko Lehtinen.
\newblock {The Role of ImageNet Classes in Frechet Inception Distance}.
\newblock {\em arXiv preprint arXiv:2203.06026}, 2022.

\bibitem{idealods2018imagequalityassessment}
Christopher Lennan, Hao Nguyen, and Dat Tran.
\newblock {Image Quality Assessment}.
\newblock \url{https://github.com/idealo/image-quality-assessment}, 2018.

\bibitem{Li2018c}
Jian Li, Yabiao Wang, Changan Wang, Ying Tai, Jianjun Qian, Jian Yang, Chengjie
  Wang, Jilin Li, and Feiyue Huang.
\newblock {DSFD: Dual Shot Face Detector}.
\newblock In {\em 2019 IEEE/CVF Conference on Computer Vision and Pattern
  Recognition (CVPR)}, pages 5055--5064, 2019.

\bibitem{Li2019}
Yining Li, Chen Huang, and Chen~Change Loy.
\newblock {Dense Intrinsic Appearance Flow for Human Pose Transfer}.
\newblock In {\em 2019 IEEE/CVF Conference on Computer Vision and Pattern
  Recognition (CVPR)}, pages 3688--3697, 2019.

\bibitem{Lin2014COCO}
Tsung-Yi Lin, Michael Maire, Serge Belongie, James Hays, Pietro Perona, Deva
  Ramanan, Piotr Doll{\'{a}}r, and C.~Lawrence Zitnick.
\newblock {Microsoft COCO: Common Objects in Context}.
\newblock In {\em Computer Vision -- ECCV 2014}, pages 740--755. Springer
  International Publishing, 2014.

\bibitem{Liu2016a}
Ziwei Liu, Ping Luo, Shi Qiu, Xiaogang Wang, and Xiaoou Tang.
\newblock {DeepFashion: Powering Robust Clothes Recognition and Retrieval with
  Rich Annotations}.
\newblock In {\em 2016 IEEE Conference on Computer Vision and Pattern
  Recognition (CVPR)}, pages 1096--1104, 2016.

\bibitem{Loper2015}
Matthew Loper, Naureen Mahmood, Javier Romero, Gerard Pons-Moll, and Michael~J.
  Black.
\newblock {SMPL}.
\newblock {\em ACM Transactions on Graphics}, 34(6):1--16, nov 2015.

\bibitem{Sun2018person}
Liqian Ma, Qianru Sun, Stamatios Georgoulis, Luc {Van Gool}, Bernt Schiele, and
  Mario Fritz.
\newblock {Disentangled Person Image Generation}.
\newblock In {\em 2018 IEEE/CVF Conference on Computer Vision and Pattern
  Recognition (CVPR)}, pages 99--108, 2018.

\bibitem{Maximov2020}
Maxim Maximov, Ismail Elezi, and Laura Leal-Taixe.
\newblock {CIAGAN: Conditional Identity Anonymization Generative Adversarial
  Networks}.
\newblock In {\em 2020 IEEE/CVF Conference on Computer Vision and Pattern
  Recognition (CVPR)}, pages 5446--5455, 2020.

\bibitem{Mokady2022}
Ron Mokady, Michal Yarom, Omer Tov, Oran Lang, Daniel Cohen-Or, Tali Dekel,
  Michal Irani, and Inbar Mosseri.
\newblock {Self-Distilled StyleGAN: Towards Generation from Internet Photos}.
\newblock 2022.

\bibitem{motpy}
Wiktor Muron.
\newblock motpy - simple multi object tracking library.
\newblock \url{https://github.com/wmuron/motpy}, 2022.

\bibitem{Neverova2018}
Natalia Neverova, Rıza {Alp G{\"{u}}ler}, and Iasonas Kokkinos.
\newblock {Dense Pose Transfer}.
\newblock In {\em Computer Vision -- ECCV 2018}, pages 128--143. Springer
  International Publishing, 2018.

\bibitem{Neverova2020}
Natalia Neverova, David Novotny, Vasil Khalidov, Marc Szafraniec, Patrick
  Labatut, and Andrea Vedaldi.
\newblock {Continuous Surface Embeddings}.
\newblock In {\em Advances in Neural Information Processing Systems}, pages
  17258----17270. Curran Associates, Inc., 2020.

\bibitem{Newton2005}
E.M. Newton, L. Sweeney, and B. Malin.
\newblock {Preserving privacy by de-identifying face images}.
\newblock {\em IEEE Transactions on Knowledge and Data Engineering},
  17(2):232--243, 2005.

\bibitem{obukhov2020torchfidelity}
Anton Obukhov, Maximilian Seitzer, Po-Wei Wu, Semen Zhydenko, Jonathan Kyl, and
  Elvis Yu-Jing Lin.
\newblock {High-fidelity performance metrics for generative models in PyTorch}.
\newblock \url{https://github.com/toshas/torch-fidelity}, 2020.

\bibitem{paszke2019pytorch}
Adam Paszke, Sam Gross, Francisco Massa, Adam Lerer, James Bradbury, Gregory
  Chanan, Trevor Killeen, Zeming Lin, Natalia Gimelshein, Luca Antiga, and
  Others.
\newblock {Pytorch: An imperative style, high-performance deep learning
  library}.
\newblock {\em Advances in neural information processing systems}, 32, 2019.

\bibitem{Radford2021CLIP}
Alec Radford, Jong~Wook Kim, Chris Hallacy, Aditya Ramesh, Gabriel Goh,
  Sandhini Agarwal, Girish Sastry, Amanda Askell, Pamela Mishkin, Jack Clark,
  Gretchen Krueger, and Ilya Sutskever.
\newblock {Learning Transferable Visual Models From Natural Language
  Supervision}.
\newblock In {\em International Conference on Machine Learning}, pages
  8748----8763, 2021.

\bibitem{Ren2018a}
Zhongzheng Ren, Yong~Jae Lee, and Michael~S Ryoo.
\newblock {Learning to Anonymize Faces for Privacy Preserving Action
  Detection}.
\newblock In {\em Computer Vision -- ECCV 2018}, pages 639--655. Springer
  International Publishing, Cham, 2018.

\bibitem{Ronneberger2015}
Olaf Ronneberger, Philipp Fischer, and Thomas Brox.
\newblock {U-Net: Convolutional Networks for Biomedical Image Segmentation}.
\newblock In {\em International Conference on Medical image computing and
  computer-assisted intervention}, pages 234--241. Springer, 2015.

\bibitem{Sarkar2021}
Kripasindhu Sarkar, Vladislav Golyanik, Lingjie Liu, and Christian Theobalt.
\newblock {Style and Pose Control for Image Synthesis of Humans from a Single
  Monocular View}.
\newblock {\em arXiv preprint arXiv:2102.11263}, feb 2021.

\bibitem{Sarkar2020}
Kripasindhu Sarkar, Dushyant Mehta, Weipeng Xu, Vladislav Golyanik, and
  Christian Theobalt.
\newblock {Neural Re-rendering of Humans from a Single Image}.
\newblock In {\em Computer Vision -- ECCV 2020}, pages 596--613. Springer
  International Publishing, Cham, 2020.

\bibitem{sjalander2019epic}
Magnus Sj\"alander, Magnus Jahre, Gunnar Tufte, and Nico Reissmann.
\newblock {EPIC}: An energy-efficient, high-performance {GPGPU} computing
  research infrastructure, 2019.

\bibitem{Sun2018}
Qianru Sun, Liqian Ma, Seong {Joon Oh}, Luc~Van Gool, Bernt Schiele, and Mario
  Fritz.
\newblock {Natural and Effective Obfuscation by Head Inpainting}.
\newblock In {\em 2018 IEEE/CVF Conference on Computer Vision and Pattern
  Recognition (CVPR)}, pages 5050--5059, 2018.

\bibitem{Sun2018a}
Qianru Sun, Ayush Tewari, Weipeng Xu, Mario Fritz, Christian Theobalt, and
  Bernt Schiele.
\newblock {A Hybrid Model for Identity Obfuscation by Face Replacement}.
\newblock In {\em Computer Vision -- ECCV 2018}, pages 570--586. Springer
  International Publishing, Cham, 2018.

\bibitem{Talebi2018}
Hossein Talebi and Peyman Milanfar.
\newblock {NIMA: Neural Image Assessment}.
\newblock {\em IEEE Transactions on Image Processing}, 27(8):3998--4011, aug
  2018.

\bibitem{Thomee2015}
Bart Thomee, David~A. Shamma, Gerald Friedland, Benjamin Elizalde, Karl Ni,
  Douglas Poland, Damian Borth, and Li-Jia Li.
\newblock {YFCC100M}: The new data in multimedia research.
\newblock {\em Commun. ACM}, 59(2):64--73, jan 2016.

\bibitem{wu2019detectron2}
Yuxin Wu, Alexander Kirillov, Francisco Massa, Wan-Yen Lo, and Ross Girshick.
\newblock {Detectron2}.
\newblock \url{https://github.com/facebookresearch/detectron2}, 2019.

\bibitem{Xie2017resnext}
Saining Xie, Ross Girshick, Piotr Dollar, Zhuowen Tu, and Kaiming He.
\newblock {Aggregated Residual Transformations for Deep Neural Networks}.
\newblock In {\em 2017 IEEE Conference on Computer Vision and Pattern
  Recognition (CVPR)}, pages 5987--5995, 2017.

\bibitem{Yang2016WIDER}
Shuo Yang, Ping Luo, Chen~Change Loy, and Xiaoou Tang.
\newblock {WIDER FACE: A Face Detection Benchmark}.
\newblock In {\em 2016 IEEE Conference on Computer Vision and Pattern
  Recognition (CVPR)}, pages 5525--5533, 2016.

\bibitem{Yu2021}
Ning Yu, Vladislav Skripniuk, Sahar Abdelnabi, and Mario Fritz.
\newblock {Artificial Fingerprinting for Generative Models: Rooting Deepfake
  Attribution in Training Data}.
\newblock In {\em 2021 IEEE/CVF International Conference on Computer Vision
  (ICCV)}, pages 14428--14437, 2021.

\bibitem{Zheng2015}
Liang Zheng, Liyue Shen, Lu Tian, Shengjin Wang, Jingdong Wang, and Qi Tian.
\newblock {Scalable Person Re-identification: A Benchmark}.
\newblock In {\em 2015 IEEE International Conference on Computer Vision
  (ICCV)}, pages 1116--1124, 2015.

\bibitem{Zhou2019}
Kaiyang Zhou, Yongxin Yang, Andrea Cavallaro, and Tao Xiang.
\newblock {Omni-Scale Feature Learning for Person Re-Identification}.
\newblock In {\em 2019 IEEE/CVF International Conference on Computer Vision
  (ICCV)}, pages 3701--3711, 2019.

\end{thebibliography}

\end{document}